\documentclass[a4paper,12pt]{report}
\usepackage[utf8]{inputenc}
\usepackage[dvipsnames]{xcolor}
\usepackage{graphicx}
\usepackage{bm}
\usepackage{inputenc}
\usepackage{xfrac} 
\usepackage{caption}
\usepackage{subcaption}
\usepackage{float}
\usepackage{caption}
\usepackage{array} 
\usepackage{multirow} 
\usepackage{titling, titlesec} 

\usepackage{hyperref}
\hypersetup{
    colorlinks,
    citecolor={blue},
    filecolor=black,
    linkcolor=blue,
    urlcolor={blue}
}

\usepackage{mathtools}

\usepackage[margin=1in]{geometry} 
\numberwithin{equation}{chapter}
\numberwithin{section}{chapter}

\usepackage[titletoc]{appendix}
\usepackage[english]{babel}

\usepackage{geometry}
\titleformat{\chapter}[display]{\normalfont\huge\bfseries\centering}   {\chaptertitlename\ \thechapter}{20pt}{\Huge}
\titlespacing{\chapter}{0pt}{-32pt}{1cm}

\titleformat*{\section}{\normalfont \fontsize{16}{19.2} \bfseries}  

\usepackage{array}

\newcolumntype{+}{!{\vrule width 2pt}}

\newlength\savedwidth

\usepackage{algorithm}
\usepackage{algpseudocode}
\usepackage{tabularx}
\usepackage{booktabs}
\usepackage{multirow}
\usepackage{hyperref}
\usepackage{lscape}

\newcommand{\BE}[1]{\noindent \textbf{#1 \quad}}

\begin{document}
\begin{titlepage}
\centering
\topskip0pt
\vspace*{\fill}
\huge{\textbf{Augmentation is AUtO-Net}: \\
Augmentation-Driven Contrastive Multiview Learning for Medical Image Segmentation
}\\

\vspace{2cm}
\LARGE{Yanming Guo}\\
\vspace{1cm}
\vspace{1 cm}

\large{A thesis submitted in partial fulfillment of \\ the requirements for the degree of \\  Bachelor of Science (Honours)} \\
\vspace{1cm}
\large{Mathematics and Statistics}\\
\vspace{2cm}
\date{June 2023}
    \includegraphics[scale=0.75]{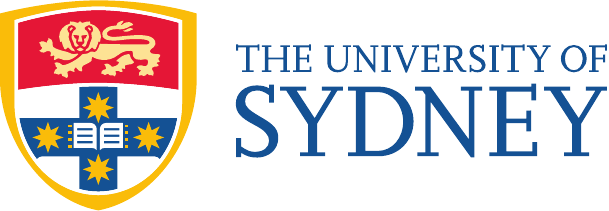}\\
\vspace{1cm}
\large{June 2023}\\
\vspace*{\fill}

\thispagestyle{empty}
\end{titlepage}

\pagenumbering{roman}

\pagebreak
\hspace{0pt}
\begin{center}
    \textbf{\large Statement of originality}\\
    \vspace{0.5cm}
\end{center}

\noindent This is to certify that to the best of my knowledge, the content of this thesis is my own work. This thesis has not been submitted for any degree or other purposes.\\
\\
\noindent I certify that the intellectual content of this thesis is the product of my own work and that all the assistance received in preparing this thesis and sources have been acknowledged.\\
\\
\\
\\
Yanming Guo

\pagebreak
\hspace{0pt}

\begin{center}
    \textbf{\large Abstract }\\
\end{center}
\vspace{0.5cm}
The utilisation of deep learning segmentation algorithms that learn complex organs and tissue patterns and extract essential regions of interest from the noisy background to improve the visual ability for medical image diagnosis has achieved impressive results in Medical Image Computing (MIC). This thesis focuses on retinal blood vessel segmentation tasks, providing an extensive literature review of deep learning-based medical image segmentation approaches while comparing the methodologies and empirical performances. The work also examines the limitations of current state-of-the-art methods by pointing out the two significant existing limitations: data size constraints and the dependency on high computational resources. To address such problems, this work proposes a novel efficient, simple multiview learning framework that contrastively learns invariant vessel feature representation by comparing with multiple augmented views by various transformations to overcome data shortage and improve generalisation ability. Moreover, the hybrid network architecture integrates the attention mechanism into a Convolutional Neural Network to further capture complex continuous curvilinear vessel structures. The result demonstrates the proposed method validated on the CHASE-DB1 dataset, attaining the highest F1 score of 83.46\% and the highest Intersection over Union (IOU) score of 71.62\% with UNet structure, surpassing existing benchmark UNet-based methods by 1.95\% and 2.8\%, respectively. The combination of the metrics indicates the model detects the vessel object accurately with a highly coincidental location with the ground truth. Moreover, the proposed approach could be trained within 30 minutes by consuming less than 3 GB GPU RAM, and such characteristics support the efficient implementation for real-world applications and deployments.

\bigskip
\bigskip

\newpage

\newpage

\begin{center}
    \textbf{\large Acknowledgements}\\
    \vspace{0.5cm}
\end{center}
   First and foremost, I would like to express my deepest gratitude to my family, especially my parents, for providing unwavering emotional and financial support during the challenging COVID era. Your constant encouragement has fueled my pursuit of new knowledge from diverse research perspectives.

    I want to thank my degree supervisors sincerely, A/Prof. Weidong Cai, Dr Dongnan Liu, and my future PhD supervisor, A/Prof. Jin Ma, for your invaluable academic guidance and mentorship.
    
    Furthermore, I am grateful for my closest friends, Mr Jiacheng Zhang and Miss Lu Liu, you have always been there to discuss new ideas and share insights, as well as all the dedicated teaching staff I have had the pleasure of encountering at the University of Sydney.
    
    I must also express my heartfelt thanks to my alma mater, the University of Melbourne, for fostering my growth by cultivating solid academic skills and helping me become the best version of myself.
    
    Looking back on the past four and a half years of my undergraduate journey, I am overwhelmed with cherished memories and gratitude for the many wonderful people who have supported me. I am genuinely thankful to all those who have contributed to my success, directly and indirectly. Without your unwavering support, I could not have overcome the challenges and reached this point.
\vfill
\hspace{0pt}
\pagebreak
\newpage
\tableofcontents
\newpage

\setcounter{page}{1}\pagenumbering{arabic}
 
\chapter{Introduction}
\label{Chapter1}
\section{Medical Image Segmentation Background}
Medical image segmentation tasks can be defined as a pixel-wise classification process to differentiate regions of interest from the background. In this context, pixels from the region of interest are categorised into one class while the background forms another. This process generates a binary image, enhancing visualisation and facilitating precise analysis of anatomical structures, tissues, and organs. Medical image segmentation has become indispensable in various clinical applications, including diagnosis, treatment planning, and disease monitoring. Its usage spans a multitude of imaging modalities, such as X-ray, Computed Tomography (CT), Magnetic Resonance Imaging (MRI), Ultrasound, and Positron Emission Tomography (PET) \cite{pham2000current, razavi2003three}.

Medical image segmentation techniques can be broadly classified into two categories. The first encompasses rule-based segmentation algorithms, including methods such as thresholding for object-background separation \cite{sezgin2004survey}, region-growing, which iteratively expands adjacent pixels from a selected seed based on predefined criteria \cite{295913}, and edge-based segmentation using operators like Canny or Sobel for detecting image-background boundaries \cite{canny1986computational}. While these rule-based algorithms offer advantages such as simplicity and efficiency in implementation, they hinge on prior knowledge - organ structure, threshold, colour, and textures. Particularly when segmentation is colour-based, these algorithms falter when applied to black-and-white images. Hence, rule-based algorithms' sensitivity to slight changes in input and their dependency on real-world images limits their utility. The second category comprises learning algorithms that eliminate the need for manually designed heuristics. These data-driven algorithms can learn semantic feature representations and recognise patterns \cite{theodoridis2006pattern} from complex real-world images. Prior studies have proven the effectiveness of deep-learning approaches in this area. Remarkably, the same network can cater to multiple biological structure segmentations across different modalities, requiring only a change in the training data. This enables the network to be adapted for new tasks, such as segmentation of prostate \cite{jia20193d, jia2019hd} and brain tumour \cite{jia2020h2nfnet, zhang2018ms} from magnetic resonance (MR) images, cell nuclei and cytoplasm segmentation in Pap smear images \cite{tareef2018multi, tareef2017automatic}, and COVID-19 lesion segmentation in CT images \cite{jia2023convolutional}.

\begin{figure}[h!]
  \centering
  \includegraphics[width = 0.8\textwidth]{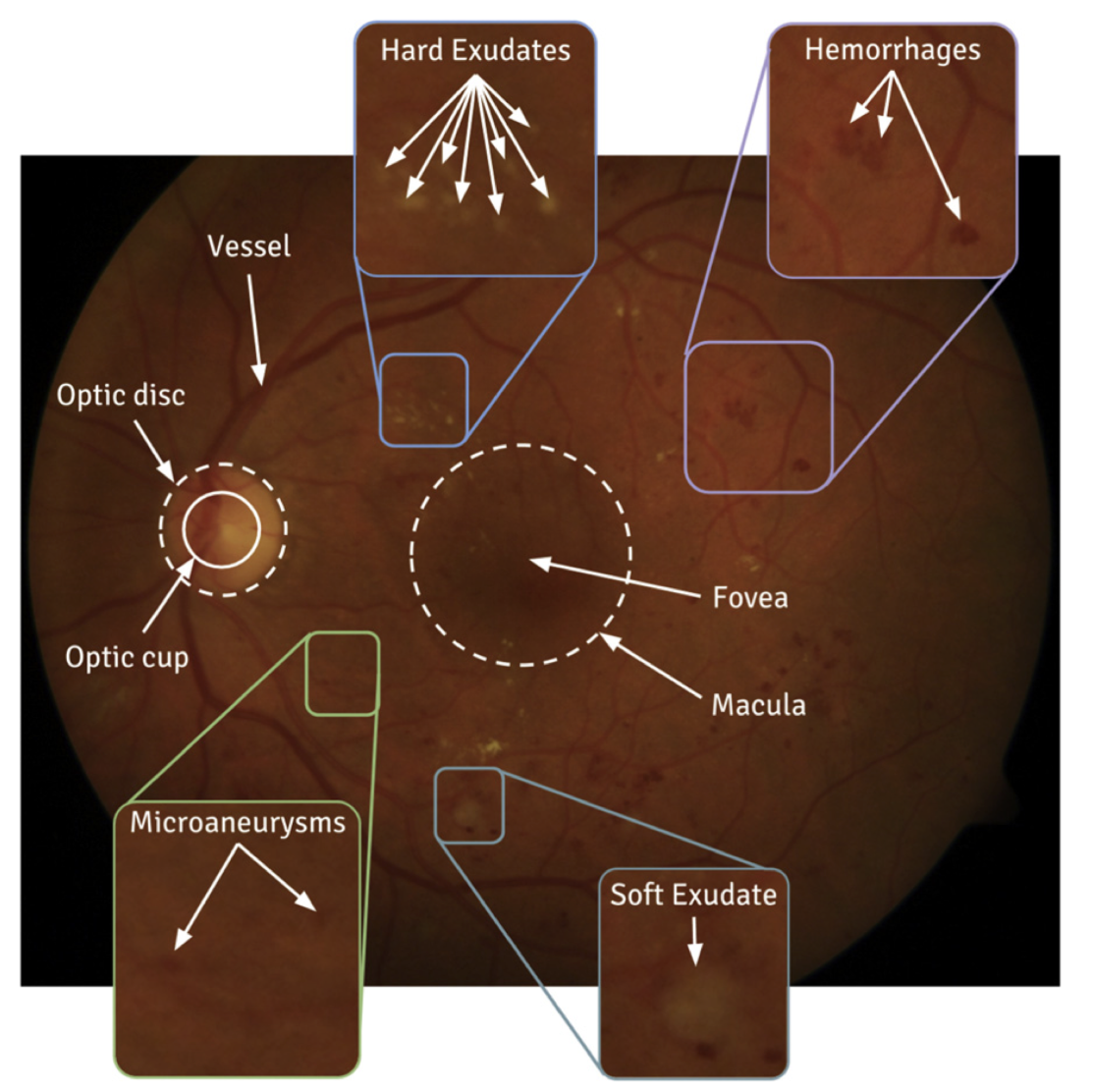}
  \caption{The fundus photograph states important retinal components \cite{sengupta2020ophthalmic}.}
  \label{fig:retina}
\end{figure}

This thesis concentrates on the task of retinal blood vessel segmentation, a critical process in diagnosing conditions such as diabetic retinopathy (DR), glaucoma, and age-related macular degeneration \cite{sengupta2020ophthalmic}. As shown in Figure \ref{fig:retina}, retinal images contain complex biological features. Therefore, visualising the blood vessel structure is vital to detecting early-stage diabetic retinopathy \cite{wu2018multiscale}. Accurately segmenting and analysing retinal blood vessels are crucial for diagnosing and treating these diseases.

\section{Aim and Importance}
\begin{figure}[h!]
  \centering
  \includegraphics[width = 0.9\textwidth]{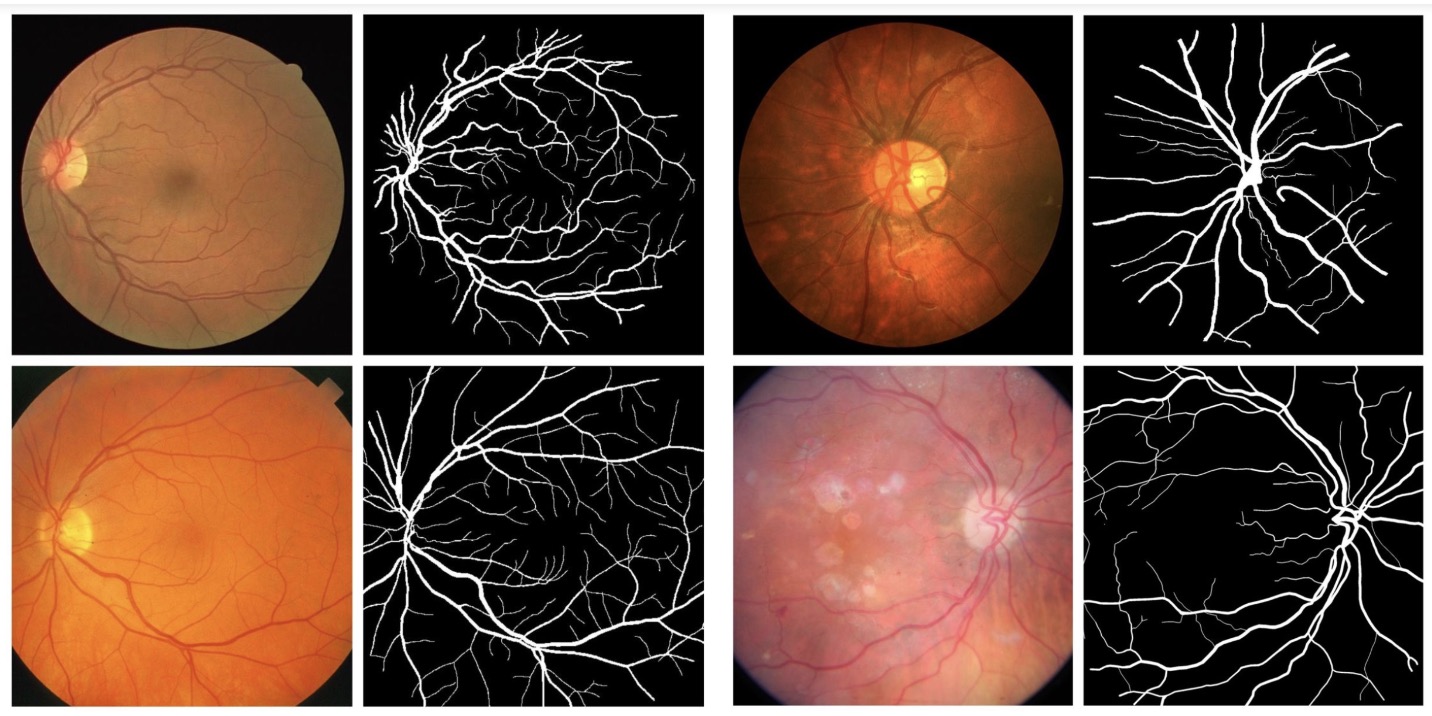}
  \caption{Upper left is the mild diabetic retinopathy retina image from DRIVE dataset \cite{staal2004ridge}. The upper right shows the retinal image of multi-ethnic children from the CHASE-DB1 dataset \cite{owen2009measuring}. Lower left is mild diabetic retinopathy retina image from STARE dataset \cite{hoover2000locating}. The glaucoma retina image from the DR-Hagis dataset \cite{holm2017dr} is the lower right.}
  \label{fig:retina_data}
\end{figure}

This thesis investigates current state-of-the-art vessel segmentation methods, aims to elucidate their methodologies and examines their advantages and limitations. This study is crucial as it provides a comprehensive overview of innovative deep learning-based retinal blood vessel segmentation techniques from 2014 to 2023, fostering an understanding of AI technologies. While existing deep learning-based approaches have achieved remarkable results in medical image segmentation, there has been a dramatic increase in computational complexity, ranging from pure convolutional neural networks (CNNs \cite{lecun2015deep}) to pure Vision Transformers (ViTs \cite{dosovitskiy2020image}). Moreover, in the medical vision domain, manual labelling depends on human experts with solid background knowledge in distinguishing the vessel object. Therefore, data limitation becomes an urgent problem to deal with. The popular vessel segmentation datasets, illustrated in Figure \ref{fig:retina_data}, typically comprise around 40 images. This small sample size makes it challenging to extract capillaries \cite{wu2020nfn+}, primarily as deep learning methods rely on larger datasets to learn robust features.

\begin{figure}[h!]
  \centering
  \includegraphics[width = \textwidth]{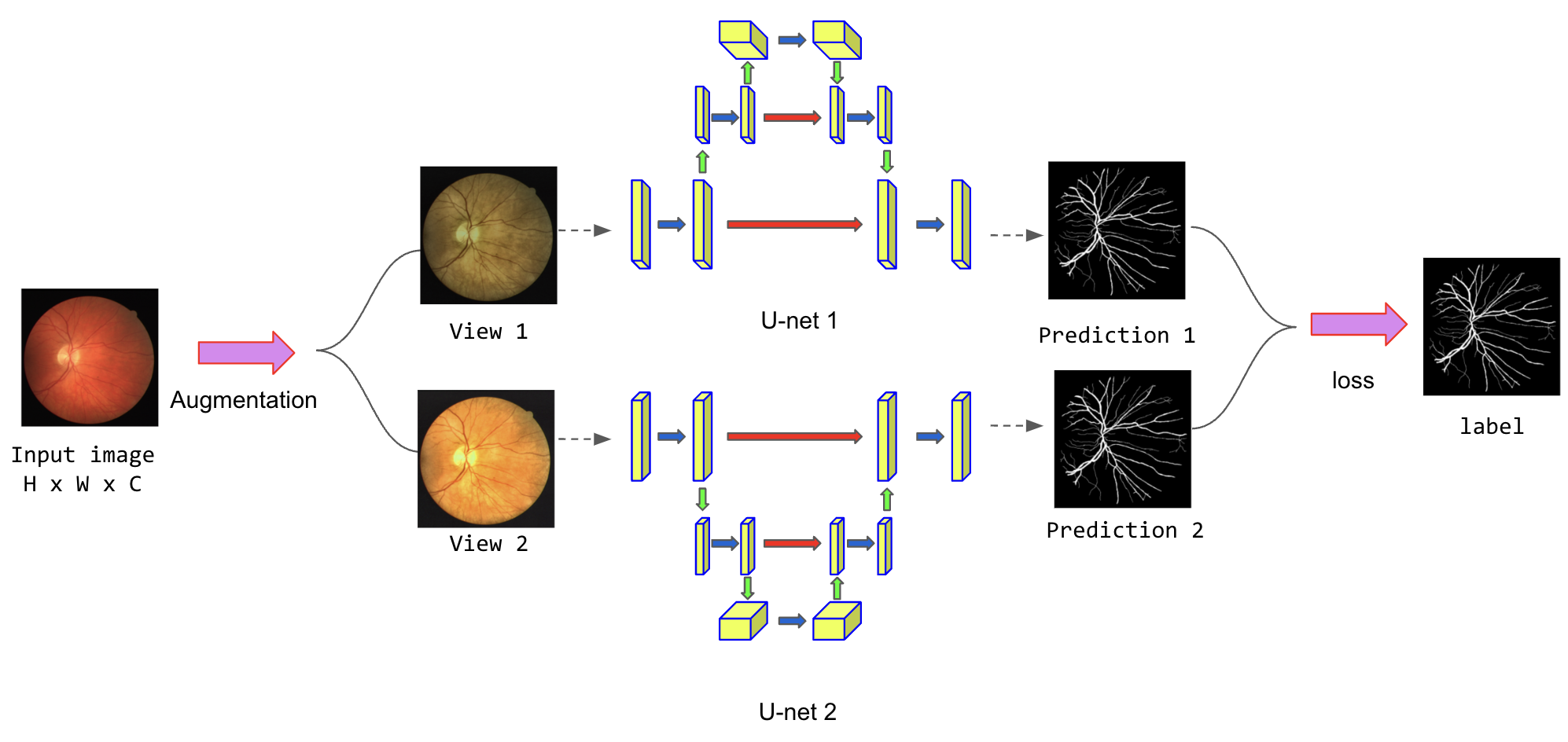}
  \caption{Structure of the AUtO-Net: \textbf{A}ugmentation-driven doubled \textbf{U}Ne\textbf{t} \textbf{O}-shape Network.  extracts robust features from multiple augmented views through the contrastive learning framework.}
  \label{fig:auto}
\end{figure}

To address the prevalent issues of limited data and high computational cost, this study rigorously tests various data augmentation methods, such as Contrastive Limited Adaptive Histogram Equalisation (CLAHE \cite{reza2004realization}) and the proposed method known as MixUp, which enriches data distribution and mitigates overfitting by randomly mixing training images or labels, aims to enhance the model's generative capabilities. To further promote learning robust invariant vessel structures across multiple augmented views, a novel learning framework is introduced by integrating data augmentation and contrastive learning \cite{he2020momentum, chen2020exploring, chen2020simple, grill2020bootstrap, caron2021emerging, caron2021unsupervised} to learn the comprehensive feature representation. Since conventional contrastive learning is the self-supervised approach. However, the proposed method is supervised learning with multiple views. Therefore, we name the modified contrastive learning framework as the contrastive multiview learning framework.  Moreover, the proposed method utilises a hybrid network structure, incorporating an attention mechanism into the CNN blocks. This balances performance enhancement and computational efficiency and allows the model to highlight and learn from global feature representations. The proposed method, referred to as AUtO-Net, can be trained on a grey image in less than 30 minutes and achieves high performance on benchmark datasets. Figure \ref{fig:auto} provides an overview of the AUtO-Net, and the details will be explained in Chapter \ref{Chapter3}.


\clearpage
\section{Thesis Structure}
The thesis is organized as follows: 

Chapter \ref{Chapter2} presents a comprehensive literature review, initially discussing rule-based medical image segmentation methods, followed by categorizing the current state-of-the-art (SOTA) deep neural networks for vessel segmentation into five classes: pure CNNs, hybrid models, pure Vision Transformers, GAN-based, and knowledge distillation-based methods.

Subsequently, Chapter \ref{Chapter3} introduces the motivation, intuition, and methodology of three modules of the proposed AUtO-Net: contrastive multiview learning framework, hybrid network structure and MixUp data augmentation.

Chapter \ref{Chapter4} showcases the experimental results of the AUtO-Net tested on two popular datasets, compared with the current benchmark method with quantitative analysis. And ablation studies for testing the effectiveness of the different modules. 

Lastly, Chapter \ref{Chapter5} summarizes the thesis, discusses the limitations and improvements and provides insights for future research direction.

\section{Contributions}
This work offers five major contributions:

\begin{itemize}
    \item \textbf{Propose of multiview learning framework} The introduction of a lightweight contrastive multiview learning framework integrated with data augmentation methods to simultaneously learn multiple views without extra training parameters where alleviates the common problems of insufficient data and limited computing resources in the field of medical image segmentation. Moreover, the ablation study shows that the contrastive multiview learning framework could improve the performance of convolutional neural networks and attention-based work. Therefore, it is a general idea rather than a specific technic, and the implementation is clean and neat, which could be suitable for any machine-learning task.

    \item \textbf{Propose of hybrid network structure} Incorporating CNN backbones with attention blocks. The performance is improved by compensating for CNN's lack of ability to capture long-range features. At the same time, maintaining the computational efficiency of training. Justified by controlled experiments, the attention mechanism indeed increases the performance sensitivity, and the model could capture detailed vessel features.
    
    \item \textbf{Comprehensive literature review} A comprehensive literature review provides a detailed explanation of existing methodologies and an in-depth quantitative analysis to better understand novel AI technology applications in medical image segmentation.

    \item \textbf{Examination various data augmentations} Data augmentations are proven to be effective in improving the model's robustness. This work explores an extensive number of augmentations to validate their performance with deep analysis.

    \item \textbf{Propose of MixUp augmentation} A novel data augmentation method called MixUp fusing information from multiple images into a single image not only allows the model to learn information from multiple images at the same time, thus saving training time and increasing vessel variation, thus allowing the model to learn key features better enhances the model's generalization ability.
\end{itemize}

\chapter{Literature Review}
\label{Chapter2}

\section{Rule-based Segmentation Algorithms}
Rule-based algorithms are unsupervised approaches that efficiently segment regions of interest. However, they often require manual parameter tuning to perform well and can be noise-sensitive. Specifically, automatic thresholding is a segmentation method based on an image's grey-scale histogram \cite{otsu1979threshold}.

The primary concept is to differentiate pixels in various regions of an image by automatically selecting a threshold. Pixels greater than or equal to the threshold are grouped into one category (e.g., the target region). In comparison, pixels smaller than the threshold are grouped into another category (e.g., the background region). Specifically, $f(x, y)$ is defined by the input image's histogram, and $T$ represents the pre-defined threshold. The segmented image $g(x, y)$ is then defined by the following criteria:

\begin{equation}
g(x,y) = \begin{cases}
1, & f(x, y) > T, \\ 0, & f(x, y) \leq T.
\end{cases}
\label{eq: relu_d}
\end{equation}

\begin{figure}[h!]
  \centering
  \includegraphics[width=\textwidth]{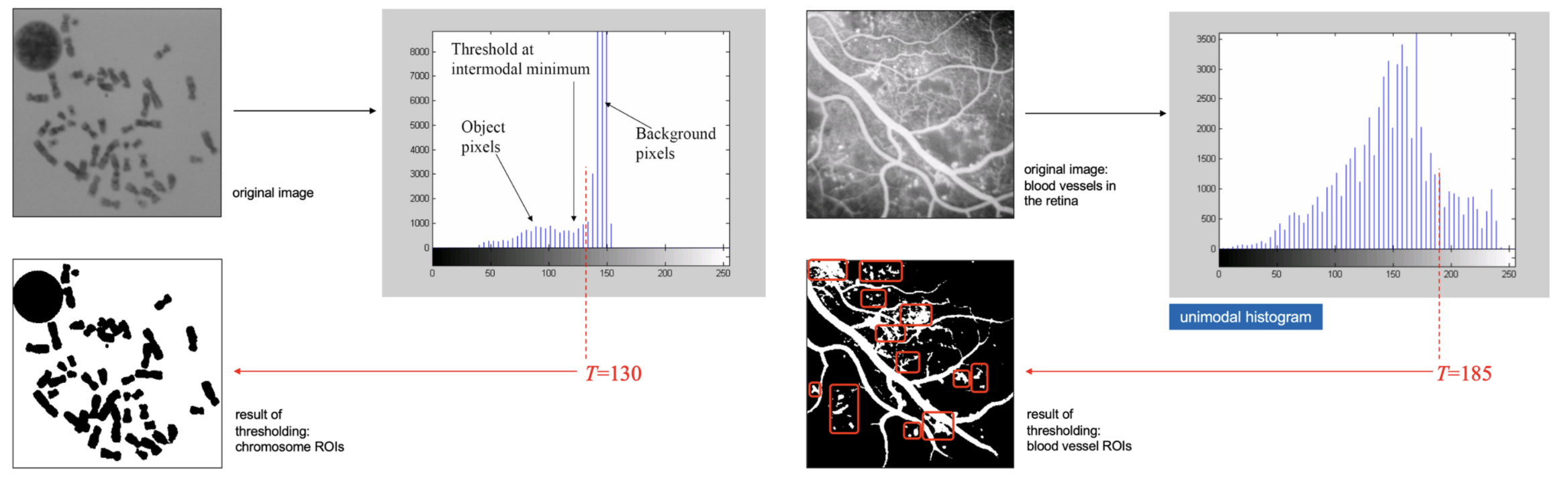}
  \caption{Left image shows that the threshold-base algorithm could clearly segment the chromosome from the background with a clear distinction between the chromosome and background in the histogram. However, the algorithm fails for vessel segmentation when the background is noisy, so the histogram no longer provides separation heuristics. The red box annotates the misclassified vessel objects  \cite{itbslides}.}
  \label{fig:threshold}
\end{figure}

The challenge of these rule-based algorithms lies in determining appropriate parameters, and performance depends on single means of information such as histogram distribution or pixel value of nearest neighbourhoods \cite{295913}. The manual design feature could have a limited use case in real-world applications. Specifically, the threshold-based algorithm only performs well when a distinct boundary exists between the background and objects. However, complex images may have unclear decision boundaries for threshold selection. Figure \ref{fig:threshold} visually illustrates the success and failure cases of the threshold-based algorithm.

\section{Deep Learning Background}
\begin{figure}[h!]
  \centering
  \includegraphics[width=\textwidth]{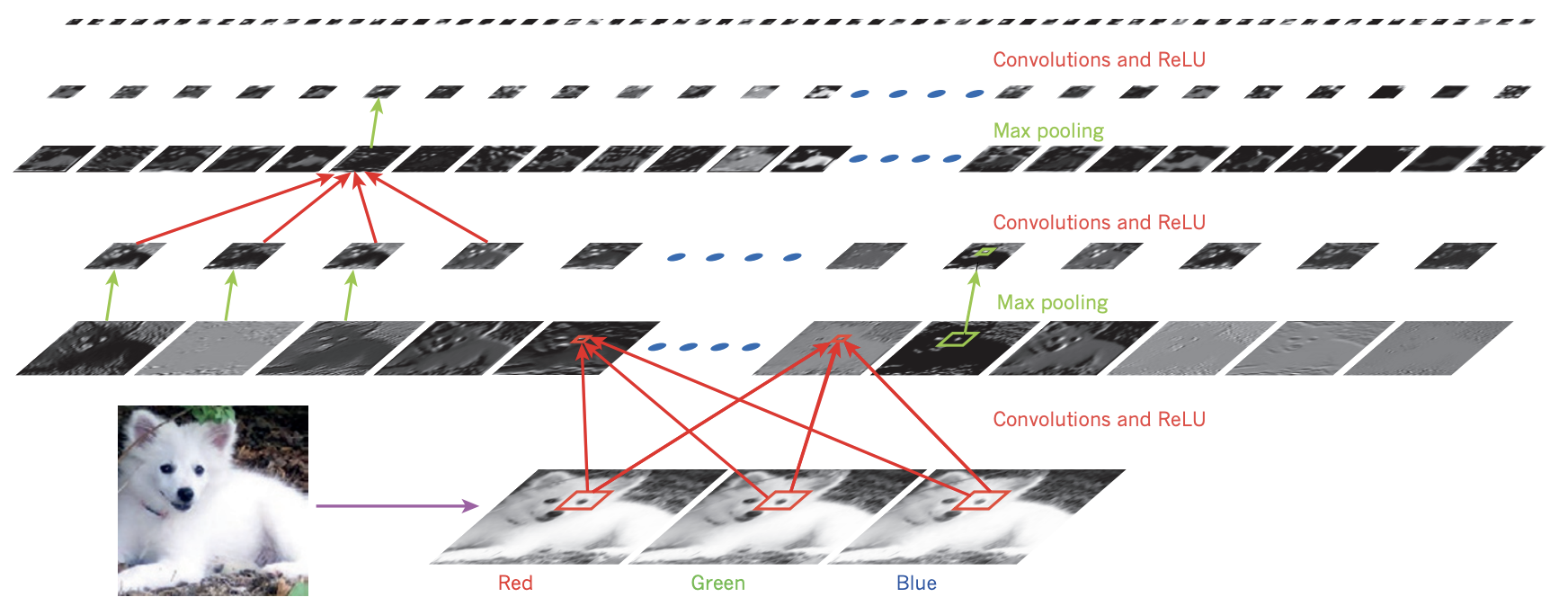}
  \caption{Structure of a convolutional neural network. Multiple kernels (filters) automatically extract the image features and feed the feature representation to the next layer. Through the depth of layers, more abstract features could be learned. The deep neural network consists of multiple modules. Max pooling operation down samples the image to increase the model's receptive field, letting the model learn global features, and the ReLU is the activation function to add non-linearities to make have more complex decision boundaries \cite{lecun2015deep}.}
  \label{fig:cnn}
\end{figure}

Compared to rule-based algorithms, deep neural networks automatically recognize and extract essential patterns and structures of target tissues or organs through the deep hidden layers, eliminating the need for manually setting hyperparameters \cite{lecun2015deep}. Additionally, they exhibit robustness to variations in noise, resulting in more reliable segmentation outcomes. Furthermore, deep learning models can effectively handle multi-modal medical imaging data, making them suitable for various medical image segmentation applications \cite{litjens2017survey}. Figure \ref{fig:cnn} shows the architecture of the convolutional neural network.

\subsection{Loss Functions}
Loss functions are crucial in deep learning architectures, often called objective functions. They are designed to quantify the discrepancy between the estimated value and the ground truth. In conjunction with optimization algorithms, loss functions help models minimize errors throughout training. In the context of medical image segmentation, the pixel value of the organ and background object belongs to different classes. Therefore, the loss function measures the misclassified pixels.

\BE{Cross-Entropy Loss} CCE Loss function, as expressed in Eq.(\ref{eq:cce}), is the summation of the Entropy of distribution $y$ and the Kullback-Leibler (KL) divergence of the ground truth $y$ and predicted $\hat{y}$ distributions. Entropy measures the uncertainty associated with a random variable, while KL divergence \cite{joyce2011kullback} evaluates the dissimilarity between two distributions. From an information theory standpoint, it calculates the information loss in approximating distribution $y$ given $\hat{y}$. Consequently, the CCE Loss function simultaneously minimizes the randomness of the ground truth. It enforces the predicted distribution to converge towards the ground truth distribution stated in Eq.(\ref{eq:cce}):

\begin{equation}
\begin{aligned}
\mathcal{L} & = -\sum_{i} y_i \log \hat{y}_i \\ 
&= -\sum_{i} y_i \log y_i + \sum_{i} y_i \log y_i - \sum_{i} y_i \log \hat{y}_i \\
&= -\sum_{i} y_i \log y_i + \sum_{i} y_i \log \frac{y_i}{\hat{y}_i}\\
&= Entropy(y) + D_{KL}(y_i|\hat{y}_i),
\label{eq:cce}
\end{aligned}
\end{equation}
where $y_{i}$ represents the ground truth distribution, and $\hat{y}_i$ denotes the predicted distribution.

\BE{Weighted Cross-Entropy Loss} Imbalanced class distribution is a prevalent issue in medical image segmentation. The number of categories can exhibit significant disparities, leading to suboptimal model performance when predicting underrepresented classes \cite{phan2020resolving}. Weighted Cross-Entropy Loss (WCCE) mitigates this problem by assigning higher weights to underrepresented classes, emphasizing their importance. WCCE is formulated in Eq.(\ref{eq:wcce}):

\begin{equation}
\mathcal{L} = - \sum_{i=1}^{C} w_i \cdot y_i \cdot log(\hat{y}_i),
\label{eq:wcce}
\end{equation}
where $C$ denotes the number of categories, and $w_i$ represents the weight of the $i$-th category.

\BE{Focal Loss} Focal Loss \cite{lin2017focal} addresses class imbalance by reducing the error weight for easily classified samples and increasing the weight for those near the decision boundary, compelling the model to concentrate on these instances, ultimately enhancing its performance. For binary classification tasks, the Focal Loss is defined as follows:

\begin{equation}
\mathcal{L} = -\alpha_t(1 - p_t)^{\gamma} \log(p_t)
\label{eq:focalloss},
\end{equation}
here, $\alpha_t$ and $\gamma$ are hyperparameters, while $p_t$ represents the probability estimated by the model, indicating the confidence level for the prediction. When $p_t$ approaches 1, the sample is deemed easily classified. Conversely, when the sample is challenging to classify, $p_t$ tends towards 0, causing $(1 - p_t)^{\gamma}$ to approach 1 and subsequently increase the loss for such samples. Focal Loss is initially developed for object detection tasks and is commonly employed in medical image segmentation tasks, particularly when segmenting small tissue structures such as blood vessel details.

\BE{Dice Loss} Dice Loss, based on the Sørensen-Dice index or F1 score, is a popular choice for medical image segmentation. Its formula is defined as follows:

\begin{equation}
\mathcal{L} = 1 - \frac{2 \sum_{i}^{N} p_i g_i}{\sum_{i}^{N} p_i^2 + \sum_{i}^{N} g_i^2}
\label{eq:dice_loss},
\end{equation}
where $p_i$ denotes the predicted value, $g_i$ is the ground truth value, and $N$ represents the number of samples. The Dice index measures the similarity between two samples, ranging from 0 to 1, where 0 indicates no match between the prediction and ground truth. Dice Loss is often combined with other loss functions, such as Cross-Entropy Loss.

\section{Benchmark Segmentation Methodologies}
\subsection{Fully Convolutional Networks}
\begin{figure}[h!]
  \centering
  \includegraphics[width = \textwidth]{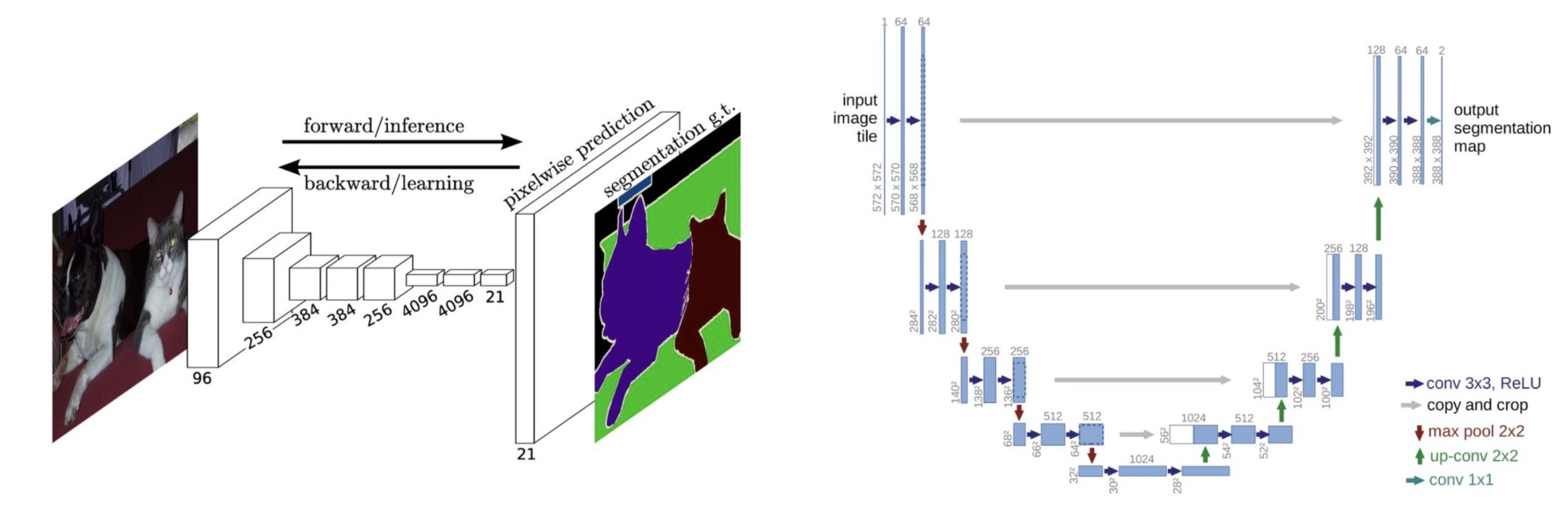}
  \caption{Left image shows the structure of Fully Connected Network \cite{long2015fully}. The right image shows the structure of UNet \cite{ronneberger2015u}.}
  \label{fig:fcn_unet}
\end{figure}

\BE{FCN} The Fully Convolutional Network (FCN \cite{long2015fully}) replaces the fully connected layers in traditional convolutional layers with up-sampling layers to make pixel-wise predictions. Figure \ref{fig:fcn_unet} illustrates the structure of FCN, which consists of downsampling and upsampling components. The downsampling part, the encoder, extracts features and reduces the size through pooling operations. The feature map is then passed to the upsampling component, called the decoder, to produce the pixel-wise dense prediction. FCN introduced skip connections between encoders and decoders to better utilise feature information for more accurate detail predictions.

\BE{UNet} Furthermore, UNet is specifically designed for medical image segmentation. Like FCN, UNet also adopts an encoder-decoder structure with skip connections and a fully convolutional architecture. However, UNet is optimised for medical image segmentation tasks. In the medical imaging domain, localisation refers to the ability to precisely detect the boundaries of the objects of interest to acquire perfect overlapping, which is crucial for accurate segmentation. UNet achieves this by integrating high-resolution features with upsampling layers through skip connections between the symmetric encoder-decoder architecture. As a result, UNet has become a popular choice for a wide range of medical imaging applications.

\BE{UNet Family} The UNet family has significantly expanded since the initial success of UNet, giving rise to numerous innovative network architectures. The Residual Recurrent UNet (R2U-Net \cite{alom2018recurrent}) incorporates residual and recurrent blocks \cite{he2015deep} into the UNet structure. The residual block, proposed to address the vanishing gradient problem encountered in deep networks, employs skip connections between layers to stabilize training and enhance performance. Recurrent modules \cite{medsker2001recurrent} enhance the network's ability to capture local and global contextual information. R2U-Net maintains the same architecture as the original UNet, achieving benchmark results in the blood vessel, skin cancer, and lung lesion segmentation tasks.

UNet++ \cite{zhou2018unet++} employs a deeply-supervised mechanism that introduces additional skip connections between the encoder and decoder, aiming to bridge the semantic gap. This technique improves the fusion of low-level and high-level features, resulting in superior performance compared to the vanilla UNet.

Similarly, the Bi-directional O-shape Network (BiO-Net \cite{xiang2020bio}) recurrently reuses network blocks. Unlike UNet++, which relies on feature fusion through deep supervision, BiO-Net does not increase the number of parameters. BiO-Net can better learn the semantic information necessary for segmentation tasks by leveraging forward and backward skip connections between encoders and decoders.

Vessel-Net \cite{wu2019vessel} is a lightweight U-shaped structure incorporating a redesigned inception-residual block, combining the benefits of inception and residual blocks. Vessel-Net is trained using four supervision paths to better learn medical features.

The Fully-Resolution Network (FR-UNet \cite{liu2022full}), inspired by HR-Net \cite{pohlen2017full} and UNet++ \cite{zhou2018unet++}, diverges from traditional convolutional network structures, which gradually downsample feature maps. Instead, HR-Net maintains high-resolution features in parallel throughout the network, enabling feature fusion across different scales. Consequently, HR-Net can capture both local and global features. FR-UNet features a redesigned feature aggregation module and employs dense skip connections for deep supervision. FR-UNet currently holds the state-of-the-art position on the DRIVE dataset.

\subsection{Vision Transformers}
The attention mechanism \cite{vaswani2017attention} in Natural Language Processing was initially utilized as an alternative solution to recurrent neural networks \cite{medsker2001recurrent}, and its parallel computing led to the success of large language models (LLMs). 


Specifically, the attention mechanism is a technique designed to capture relationships between elements within an input sequence. Its core concept involves computing a weight distribution for each element in the input sequence. In contrast to traditional RNNs, self-attention can process the entire input sequence in parallel without the need to traverse the elements sequentially. Previous studies \cite{dosovitskiy2020image} have integrated pure Transformers and attention blocks into the field of computer vision, achieving state-of-the-art (SOTA) performance.

\begin{figure}[h!]
  \centering
  \includegraphics[scale=0.6]{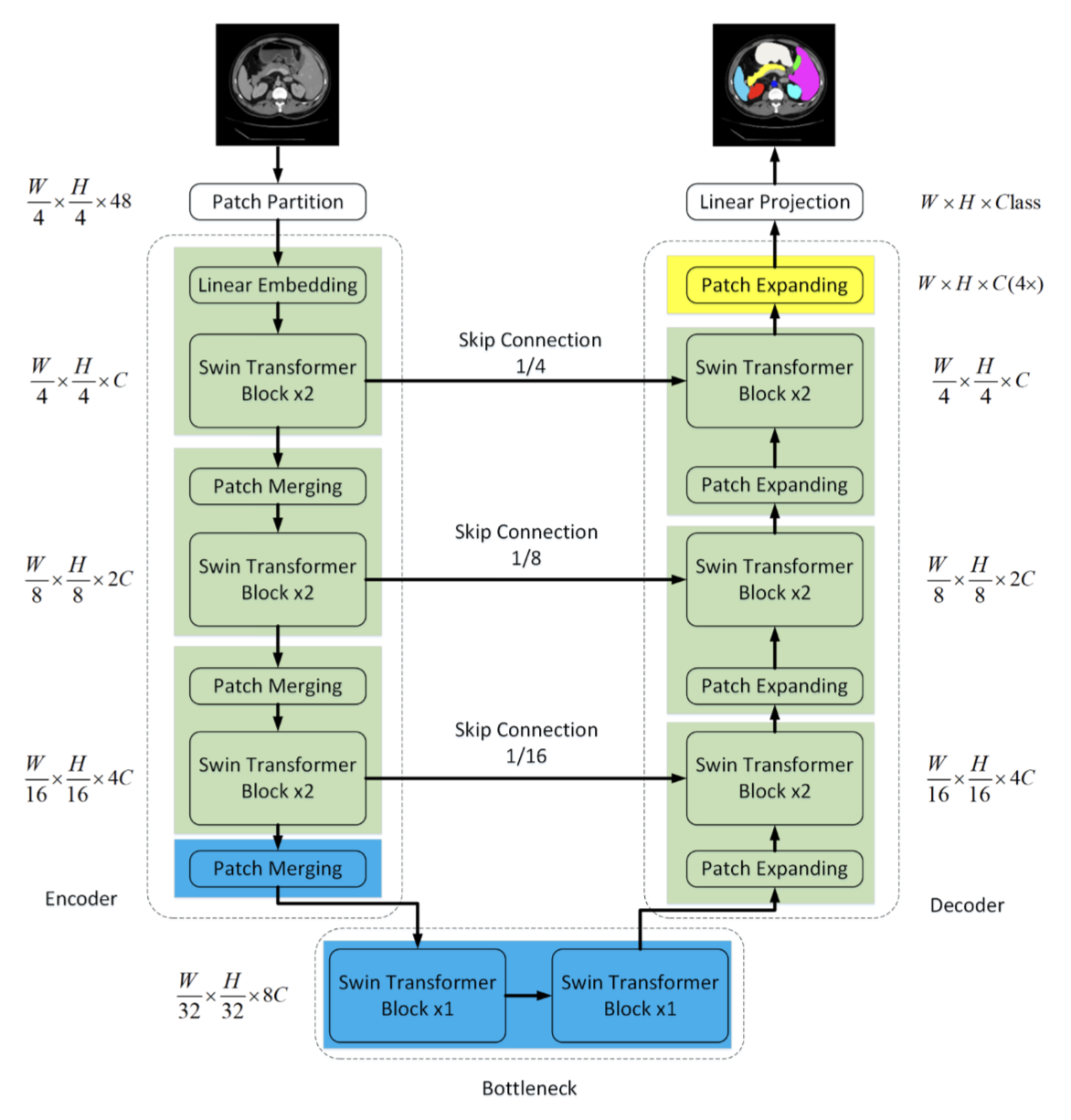}
  \caption{The structure of Swin-UNet, where Swin Transformer blocks replace the CNN blocks \cite{cao2023swin}. The skip connections help better fuse the feature between encoders and decoders.}
  \label{fig:swinunet}
\end{figure}

\BE{Swin-UNet} The medical image processing domain has also witnessed the adoption of Transformers as the backbone instead of CNNs, yielding impressive results. Swin-UNet \cite{cao2023swin} is proposed for organ segmentation, utilizing the Swin Transformer \cite{liu2021swin} as the backbone network block illustrated in Figure \ref{fig:swinunet}.

\begin{figure}[h!]
  \centering
  \includegraphics[scale=0.3]{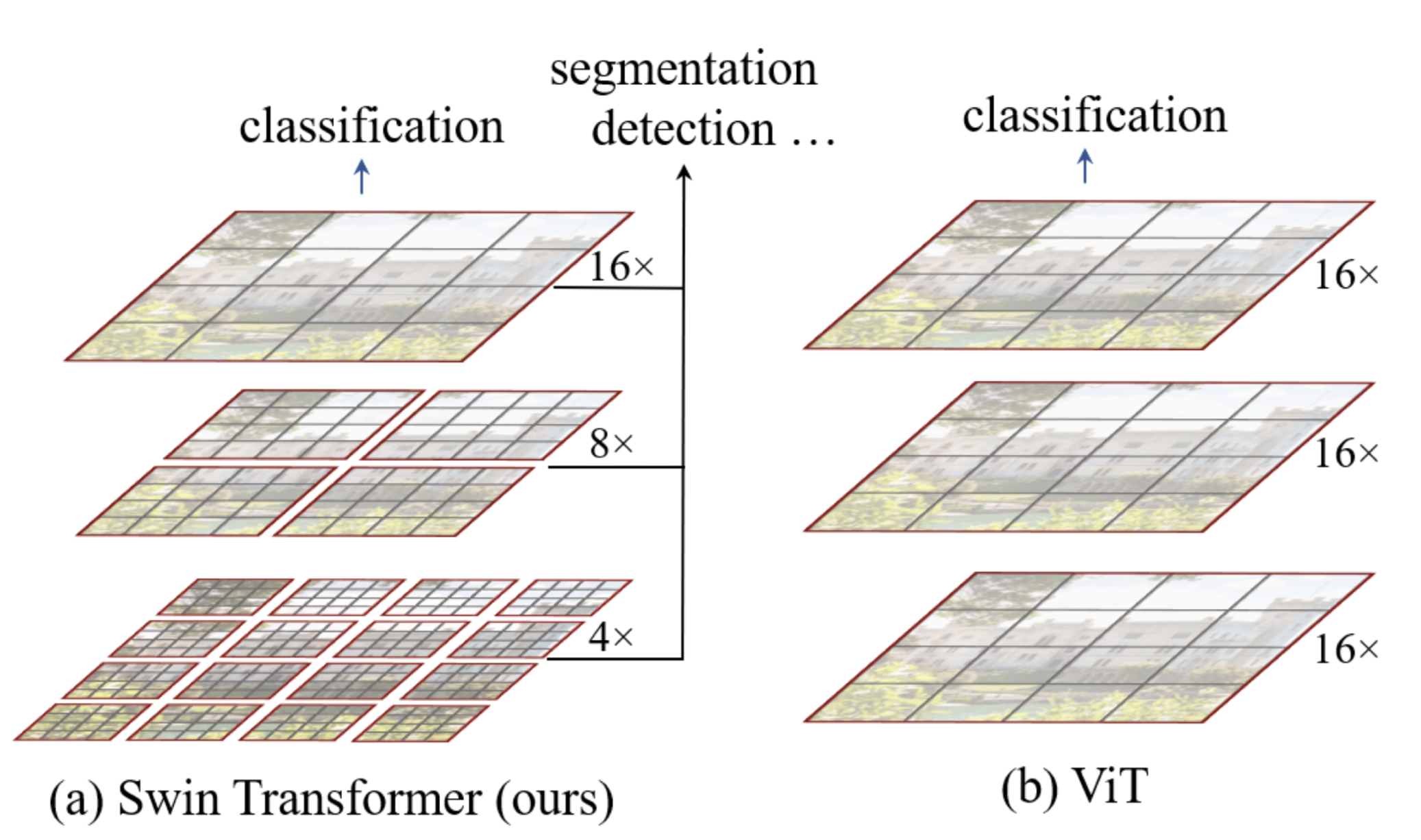}
  \caption{The comparison of the hierarchical structure of Swin Transformer and Vision Transformer. The self-attention is calculated only with small windows in Swin Transformer to utilise the computational resource better \cite{liu2021swin}.}
  \label{fig:swin}
\end{figure}

The Swin Transformer, as demonstrated in Figure \ref{fig:swin}, emulates the hierarchical architecture and localisation capability of CNNs to better exploit the inductive bias of vision tasks. Swin stands for shifting window, a design intended to incorporate the inductive bias of CNNs into Transformers. While the original ViT computes self-attention across the entire image, which is computationally expensive, the Swin Transformer calculates attention maps across windows, reducing computational complexity. This parallel approach addresses the challenges of training high-resolution images, making it particularly suitable for medical image backgrounds where image sizes are typically high resolution.

\subsection{Hybrid Networks}
Nonetheless, the complexity of the model is contingent on the size of the training data. Scaling the parameters of Transformers can lead to heightened computational demands and a predisposition to overfitting. Consequently, hybrid networks are the alternative that combines the advantages of CNNs and Vision Transformers that have been proposed for vessel segmentation.

\begin{figure}[h!]
\centering
\includegraphics[width= 0.8\textwidth]{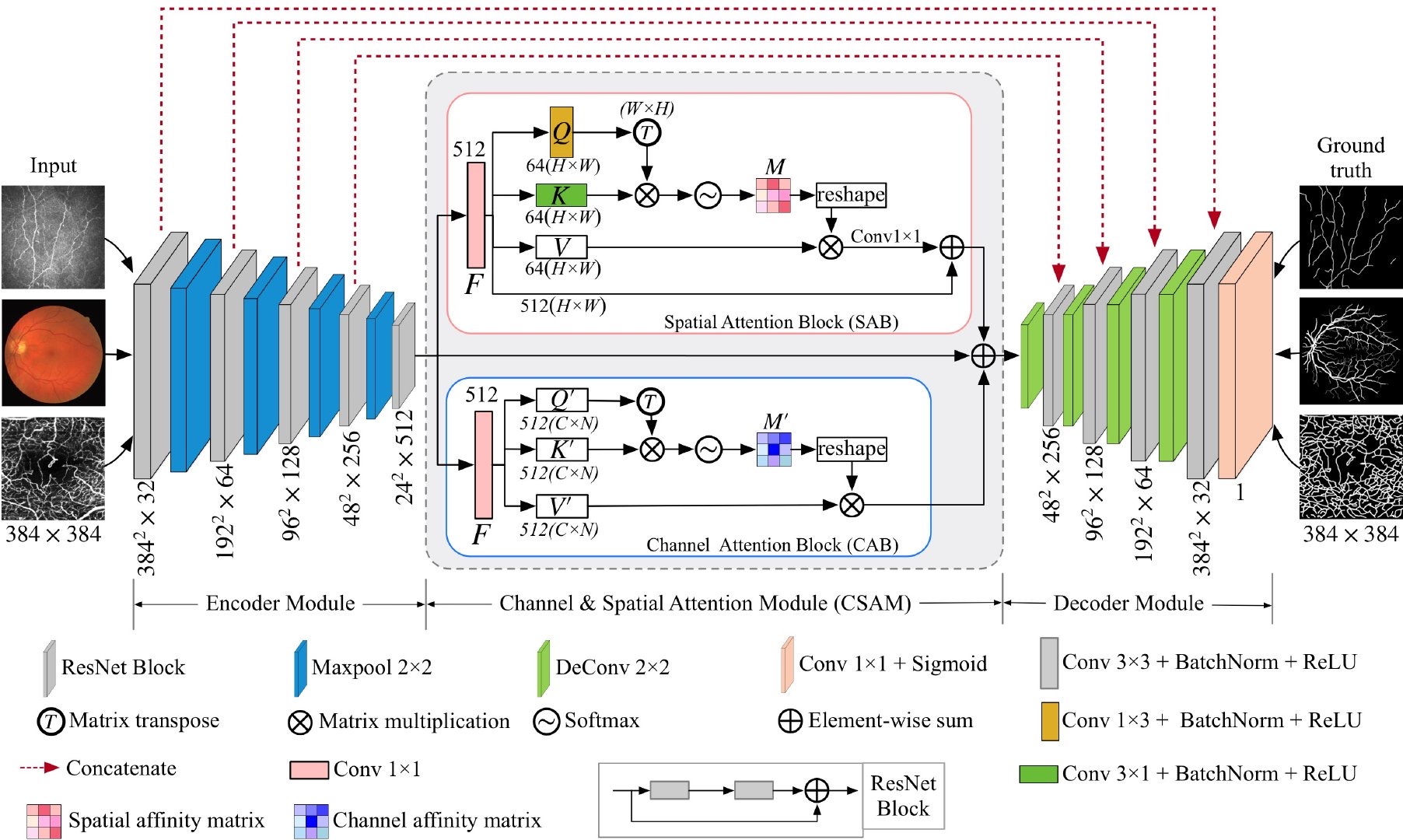}
\caption{The CS-Net comprises three parts: encoder, attention module, and decoder \cite{mou2019cs}.}
\label{fig:csnet}
\end{figure}

\BE{CS-Net (Dual Attention)} CS-Net \cite{mou2019cs} incorporates spatial and channel attention to detect the global features of curvilinear structures in retinal images. Its backbone network is based on UNet, with the network structure depicted in Figure \ref{fig:csnet}.

Dual attention \cite{fu2019dual} was initially proposed for scene segmentation, employing attention mechanisms in both spatial and channel dimensions to capture global semantic context information. Specifically, the spatial attention mechanism learns the dependencies between each position of the input feature map and other positions, capturing long-range dependencies in an image. This enables the model to concentrate on the image's local details and global contextual information. Similarly, the channel attention mechanism focuses on the inter-channel relationships of the input feature maps. By learning the dependencies between different channels, the model can better capture the semantic information of each channel, facilitating improved identification and segmentation of various objects and scene elements. The formulas for spatial and channel attention are as follows:

\begin{equation}
S_{(x, y)} = \dfrac{exp(K_{(x)} \cdot Q_{(y)}^T)}{\sum^N_{x = 1} exp(K_{(x)} \cdot Q_{(y)}^T)},
\end{equation}
where $K$ and $Q$ are two views generated by $1 \times 1$ convolutional layers demonstrated. And the channel attention is by the following:

\begin{equation}
C_{(x, y)} = \dfrac{exp(F_{(x)} \cdot F_{(y)}^T)}{\sum_{x =1 }^C exp(F_{(x)} \cdot F_{(y)}^T)}.
\end{equation}

\BE{Other Hybrid Models} Attention-UNet \cite{oktay2018attention} incorporates the attention gate into the skip connections between the encoders and decoders to emphasize detailed features better and enhance performance sensitivity. More specifically, the attention gate reassigns weights during the feature fusion process, aiming to increase attention to essential regions and thus bolstering segmentation performance.

Spatial Attention UNet (SA-UNet \cite{guo2021sa}) introduces a spatial attention block. Like other attention mechanisms, the primary advantage is spatially accentuating the weight of features, selectively focusing on detail-rich regions.

\subsection{Generative Adversarial Netwroks}
Generative adversarial networks (GAN \cite{goodfellow2014generative}) were first introduced for unsupervised learning, composed of two parts the generator and the discriminator. The optimisation of GAN is during the minimax advertised game. The generator will generate a fake image to cheat the discriminator. In contrast, the discriminator is aimed to increase the ability to discover whether the image is from the actual distribution or generated by the generator. Therefore, they are iteratively optimised until they converge, finally, with the generator striving to create more realistic images and the discriminator aiming to better differentiate between natural and synthetic images. This adversarial process generates high-quality synthetic images that resemble authentic images like data augmentation.

\begin{figure}[h!]
\centering
\includegraphics[width= 0.9 \textwidth]{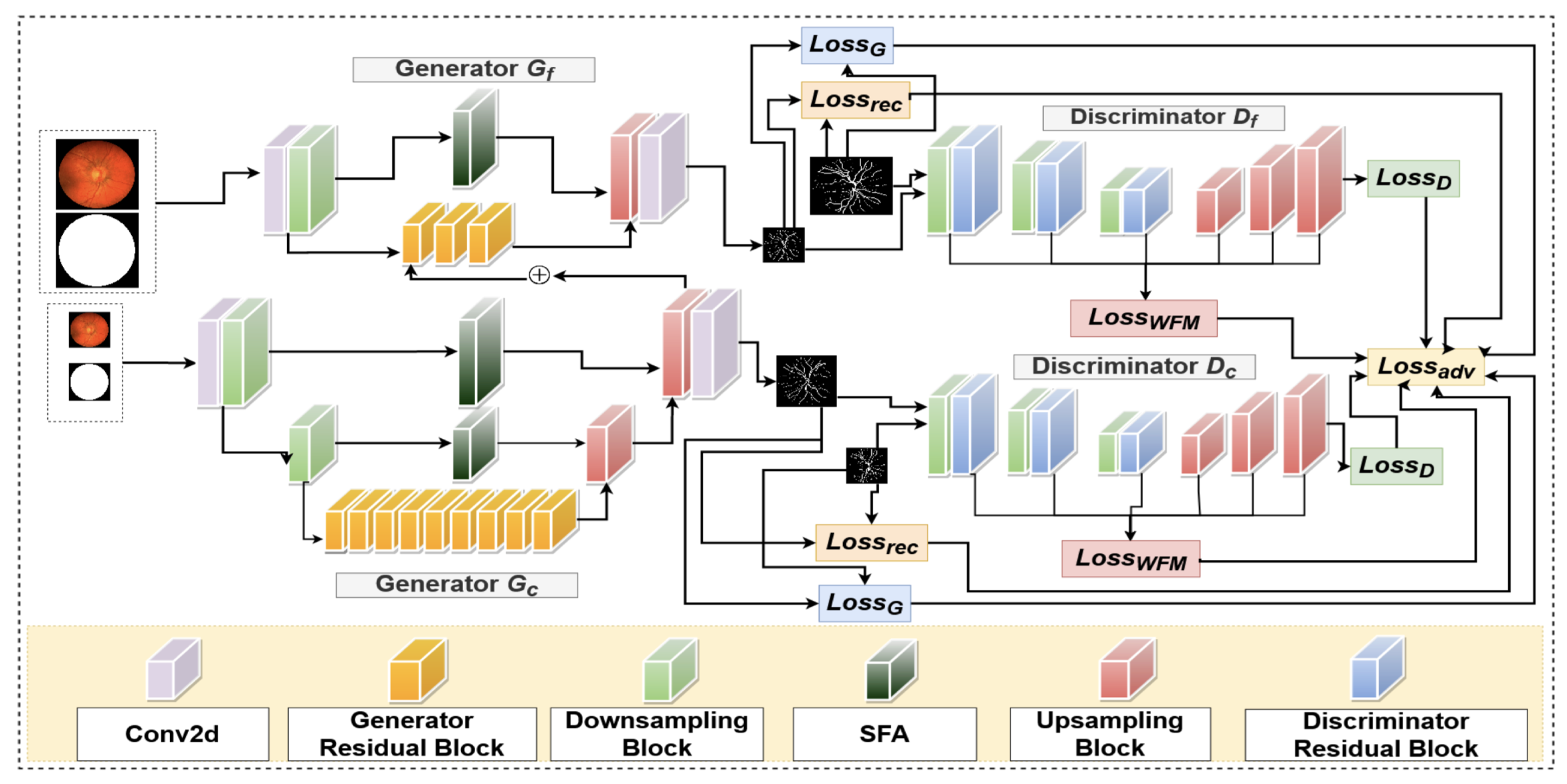}
\caption{Network structure of RV-GAN is composed of two generators and two discriminators \cite{kamran2021rv}.}
\label{fig:rvgan}
\end{figure}

\BE{RV-GAN} The GAN approach could concur with the limitation of training data by training on synthesis pseudo-labels. RV-GAN \cite{kamran2021rv} utilises this idea in vessel segmentation. The network structure is shown in Figure \ref{fig:rvgan}. The main idea of RV-GAN is to exploit the multi-scale information of input images by integrating two generators and two discriminators to learn the detailed vessel structure. However, GAN-based networks are notoriously difficult to optimize, as they require maintaining consistency between the generator and the discriminator. This stems from the need to balance the competition between the generator and the discriminator during training. If this balance is disrupted, the training process may become unstable or even experience mode collapse. For instance, if the generator significantly outperforms the discriminator, the discriminator cannot provide an adequate supervisory signal. This may lead to the generator overfitting the training data and generating images lacking diversity. Additionally, the gradient information provided by the discriminator may become weak, resulting in insufficient guidance for the generator during training. Conversely, if the discriminator is stronger than the generator, it can quickly discern the differences between authentic and generated images. At this point, the generator will struggle to deceive the discriminator, causing issues with its gradient updates. This may slow down the training process, resulting in poor-quality images produced by the generator and even leading to pattern collapse, where the generator can only produce a minimal number of image samples. To address this issue, the learning rate of the discriminator can be reduced appropriately, or regularization methods (e.g., gradient penalty) can be employed to limit the discriminator's capability. The RV-GAN specially designed a modified loss function to balance the inconsistency between the generator and the discriminator and finally achieved impressive segmentation results.

\subsection{Knowledge Distillations}
The original concept of knowledge distillation aims to train a student network with fewer parameters to approximate the predictions of a larger model, also known as the teacher network, to achieve high computational efficiency.

\begin{figure}[h!]
\centering
\includegraphics[width=0.9\textwidth]{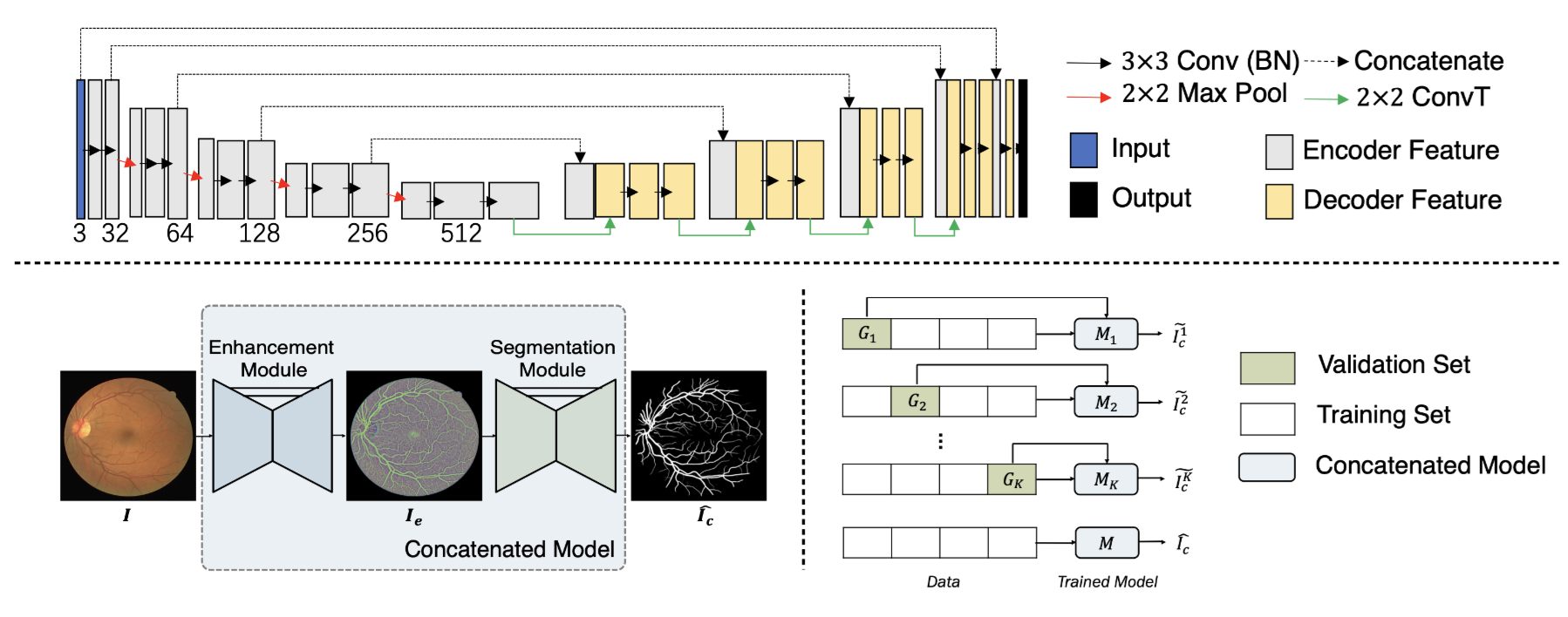}
\caption{Network structure of SGL. The training set is divided into $k$ subsets, fed to the model set {$M_k$}, and updated by the sum of the loss as the supervision signal \cite{zhou2021study}.}
\label{fig:sgl}
\end{figure}

\BE{SGL} Training on a small dataset, particularly medical images, carries a high risk of overfitting. Furthermore, image segmentation labels require manual annotation by ophthalmologists, which can be error-prone and lead to noisy labelling problems. Study Group Learning (SGL \cite{zhou2021study}) is proposed to address these issues, drawing inspiration from knowledge distillation cross-validation and treating the ground truth as noisy labels. However, in SGL, models are trained in parallel using K-fold cross-validation and divided into different $k$ training models, denoted as the model set {$M_k$}. This approach seeks to enhance the generalization capability and robustness of the models. The pseudo-label is denoted as $\tilde{I}_{ck}$, and the ground truth is represented by $I_c$. The pseudo-label set is denoted by $\tilde{I_c} = \bigcup^K_{k = 1} \tilde{I}_{ck}$. The final loss function is the sum of two parts. The first part is the loss of prediction $\hat{I}_c$ with the actual ground truth $I_c$ and pseudo label $\tilde{I_c}$:

\begin{equation}
\mathcal{L}_{SGL} = \mathcal{L}(\hat{I}_c, I_c) + \mathcal{L}(\hat{I}_c, \tilde{I}_c).
\end{equation}
The model $M$ trained on the entire dataset is expected to have higher prediction accuracy and consolidate knowledge from different study groups, ultimately achieving enhanced performance and robustness. The network structure of SGL is depicted in Figure \ref{fig:sgl}.

\section{Contrastive Learning}
Contrastive learning is a versatile self-supervised technique that facilitates learning feature representations without the need for labelled data. Specifically, previous studies \cite{wu2018unsupervised, ye2019unsupervised} posited that analogous features from a single instance could be mapped into the latent feature space with close proximity. In contrast, disparate features from different instances should be distinctly separated. Consequently, the objective function in contrastive learning frameworks seeks to maximize the similarity with positive samples, defined as instances exhibiting invariant features, while minimizing the similarity with the irrelevant features of negative samples. The evolution of contrastive learning frameworks can be categorized into two parallel trajectories: refining the storage of positive and negative examples and streamlining the network structure.

\BE{Strategies for positive and negative samples} Some contrastive learning approaches \cite{wu2018unsupervised, ye2019unsupervised} employ other instances within the mini-batch as negative samples. However, this approach inadvertently escalates computational costs. SwAV \cite{caron2020unsupervised} integrates the principles of deep clustering, selecting cluster centroids as negative samples, and training the model by swapping predictions using various data augmentation methods. Conversely, BYOL \cite{grill2020bootstrap} maps images from different views into the same latent feature space, utilizing information from one view to predicting another, thereby enabling training without negative samples.

\begin{figure}[h!]
\centering
\includegraphics[width=0.8\textwidth]{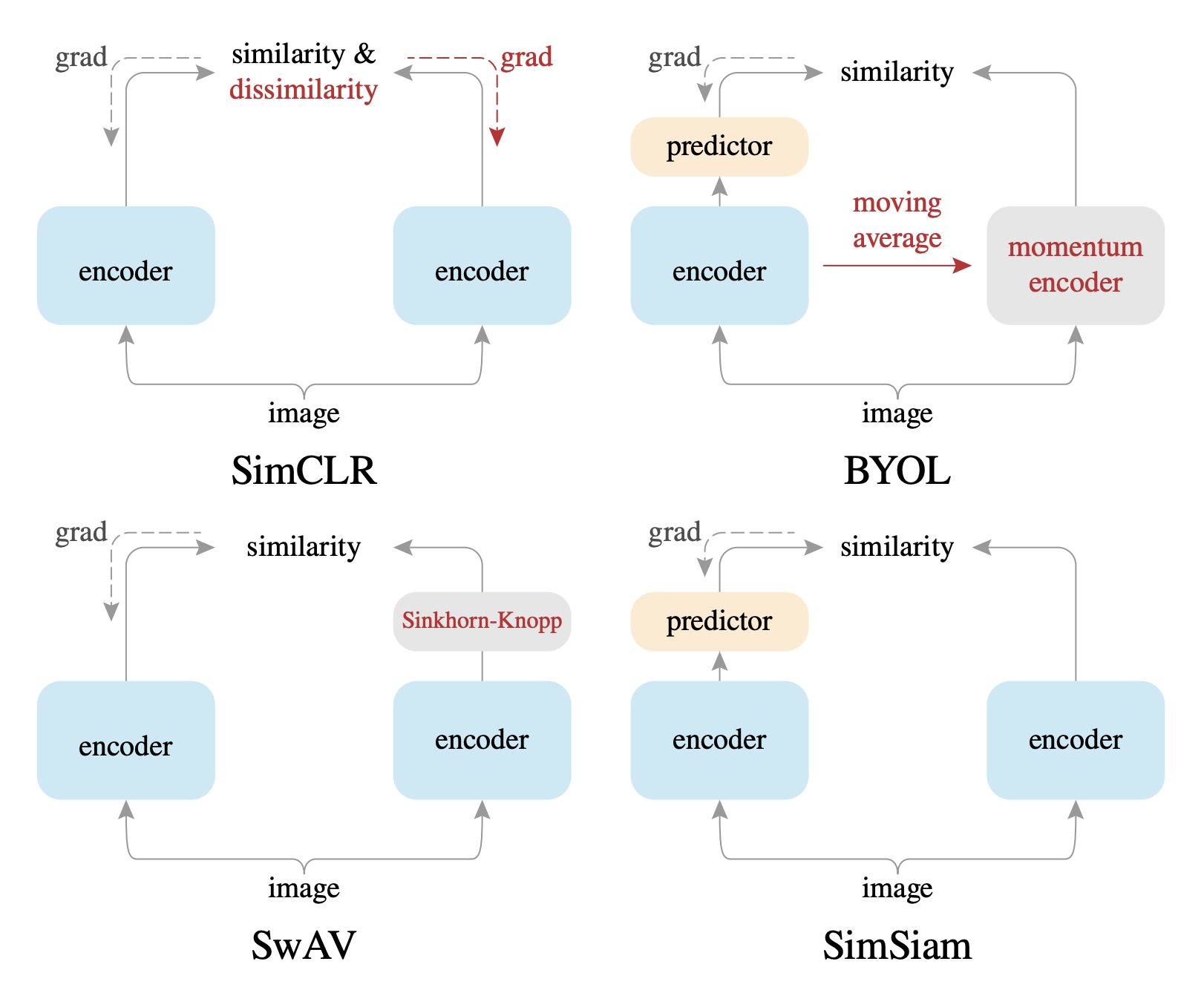}
\caption{Comparison of siamese architectures for contrastive learning \cite{chen2020exploring}.}
\label{fig:contrastiveframework}
\end{figure}

\BE{Simpification in learning frameworks} Another facet of contrastive learning development lies in enhancing the network structure. MoCo \cite{he2020momentum} contended that a sizable, consistent dictionary is imperative for achieving superior performance, subsequently proposing a momentum encoder to tackle this challenge. SimCLR \cite{chen2020simple} explored a wide range of data augmentation methods and introduced an MLP layer after the encoder, which bolstered accuracy by over 10 per cent. As a result, subsequent contrastive learning networks consistently incorporate an MLP to optimize performance. Ultimately the Siamese architectures are well-summarised \cite{chen2020exploring} in Figure \ref{fig:contrastiveframework}.

Given the vast availability of unlabeled data on the internet, contrastive learning has emerged as an incredibly adaptable framework capable of being trained for many tasks with suitably defined contrastive objectives. These tasks span image classification \cite{wu2018unsupervised, ye2019unsupervised, he2020momentum, chen2020improved, chen2020simple, caron2020unsupervised, grill2020bootstrap, chen2020exploring, tian2020makes, zbontar2021barlow, gidaris2021obow, tian2021divide, caron2021emerging, Radford2021LearningTV}, object detection \cite{Kim2021ViLTVT, Gu2021OpenvocabularyOD, Li2021GroundedLP}, semantic segmentation \cite{Li2022LanguagedrivenSS, Xu2022GroupViTSS}, video understanding \cite{Luo2021CLIP4ClipAE, Wang2021ActionCLIPAN}, and depth estimation \cite{Zhang2022CanLU, tian2020contrastive}. The state-of-the-art contrastive learning frameworks are concisely summarized in Table \ref{tab:contrastive}, illustrating their immense potential to surpass supervised learning by addressing data limitations, noisy labelling, and accuracy performance challenges.

\begin{landscape}
\begin{table}[!h]
    \centering
    \resizebox{\linewidth}{!}{
    \begin{tabular}{|c||c|c|c|c|c|c|}
    \hline
    Method & Positive Sample & Negative Sample & Contributions & Time & Task\\
    \hline
    InstDis \cite{wu2018unsupervised} & Augmented views & Other instances & Instance discrimination, memory bank & 2018 & Image classification\\
    \hline
    CPC \cite{oord2018representation} & Temporally nearby frames & Temporally distant frames & Autoregressive predictive coding & 2018 & Audio prediction\\
    \hline
    ISIF \cite{ye2019unsupervised} & Neighboring instances & Distant instances & Invariant and spreading feature learning & 2019 & Image classification\\
    \hline
    CMC \cite{tian2020contrastive} & Augmented views & Other samples in batch & Multiview encoding & 2020 & Depth estimation\\
    \hline
    MoCo-v2 \cite{he2020momentum, chen2020improved} & Augmented views & Queue-based samples & Momentum encoder & 2019 & Image classification\\
    \hline
    SimCLR \cite{chen2020simple} & Augmented views & Other samples in batch & New structure with MLP, strong data augmentation & 2020 & Image classification\\
    \hline
    SwAV \cite{caron2020unsupervised} & Augmented views & Cluter centroids & Swapped prediction & 2020 & Image classification\\
    \hline
    BYOL \cite{grill2020bootstrap} & Augmented views & None & Non-negative sample learning, stop-gradient & 2020 & Image classification\\
    \hline
    SimSiam \cite{chen2020exploring} & Augmented views & None & Simplified architecture & 2020 & Image classification\\
    \hline
    InfoMin \cite{tian2020makes} & Augmented views & Other samples in batch & Mutual information minimization & 2021 & Image classification\\
    \hline
    BarlowTwin \cite{zbontar2021barlow} & Augmented views & None & Cross-correlation reduction & 2021 & Image classification\\
    \hline
    OBoW \cite{gidaris2021obow} & Augmented views & Other samples in batch & Bag-of-Words & 2021 & Image classification\\
    \hline
    DC-v2 \cite{tian2021divide} & Augmented views & None & Local and global feature learning & 2021 & Image classification\\
    \hline
    DINO \cite{caron2021emerging} & Augmented views & None & Teacher-student distillation architecture& 2021 & Image classification\\
    \hline
    CLIP \cite{caron2021emerging} & Language-image pair & None & Multimodle learning & 2021 & Image classification\\
    \hline
    
    \end{tabular}}
    \caption{Summary of contrastive learning methods.}
    \label{tab:contrastive}
\end{table}
\end{landscape}

\chapter{Methodology}
\label{Chapter3}

\section{Major Challenges}
Two significant challenges exist in segmentation tasks, particularly in the medical imaging segmentation domain. The first challenge is data limitation, and the second is the high computational cost associated with high-resolution input data.

\BE{Data limitation} In medical image segmentation tasks, acquiring labelled data is extremely costly. Firstly, it requires manual annotation from human experts with a strong background in the biomedical domain to distinguish between the target object and noise. Furthermore, training data contains patients' private information, which poses challenges for data publication to protect patient's privacy. These two significant factors result in an average of only around 20 images for each dataset in vessel segmentation tasks.

\BE{High computational Cost} In contrast to general image classification tasks, such as ImageNet \cite{ILSVRC15} (size of $224 \times 224 \times 3$) and CIFAR10 \cite{cifar} (size of $32 \times 32 \times 3$), the input images for segmentation are used for real-world disease visualization and diagnosis. Examples of vessel segmentation datasets include DRIVE \cite{staal2004ridge} (size of $568 \times 585 \times 3$) and CHASE-DB1 \cite{owen2009measuring} (size of $1008 \times 1008 \times 3$). Consequently, the input data for medical image segmentation are generally high-resolution images. It is well known that computational complexity is exponentially proportional to the dimensions of the input image. Thus, an efficient network structure is essential for real-world medical image segmentation tasks to handle high-resolution input images.

\section{Methodology Overview}
\begin{figure}[h!]
  \centering
  \includegraphics[width = \textwidth]{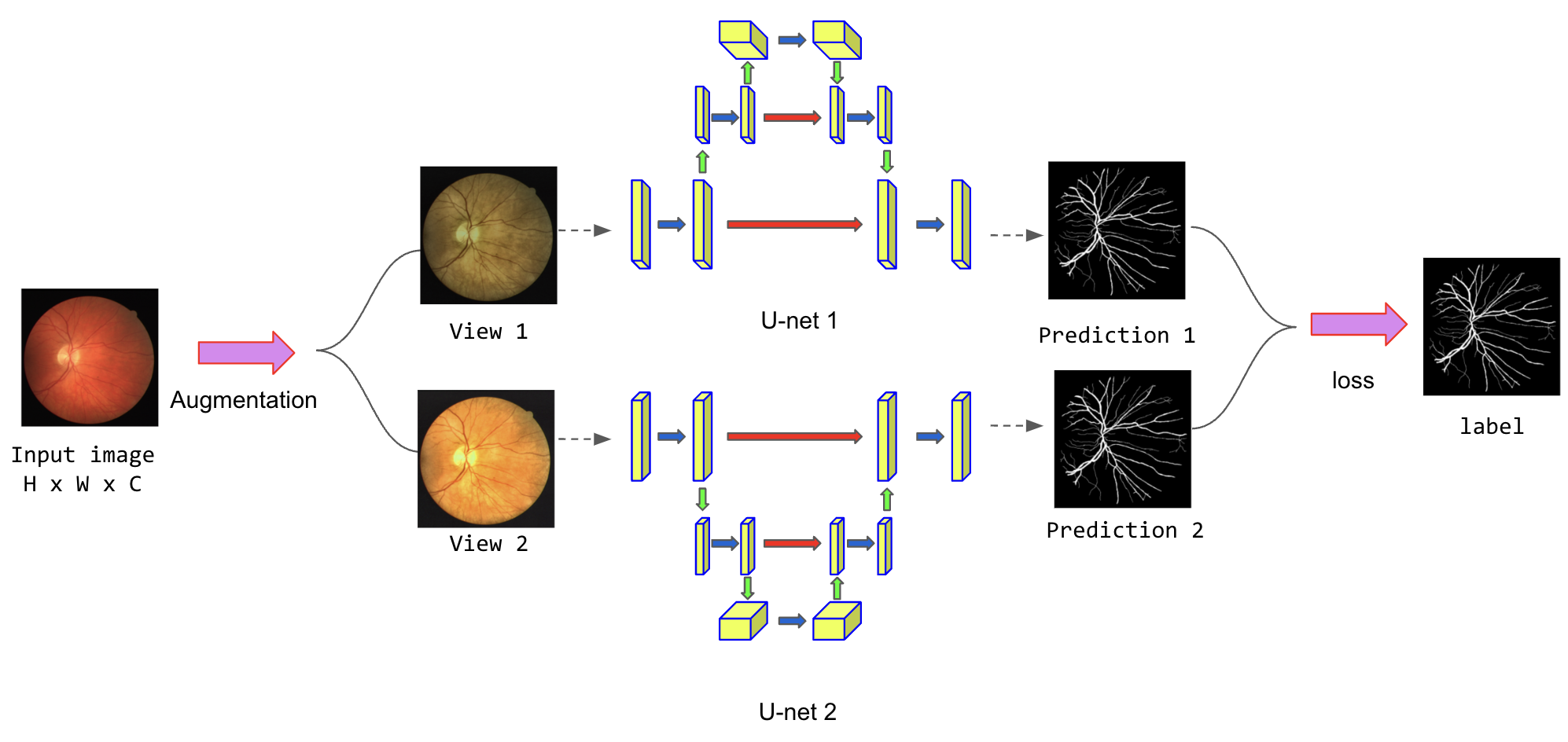}
  \caption{Structure of the AUtO-Net, which receives two views from the data augmentation from the input image to generate two predictions, applies an average loss of different views as the objective function.}
  \label{fig:auto_method}
\end{figure}

Our method addresses the two significant challenges mentioned above: data limitation and high computational cost. We designed a novel learning framework named AUtO to tackle these problems, which stands for \textbf{A}ugmentation-driven doubled \textbf{U}Ne\textbf{t} \textbf{O}-shape Network. The framework integrates three primary modules described in Figure \ref{fig:auto_method}.

Firstly, we introduce a novel contrastive framework to extract invariant features, aiming to reduce the variants caused by factors such as lighting, colour, and noise. It is worth noting that our method does not introduce extra training parameters.

Secondly, after reviewing the existing backbone network methodologies, we specifically designed our backbone network to achieve high performance with efficient computational cost. We found that pure CNNs excel at extracting local features but cannot capture global features. On the other hand, pure Transformers are good at extracting long-distance dependencies but require high computational resources. Therefore, AUtO is designed to combine these two advantages with high efficiency.

Finally, we review and implement extensive existing data augmentation methods, testing their effectiveness in addressing the data limitation problem. We have also proposed a new data augmentation method called MixUp.

\section{Contrastive Multiview Learning}
\subsection{Contrastive Learning Limitations}
Inspired by the impressive performance of emerging contrastive learning networks across various research domains, such as Natural Language Processing, Computer Vision, and Audio Recognition (see Chapter \ref{Chapter2} for a comprehensive literature review), contrastive learning has proven to be a versatile and adaptable self-supervised learning framework with outstanding transferability. This framework can achieve high performance in downstream tasks, including object detection and segmentation, thus representing a promising research direction. Employing the contrastive learning framework for downstream tasks like segmentation typically involves a two-step training process. The first step entails training the siamese encoder structure with a predictor layer, usually an MLP, on a substantial dataset like ImageNet \cite{ILSVRC15} with unlabelled images. Only the encoder part is retained in the second step, while the other components are discarded. Subsequently, the decoder is trained separately on the downstream dataset.

Despite the promising results demonstrated by the use of contrastive learning for medical image segmentation tasks, some concerns warrant critical evaluation. The following are four primary issues raised in the given statement:

\BE{Disruption of End-to-End learning} Contrastive learning deviates from the conventional end-to-end training process, as it involves a two-step procedure comprising pretraining and fine-tuning. This separation may render the learning process more intricate and potentially less efficient than a unified end-to-end approach.

\BE{Resource wastage} Contrastive learning approaches involve retaining only the encoder for downstream tasks while discarding other network components, such as the projector and predictor. This scenario can result in resource wastage to some extent, as the discarded elements might have been beneficial for the specific downstream task.

\BE{Necessity of datasets} Contrastive learning frequently depends on a sizeable pretraining dataset and the dataset required for the specific downstream task. While this method is prevalent in mainstream computer vision, it poses challenges in the medical image segmentation domain, where datasets are often smaller and procuring a sizable pretraining dataset is infeasible.

\begin{figure}[h!]
\centering
\includegraphics[scale=0.2]{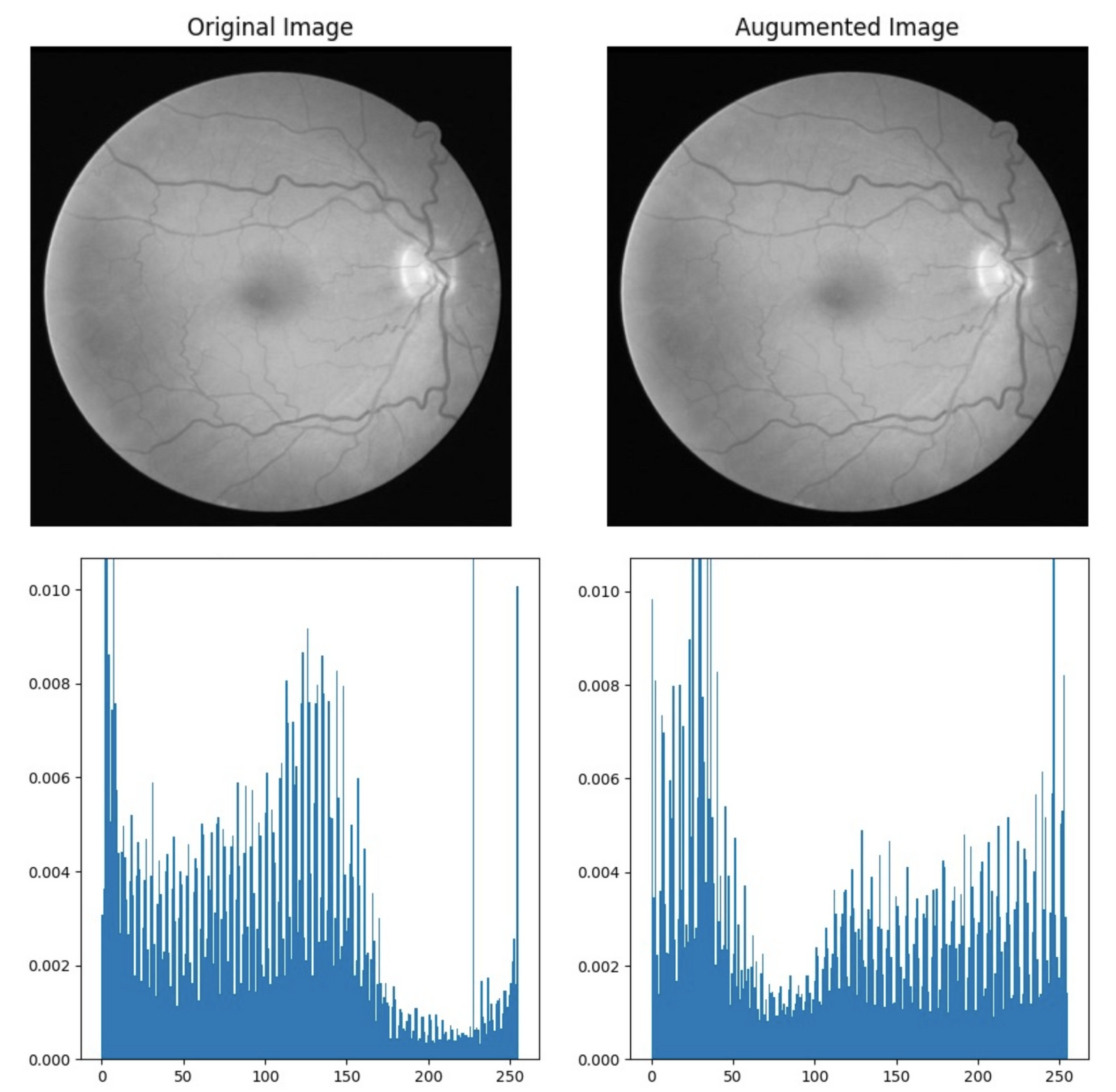}
\caption{Distribution gap between the training and testing data caused by data augmentation, such as colour transformation, where the difference is difficult to detect by the human eye. Nevertheless, it significantly shifts image intensity distribution for small datasets.}
\label{fig:distributiongap}
\end{figure}

\BE{Distribution gap by augmentations} Since medical image segmentation contains limited data, data augmentation may cause a distribution gap between the training and testing data, as visually illustrated in Figure \ref{fig:distributiongap}. This phenomenon may not be problematic for general computer vision tasks, as the dataset is typically large and diverse. However, small datasets are sensitive to data augmentations, which may lead to high generalisation ability but with a performance drop due to overfitting.

\subsection{Multiview Learning Module}
\begin{figure}[h!]
  \centering
  \includegraphics[width = \textwidth]{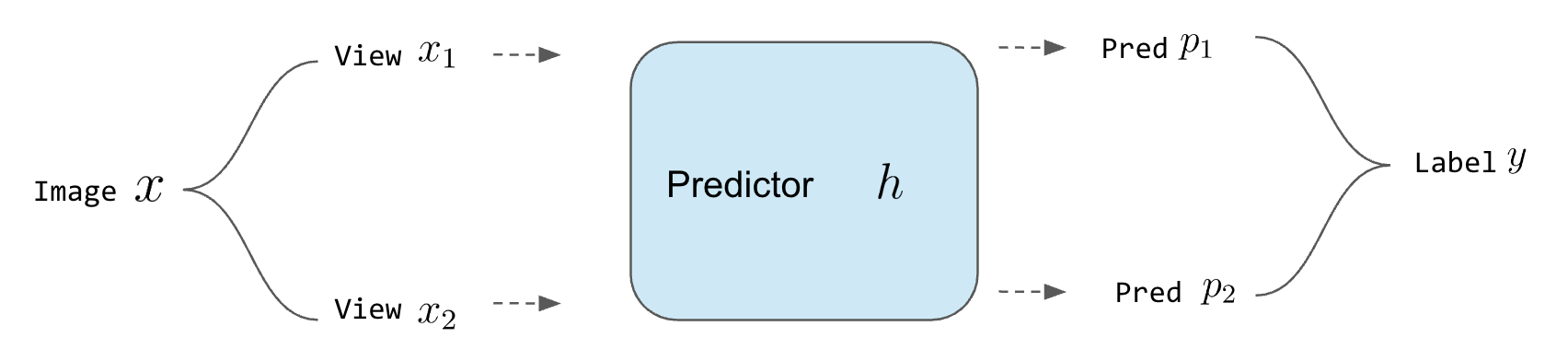}
  \caption{Detailed contrastive multiview learning framework of AUtO, composited with three parts. The data augmentation, a predictor and a loss function.}
  \label{fig:auto_detail_method}
\end{figure}

Therefore, we simplified the contrastive learning framework and developed it into a lightweight end-to-end multi-views supervised learning architecture without introducing extra parameters and reducing the reliability of the enormous data set. As illustrated in Figure \ref{fig:auto_detail_method}, the forward process of the framework can be summarised as the following.

Firstly, the input image $x$ is transferred with the data augmentation to two views, $x_1$ and $x_2$. We assume that the data augmentation will not change the semantic information of the original image. Specifically, the augmented image should keep the continuous curvilinear structure of the blood vessel but the different background information such as the position, colour, and light. Therefore, the model is expected to be robust against noise, extracting the invariant vessel feature and mapping views into the same latent space. This could be achieved by passing two views into the predictor, a segmentation predictor $h$ such as UNet, which is an encoder and decoder structure mapping the input image to the segmented images, to obtain two predictions denoted as $p_1$ and $p_2$. Then, the two predictions should have a similar output. Finally, two predictions are compared with the label. The final objective function could be represented as the sum of the individual loss of the prediction with the label, mathematically illustrated by Eq.(\ref{eq:contrastloss}):

\begin{equation}
    \mathcal{L}_{joint} = \mathcal{L}(pred_1, label) + \mathcal{L}(pred_2, label).
    \label{eq:contrastloss}
\end{equation}

\begin{figure}[h!]
  \centering
  \includegraphics[width = \textwidth]{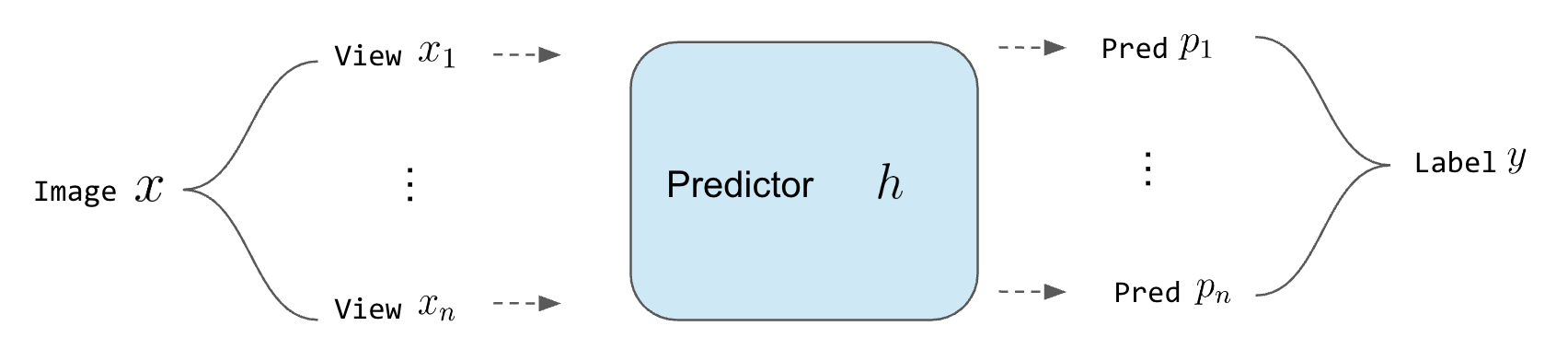}
  \caption{Generalised version of the contrastive multiview learning module of AUtO-Net from two to multiple views.}
  \label{fig:auto_detail_method2}
\end{figure}

Moreover, the performance could be far more extended from two to multiple views, as demonstrated in Figure \ref{fig:auto_detail_method2}. Then, the model could learn more views and the same time. To generalise, the final loss function is shown by Eq.(\ref{eq:contrastlossgeneral}):

\begin{equation}
    \mathcal{L}_{joint} = \sum_i^n \mathcal{L}(pred_i, label),
    \label{eq:contrastlossgeneral}
\end{equation}
where $\mathcal{L}$ could be any segmentation loss such as Binary Cross-Entropy Loss, Focal Loss or Dice Loss (See Chapter \ref{Chapter2} for the literature review). Since only modifying with the loss function. Therefore, no extra parameters are introduced. The Algorithm \ref{alg:auto} explains the pseudo-code of the proposed method.

\begin{algorithm}
\caption{Pseudocode of AUtO Contrastive Multiview Learning with Two Views}
\label{alg:auto}
\begin{algorithmic}
\Repeat
\State $x, y \sim \mathcal{D}$ \Comment{Simple image and label from the population.}
\State $x_1, x_2 \gets aug(x), aug(x)$ \Comment{Get two augmented images.}
\State $p_1, p_2 \gets h(x_1), h(x_2)$ \Comment{Get two predicted images from the model.}
\State $\mathcal{L} \gets \left(\mathcal{L}(p_1, y) + \mathcal{L}(p_2, y) \right)/2$ \Comment{Calculate the joint loss for both views.}
\State $\mathcal{L} \gets \mathcal{L} - \nabla \mathcal{L}$ \Comment{Backpropaget the loss to update weights.}
\Until{converage}
\end{algorithmic}
\end{algorithm}

\section{Hybrid Model Strucutre}
\subsection{Analyis of Backbone Networks}
The design of the hybrid model comes from the analysis of the existing backbone models with their advantages and limitations (See Chapter \ref{Chapter2} for the reviews of network structures). The benchmark methods highlight three distinct stages of deep learning approach development: pure CNN-based backbones, hybrid CNN models incorporating attention mechanisms, and pure Transformer networks.

\BE{Pure CNNs} Conventional approaches for vessel segmentation primarily stemmed from the UNet family, relying on CNN-based backbones. The essence of CNNs involves using filters (kernels) to process intricate image inputs, targeting and extracting features such as shapes, edges, or textures. Combining various filters facilitates mapping complex images into a rich feature space. With increasing hidden layers, models can learn simple texture and shape features and progress to higher-level abstract patterns with semantic meaning. The success of CNNs hinges on the inductive bias optimized for vision tasks.

\begin{itemize}
    \item \textbf{Localisation:} CNNs could effectively capture the local structures and features.

    \item \textbf{Translation invariance:} CNNs employ pooling layers and increasing receptive fields, enabling the identification of image features irrespective of position transformation \cite{kaudererabrams2017quantifying}.

    \item \textbf{Weight sharing:} CNNs filter share parameters, reducing trainable parameters while maintaining high efficiency.
\end{itemize}

Additionally, the residual block \cite{he2015deep} has been proposed, prompting many methods to adopt the residual mechanism for more stable training and enhanced performance. However, CNNs' primary shortcoming is its inability to capture global features effectively due to the down-pooling in its architecture. While down-pooling allows CNNs to have larger receptive fields and learn global information, it simultaneously reduces image resolution and sacrifices detail, which results in the poor ability to extract the global features.

\BE{Vision Transformers} Compared with CNNs, the Vision Transformer exhibit outstanding scaling properties, maintaining robust performance as model parameters increase. Previous studies \cite{dosovitskiy2020image} demonstrated that the self-attention mechanism excels at learning long-distance dependencies, visualizing these dependencies during only a few training epochs. This advantage is particularly beneficial for segmenting non-linear structures, such as retinal vessels. The self-attention mechanism has proven to be a powerful tool for accentuating essential features and improving performance. However, there are two major shortcomings of the pure Vision Transformer. The first one is it does not optimise for the image tasks and is without two essential inductive biases like CNNs do. Therefore, the model must learn the basic assumptions of the large data set and requires a longer training time. Moreover, with increased parameters, the model becomes more complicated and prone to overfitting. In conclusion, current Vision Transformers rely heavily on computational resources, making training time-consuming and financially costly.

\subsection{Modified Residual Attention Block}
\begin{figure}[h!]
\centering
\includegraphics[width=0.8\textwidth]{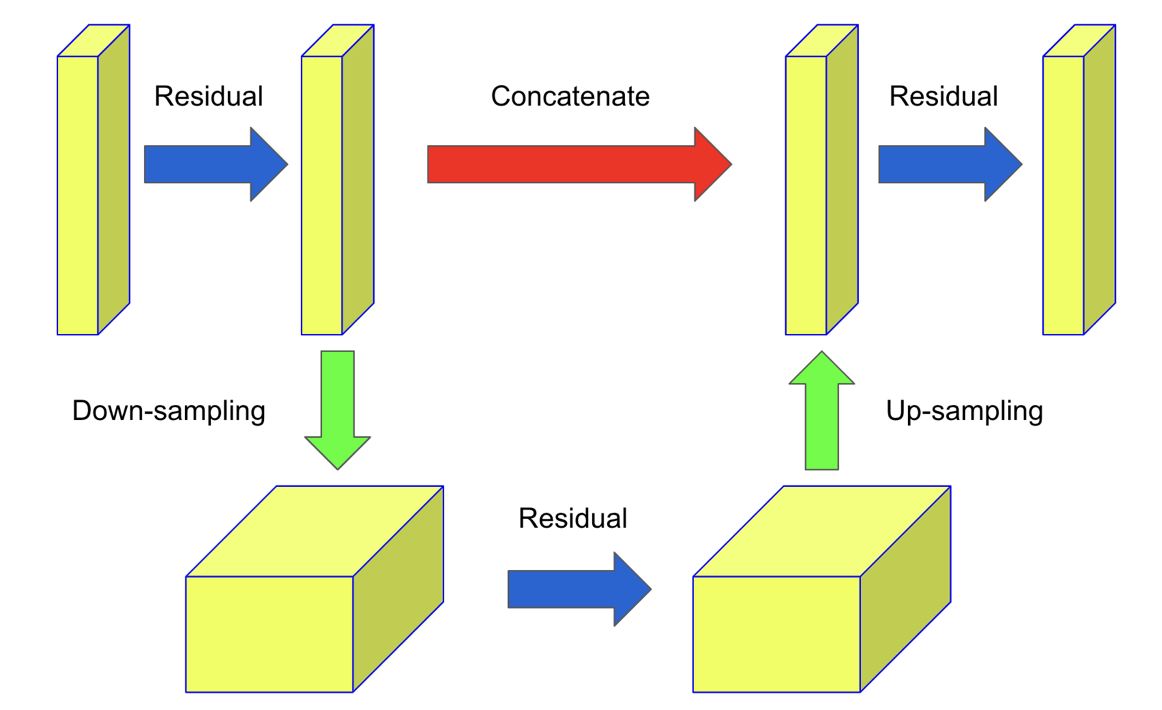}
\caption{The detail of the UNet block is composited by three major parts: residual block, concatenation and down-sampling.}
\label{fig:detailedunet}
\end{figure}

A modified residual attention block has implemented the proposed hybrid model to optimise the segmentation process. Specifically, we leverage residual network (ResNet \cite{he2015deep}) with an attention block. This unique configuration amalgamates the innate efficiency and inductive bias of CNNs with the exceptional global feature learning capability of Vision Transformers, culminating in a unified, high-performance structure. The intricate structure of the UNet model is illustrated in Figure \ref{fig:detailedunet}.

The feature map is primarily forwarded into a residual block, signified by the blue arrow in the figure. This residual operation is a potent solution to the vanishing gradient issue, stabilizing the training process. Subsequently, the feature map embarks on two separate paths. One trajectory incorporates a skip connection, which concatenates the feature map from the encoder with that of the decoder, enhancing the reconstruction of intricate vessel details. In contrast, the other path employs a down-sampling operation, effectively reducing the feature map's size to facilitate high-level feature extraction.

\begin{figure}[h!]
\centering
\includegraphics[width=0.7\textwidth]{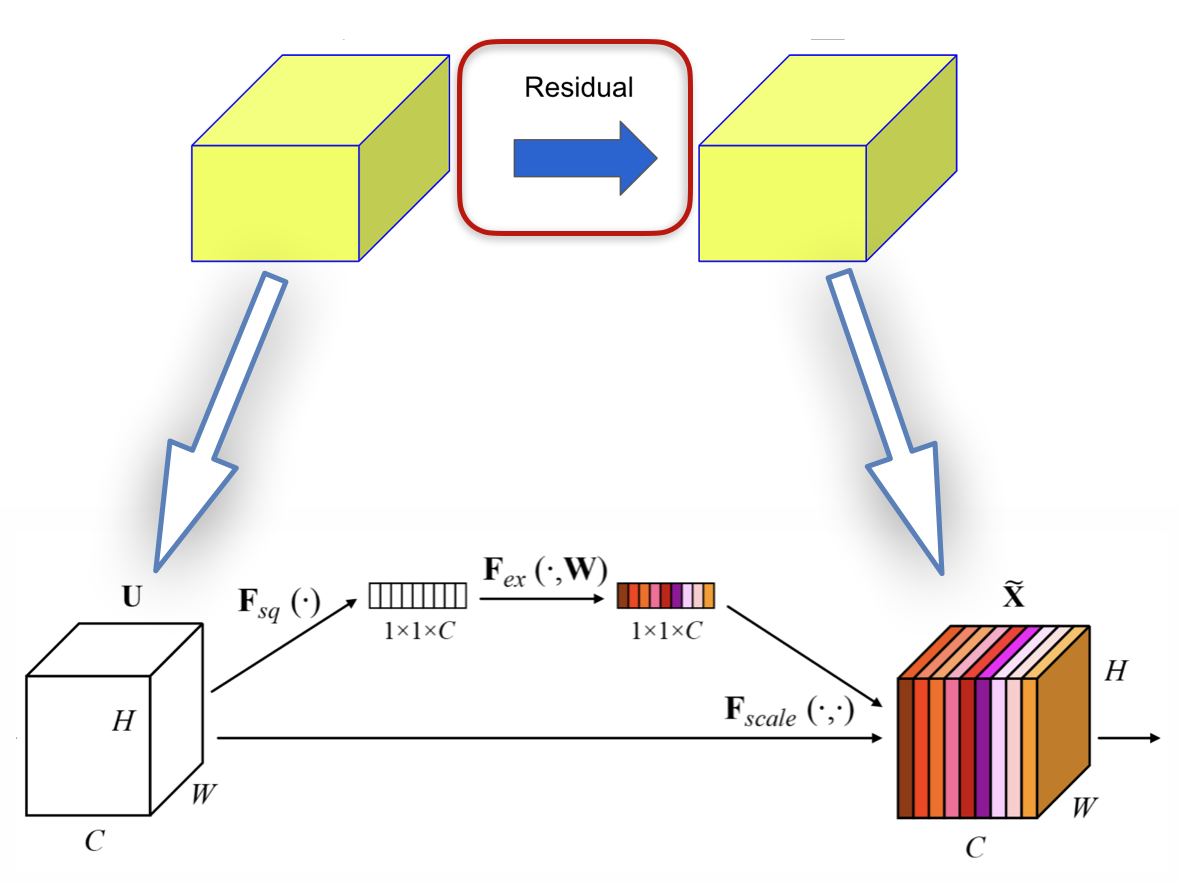}
\caption{Modified UNet residual block with SE block \cite{hu2018squeeze}.}
\label{fig:seblock}
\end{figure}

The proposed method integrates an attention mechanism into the residual block by leveraging the Squeeze-and-Excitation block (SE block \cite{hu2018squeeze}) to calculate channel attention. This modification creates a unified AUtO residual block encapsulating the SE block. The operation primarily involves two steps: the squeeze operation, which encapsulates global information, and the excitation operation, which captures channel-wise dependencies, as illustrated in Figure \ref{fig:seblock}.

\BE{Squeeze} The first step is to compute the weights $z$ across channels. The squeeze operation, denoted as $F_{sq}(\cdot)$, integrates spatial information by reshaping the feature map $U$ from dimensions $H \times W \times C$ to $1 \times 1 \times C$. The mathematical formula for this operation is shown in Eq.(\ref{eq:squzee}):

\begin{equation}
z = F_{sq}(U) = \frac{1}{H \times W} \sum_{i=1}^{H} \sum_{j=1}^{W} U(i, j).
\label{eq:squzee}
\end{equation}

\BE{Excitation} After obtaining the one-dimensional channel feature map via the squeeze operation, the next step is to compute each channel's weight $s$ and dependencies using the attention mechanism. The excitation operation, $F_{ex}$, is illustrated in Eq.(\ref{eq:excitation}):

\begin{equation}
s = F_{ex}(z, W) = \sigma(g(z, W)) = \sigma(W_2 \delta(W_1z)),
\label{eq:excitation}
\end{equation}
here, $\sigma$ represents the activation function \cite{nair2010rectified, xu2015empirical, ramachandran2017searching, hendrycks2016gaussian, he2015delving} such as ReLU. The matrices $W_1 \in \mathcal{R}^{\frac{C}{r} \times C}$ and $W_2 \in \mathcal{R}^{C \times \frac{C}{r}}$ control the computational complexity through a reduction ratio $r$. Finally, the computed channel weight is multiplied by the original feature map $U$ to obtain the final output $\widetilde{x}$ via the scaling operation $F_{scale}$, as denoted in Eq.(\ref{eq:scale}):

\begin{equation}
\widetilde{x} = F_{scale}(U, s) = U \cdot s.
\label{eq:scale}
\end{equation}

\section{Data Augmentations}
\subsection{Summary of Augmentation Methods}
Data augmentation is an integral operation in machine learning processes. It is instrumental in creating a robust model capable of learning effective feature representations. Significantly, it enhances the performance of medical image segmentation tasks without necessitating additional training time, thus optimizing resource utilization. \cite{uysal2021exploring} shows the vanilla UNet, when equipped with the appropriate implementation of data augmentations, can achieve benchmark results, thereby proving the effectiveness of these augmentation techniques.

\begin{figure}[h!]
\centering
\begin{subfigure}{.19\textwidth}
  \centering
  \includegraphics[width=.95\linewidth]{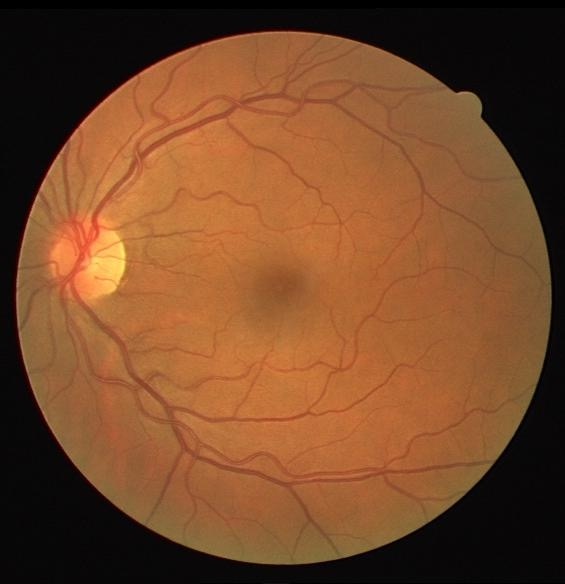}
  \caption{Original}
  \label{fig:sub1}
\end{subfigure}%
\hfill
\begin{subfigure}{.19\textwidth}
  \centering
  \includegraphics[width=.95\linewidth]{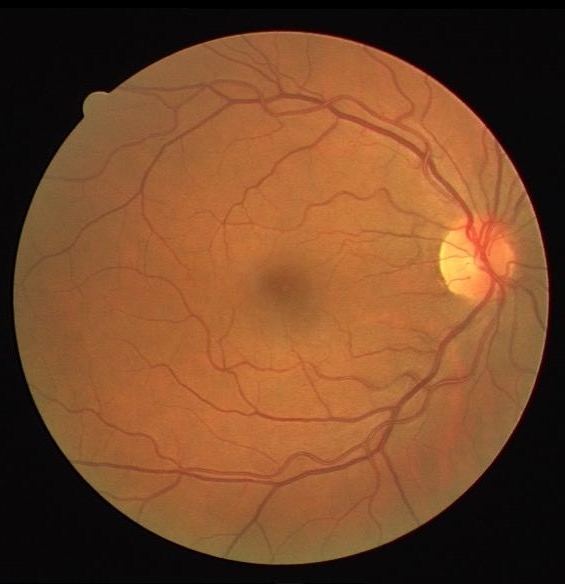}
  \caption{Horizon Flip}
  \label{fig:sub2}
\end{subfigure}%
\hfill
\begin{subfigure}{.19\textwidth}
  \centering
  \includegraphics[width=.95\linewidth]{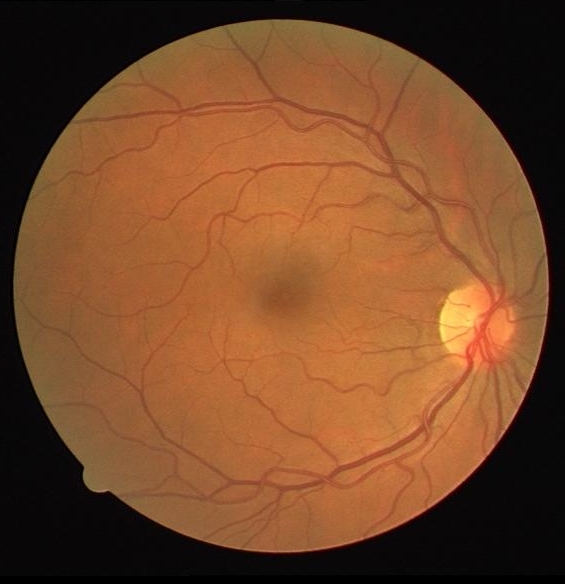}
  \caption{Vertial Flip}
  \label{fig:sub3}
\end{subfigure}
\hfill
\begin{subfigure}{.19\textwidth}
  \centering
  \includegraphics[width=.95\linewidth]{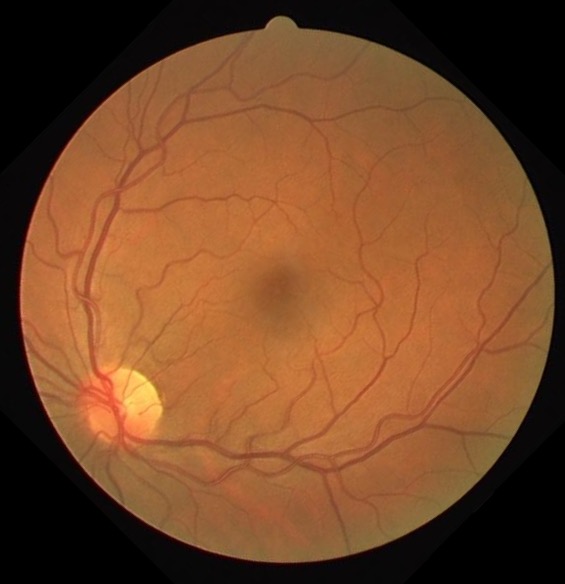}
  \caption{Rotation 45 $^\circ$}
  \label{fig:sub4}
\end{subfigure}%
\hfill
\begin{subfigure}{.19\textwidth}
  \centering
  \includegraphics[width=.95\linewidth]{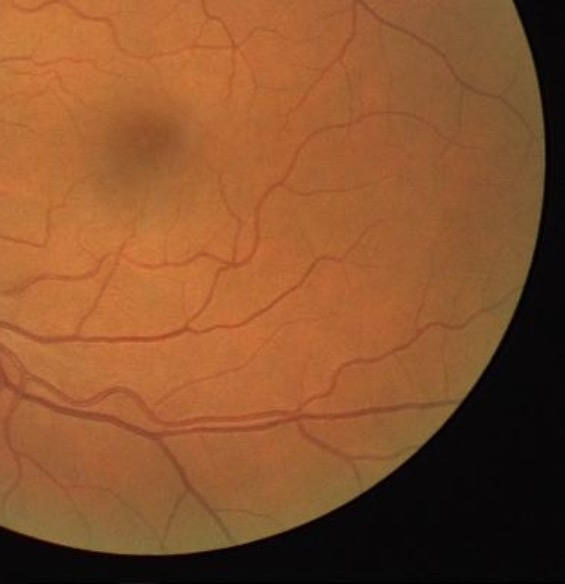}
  \caption{Crop}
  \label{fig:sub5}
\end{subfigure}

\medskip

\centering
\begin{subfigure}{.19\textwidth}
  \centering
  \includegraphics[width=.95\linewidth]{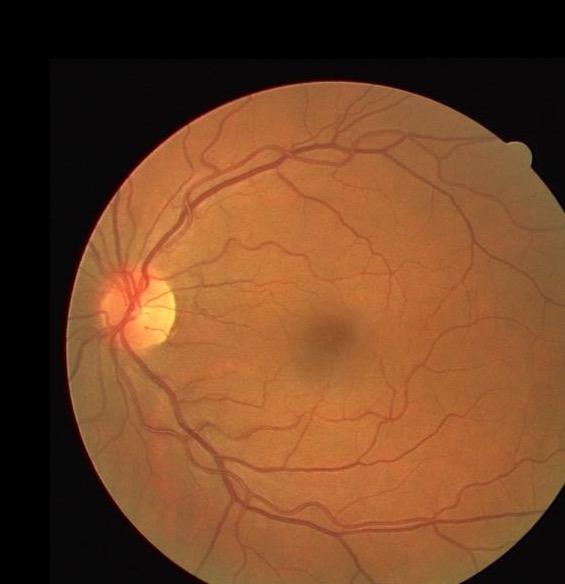}
  \caption{Translation}
\end{subfigure}%
\hfill
\begin{subfigure}{.19\textwidth}
  \centering
  \includegraphics[width=.95\linewidth]{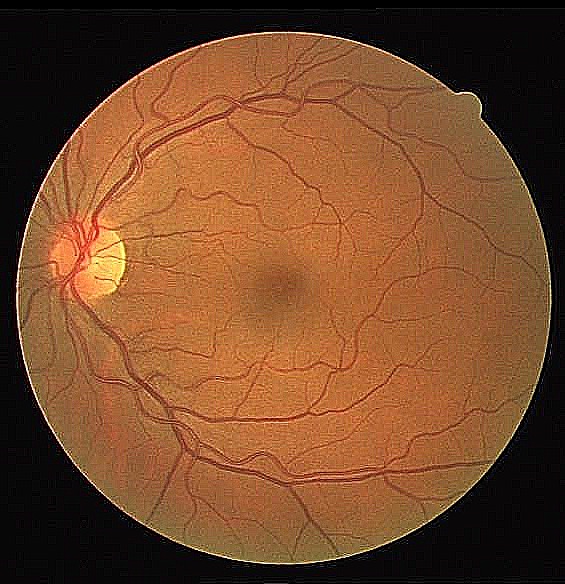}
  \caption{Sharpen}
\end{subfigure}%
\hfill
\begin{subfigure}{.19\textwidth}
  \centering
  \includegraphics[width=.95\linewidth]{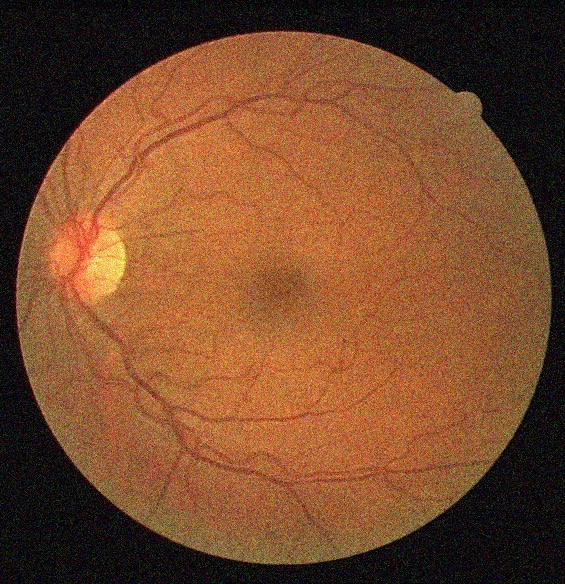}
  \caption{Gauss Noise}
\end{subfigure}
\hfill
\begin{subfigure}{.19\textwidth}
  \centering
  \includegraphics[width=.95\linewidth]{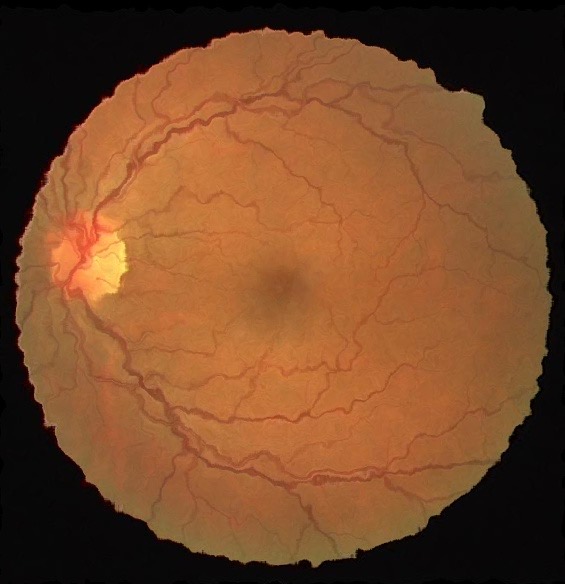}
  \caption{Elastic}
\end{subfigure}%
\hfill
\begin{subfigure}{.19\textwidth}
  \centering
  \includegraphics[width=.95\linewidth]{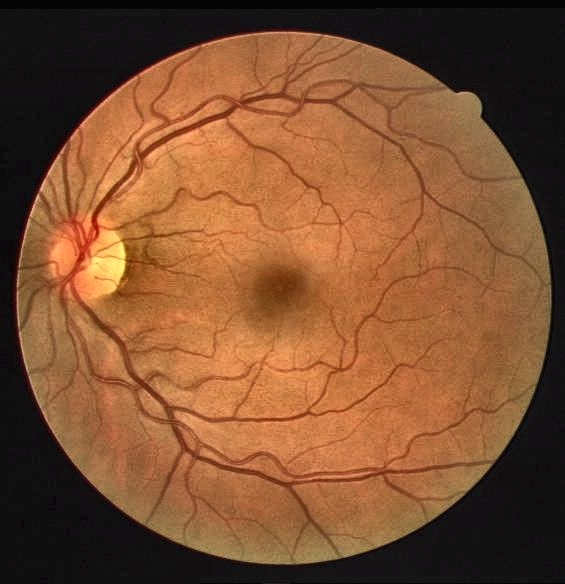}
  \caption{CLAHE}
\end{subfigure}%

\medskip 

\begin{subfigure}{.19\textwidth}
  \centering
  \includegraphics[width=.95\linewidth]{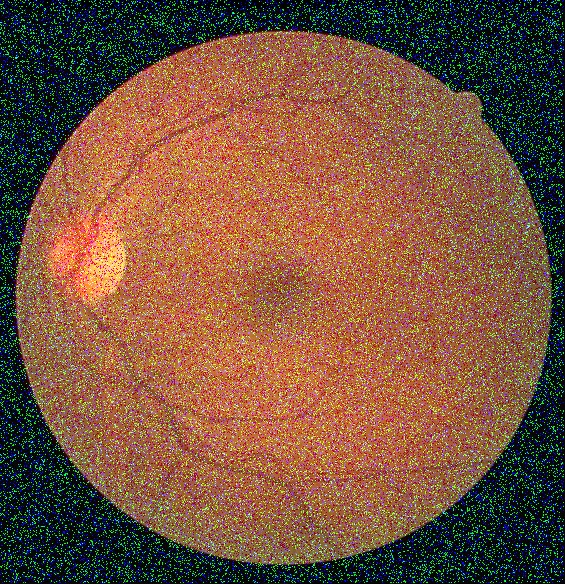}
  \caption{Salt Noise}
\end{subfigure}
\hfill
\begin{subfigure}{.19\textwidth}
  \centering
  \includegraphics[width=.95\linewidth]{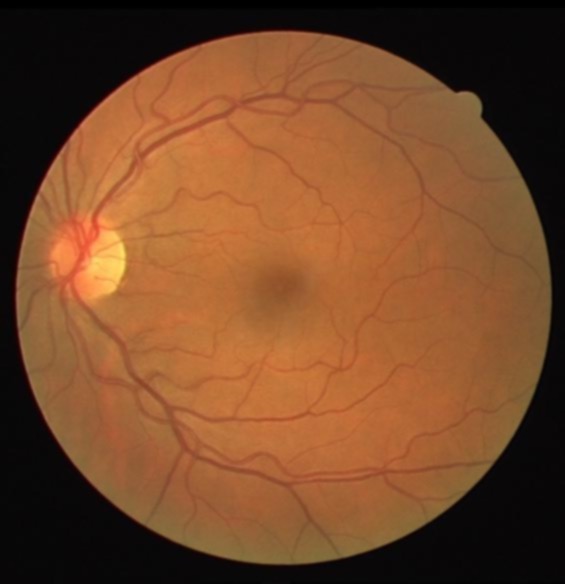}
  \caption{Gauss Blur}
\end{subfigure}
\hfill
\begin{subfigure}{.19\textwidth}
  \centering
  \includegraphics[width=.95\linewidth]{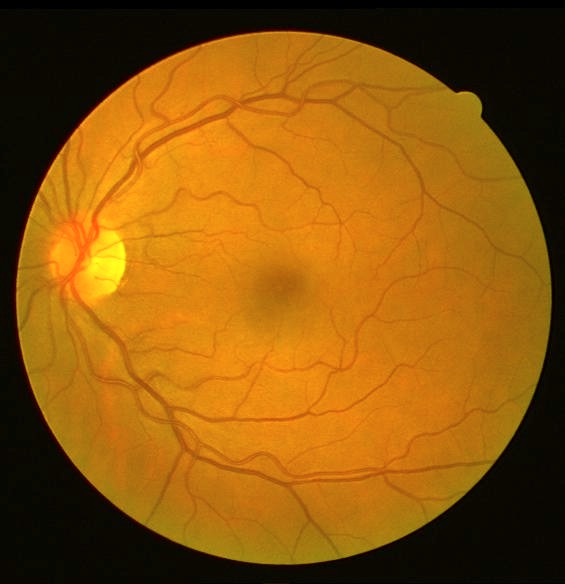}
  \caption{Colour}
\end{subfigure}
\hfill
\begin{subfigure}{.19\textwidth}
  \centering
  \includegraphics[width=.95\linewidth]{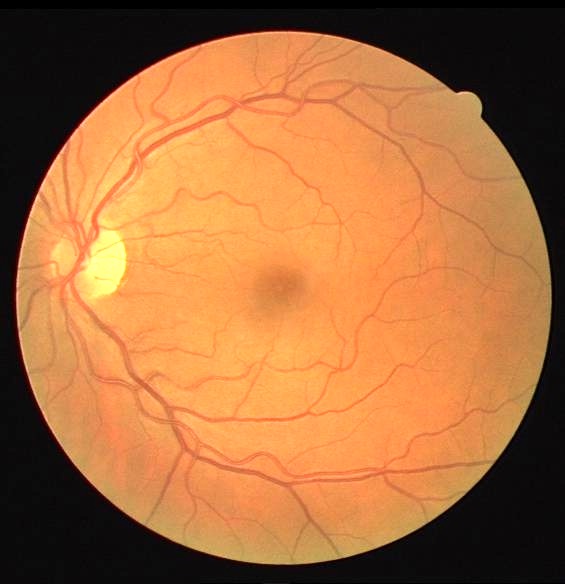}
  \caption{Brightness}
\end{subfigure}
\hfill
\begin{subfigure}{.19\textwidth}
  \centering
  \includegraphics[width=.95\linewidth]{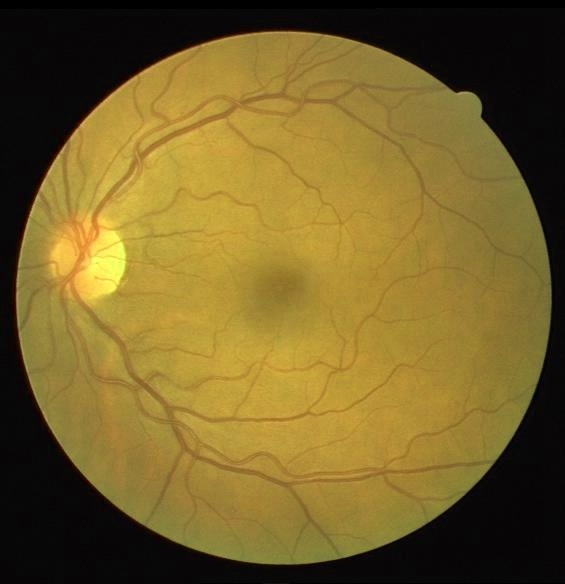}
  \caption{Hue}
\end{subfigure}

\medskip 

\begin{subfigure}{.19\textwidth}
  \centering
  \includegraphics[width=.95\linewidth]{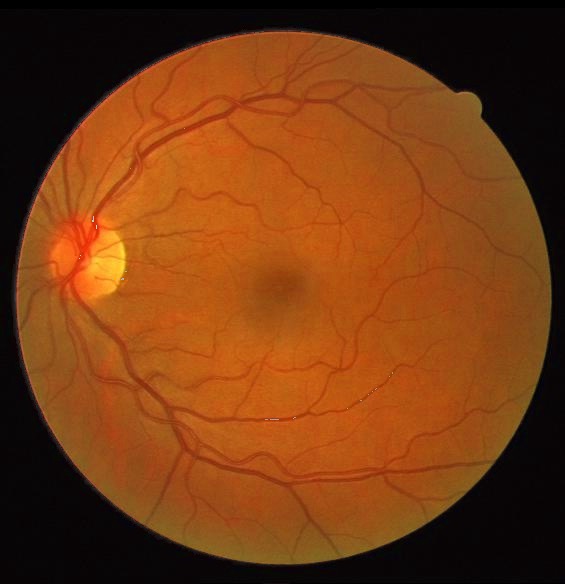}
  \caption{Saturation}
\end{subfigure}
\hfill
\begin{subfigure}{.19\textwidth}
  \centering
  \includegraphics[width=.95\linewidth]{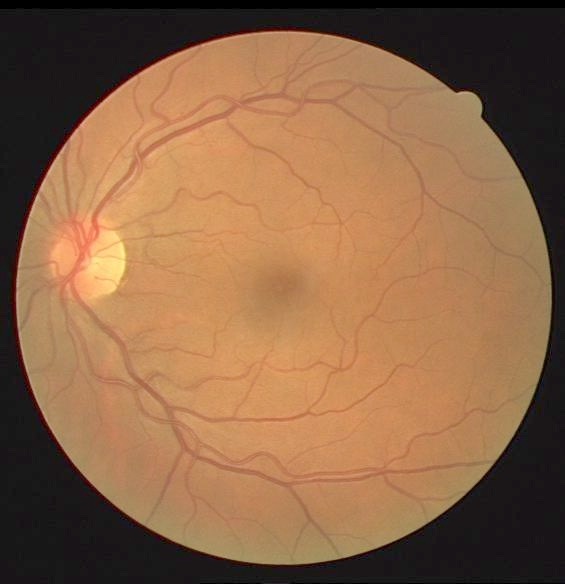}
  \caption{$\gamma$ Correction}
  
\end{subfigure}
\hfill
\begin{subfigure}{.19\textwidth}
  \centering
  \includegraphics[width=.95\linewidth]{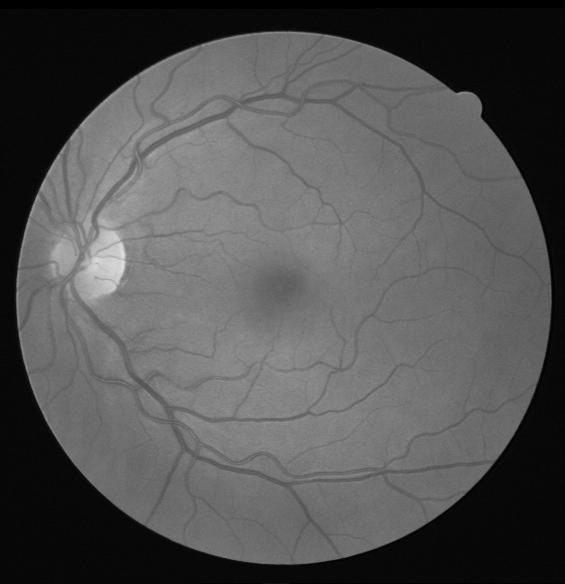}
  \caption{Grey}
  
\end{subfigure}
\hfill
\begin{subfigure}{.19\textwidth}
  \centering
  \includegraphics[width=.95\linewidth]{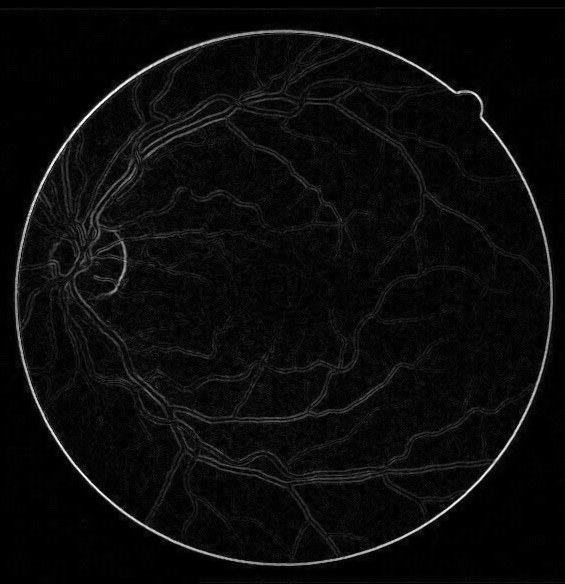}
  \caption{Sobel}
  
\end{subfigure}
\hfill
\begin{subfigure}{.19\textwidth}
  \centering
  \includegraphics[width=.95\linewidth]{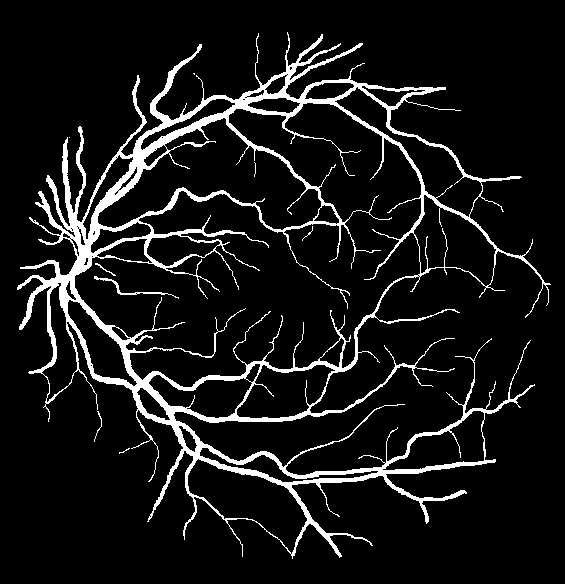}
  \caption{Label}
  
\end{subfigure}

\caption{Visual examples of data augmentation methods for the retina image from the DRIVE dataset. Where horizon and vertical file, rotation, crop and translation belong to spatial transformation, the last image is the ground truth label for better visual comparison among the methods. And remaining augmentation methods are pixel-wise transformations. This figure serves as a particular visual illustration of the transformed retina images.}
\label{fig:augall}
\end{figure}

Previous studies have also validated the effectiveness of data augmentations, particularly their composition, in image classification tasks \cite{chen2020simple}. These augmentation methodologies for segmentation tasks can broadly be categorized into spatial and pixel-level transformations. Figure \ref{fig:augall} illustrates the augmentations implemented to build a visual concept of the effect of various data augmentation methods.

\BE{Spatial transformation} Spatial or geometric transformations are the techniques that manipulate the structure of the image without altering the object's inherent attributes. By doing so, they compel the model to learn position-invariant features. This strategy effectively mitigates the potential issue of the model relying on simple memorization of geometric positions, a shortcut solution that could hinder the model's generalization ability. Examples of spatial transformations encompass various operations such as horizontal or vertical flipping, resizing or rotation \cite{gidaris2018unsupervised}, random cropping, and zooming in or out. These operations enhance the model's robustness to object orientation, size, and position changes.

\BE{Pixel-level transformation} The other class of transformations, pixel-level transformations, involves more complexity by altering the pixel value of the image while preserving the semantic information of the image. The primary objective of these transformations is to guide the model to learn invariant features despite changes in colour, texture, or other pixel-level details. For instance, Contrast Limited Adaptive Histogram Equalisation (CLAHE \cite{reza2004realization}) is a technique that enhances the contrast between vessel structures and the background, thereby making the objects of interest more distinguishable by reducing the effects of noise. Colour jitter, encompassing brightness transformations, hue value saturation, and contrast \cite{howard2013some, szegedy2015going}, simulates real-world noise caused by lights and shadows, thereby preparing the model to handle various lighting conditions. Additionally, Gaussian blur simulates the effect of an unfocused camera, introducing a certain level of uncertainty in object boundaries. Gamma correction, on the other hand, improves the image's contrast or ensures colour and brightness consistency across different devices, ensuring that the model is not confused by device-specific image rendering differences. Elastic transformation is another noteworthy technique that simulates tissue shape changes, thereby providing additional training samples that account for potential variations in the shape and structure of the tissues in different medical images.

\clearpage
\subsection{Proposed Augmentation MixUp}
To improve performance in medical image segmentation tasks, we propose an innovative data augmentation method known as MixUp. This technique is designed with two distinct variants in mind: the first merges training images randomly, while the second integrates the image with its associated label.

\subsection{MixUp by Image}
\begin{figure}[h!]
\centering
\includegraphics[width =\textwidth]{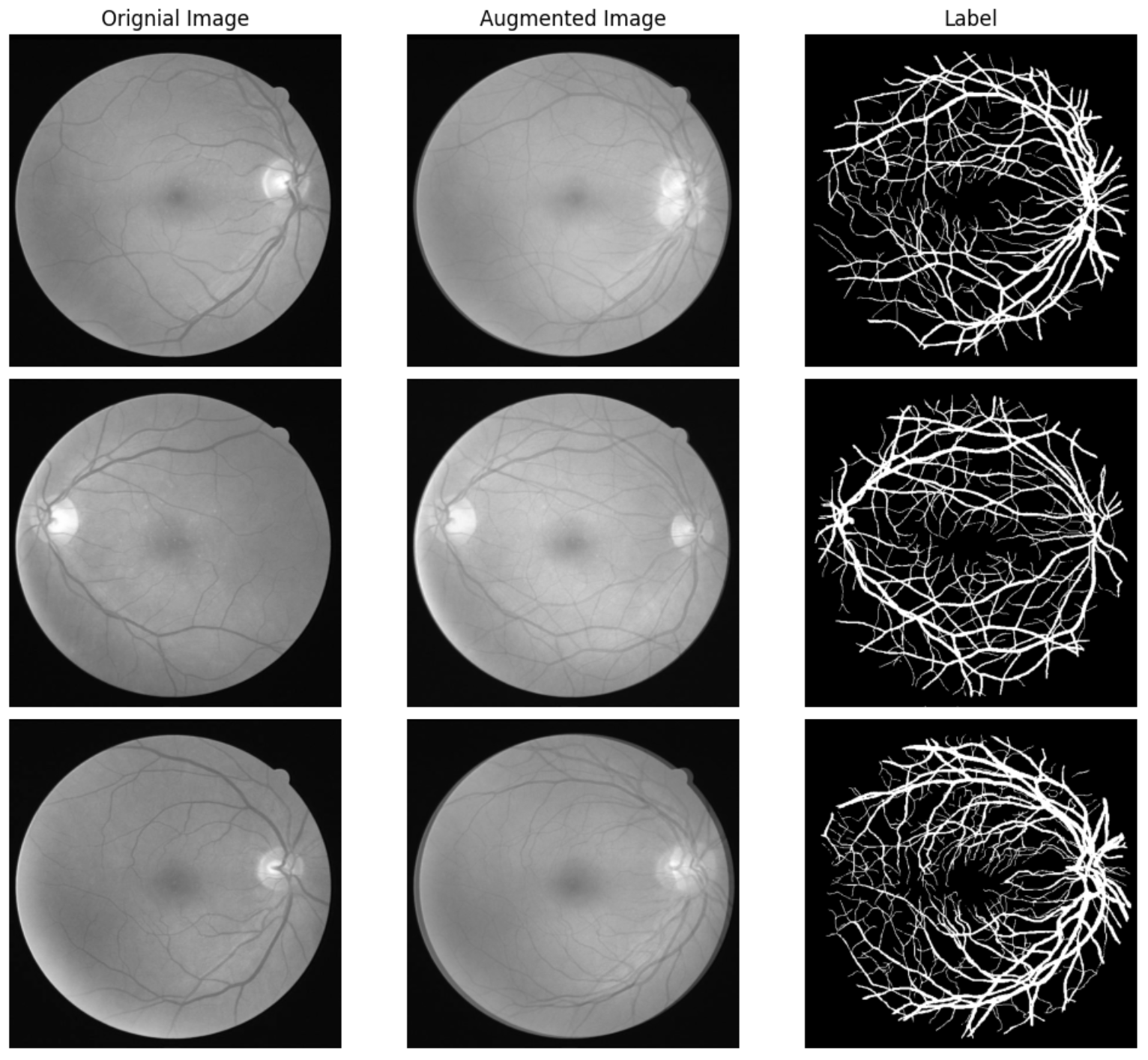}
\caption{MixUp augmentation: a visual depiction of the original images versus their randomly summed counterparts.}
\label{fig:mixupimage}
\end{figure}

Drawing inspiration from MixGen's concept \cite{hao2023mixgen}. The MixUp technique advocates for the random summation of images. In the realm of vessel segmentation, this technique demands the fusion of labels whenever their corresponding images are merged. This augmented approach introduces more significant variance, thus ensuring the model understands that the summation operation does not distort the fundamental semantic information of the vessel structure. Figure \ref{fig:mixupimage} visually represents the contrast between the original and augmented images using MixUp.

The MixUp technique offers several benefits in the realm of dataset handling, primarily by mitigating the imbalanced distribution between the background and vessel classes and promoting simultaneous multi-view learning to introduce additional variety.

\begin{figure}[ht!]
\centering
\begin{subfigure}{.495\textwidth}
  \centering
  \includegraphics[width=.8\linewidth]{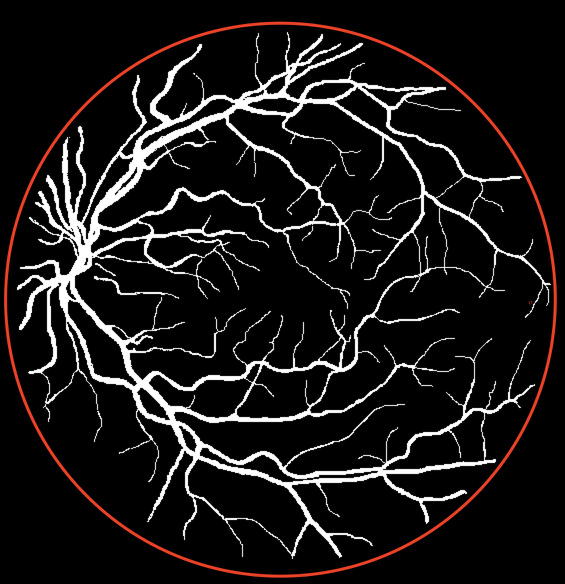}
  \caption{Label}
\end{subfigure}%
\hfill
\begin{subfigure}{.495\textwidth}
  \centering
  \includegraphics[width=.8\linewidth]{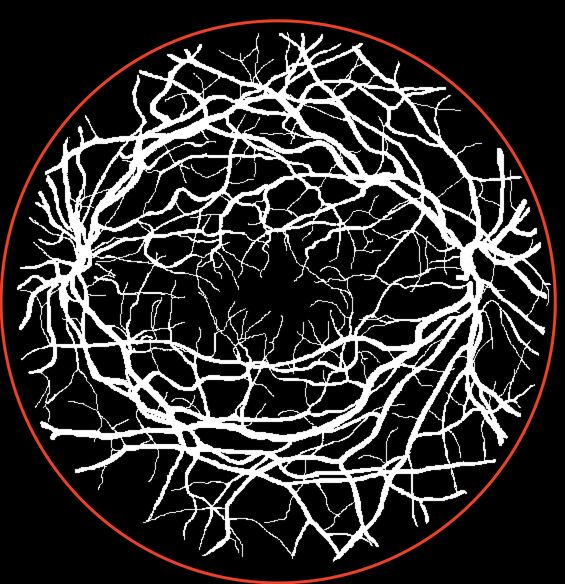}
  \caption{MixUped Label}
\end{subfigure}
\hfill
\medskip 

\begin{subfigure}{.495\textwidth}
  \centering
  \includegraphics[width=.9\linewidth]{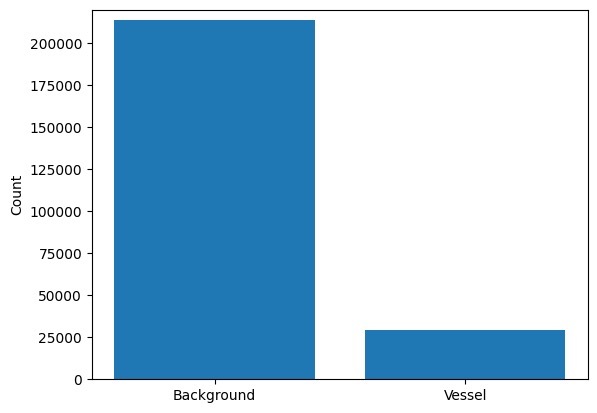}
  \caption{Histogram}
\end{subfigure}%
\hfill
\begin{subfigure}{.495\textwidth}
  \centering
  \includegraphics[width=.9\linewidth]{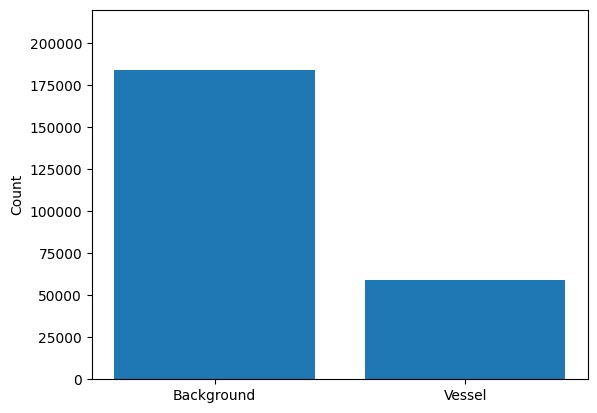}
  \caption{MixUped Histogram}
\end{subfigure}
\hfill
\caption{Sample imbalance for vessel class and background class inside the region annotated by the red cycle.}
\label{fig:mixuphist}
\end{figure}

\BE{Mitigating unbalanced distribution} In the field of medical image segmentation, the background class predominantly overshadows the vessel class. This disproportionate representation can induce the model to misclassify pixels belonging to vessels as background, especially those located near the decision boundary, culminating in sub-optimal performance. A plausible solution to this imbalance is attributing more significant weight to the vessel class. For instance, implementing Focal Loss \cite{lin2017focal} can effectively counterbalance the skewed distribution between background and vessel classes, as demonstrated in Figure \ref{fig:mixuphist}.

\BE{Simultaneous multi-view learning} The deployment of the MixUp technique enables the model to process multiple images concurrently, accelerating the training process by introducing additional data. While introducing more significant variance, this method also signals to the model that the summation operation can preserve the semantic information, thereby improving the overall learning efficacy.

Nonetheless, the MixUp technique is not devoid of limitations. The summation operation intrinsic to MixUp tends to inflate the pixel intensity beyond the ordinary. This could engender disparity between the training and testing data due to the increased representation of vessel structures. Consequently, while the model becomes more attuned to minor vessels, it may also become more noise-resistant. 




\subsection{MixUp by Label}
\begin{figure}[h!]
\centering
\includegraphics[width=\textwidth]{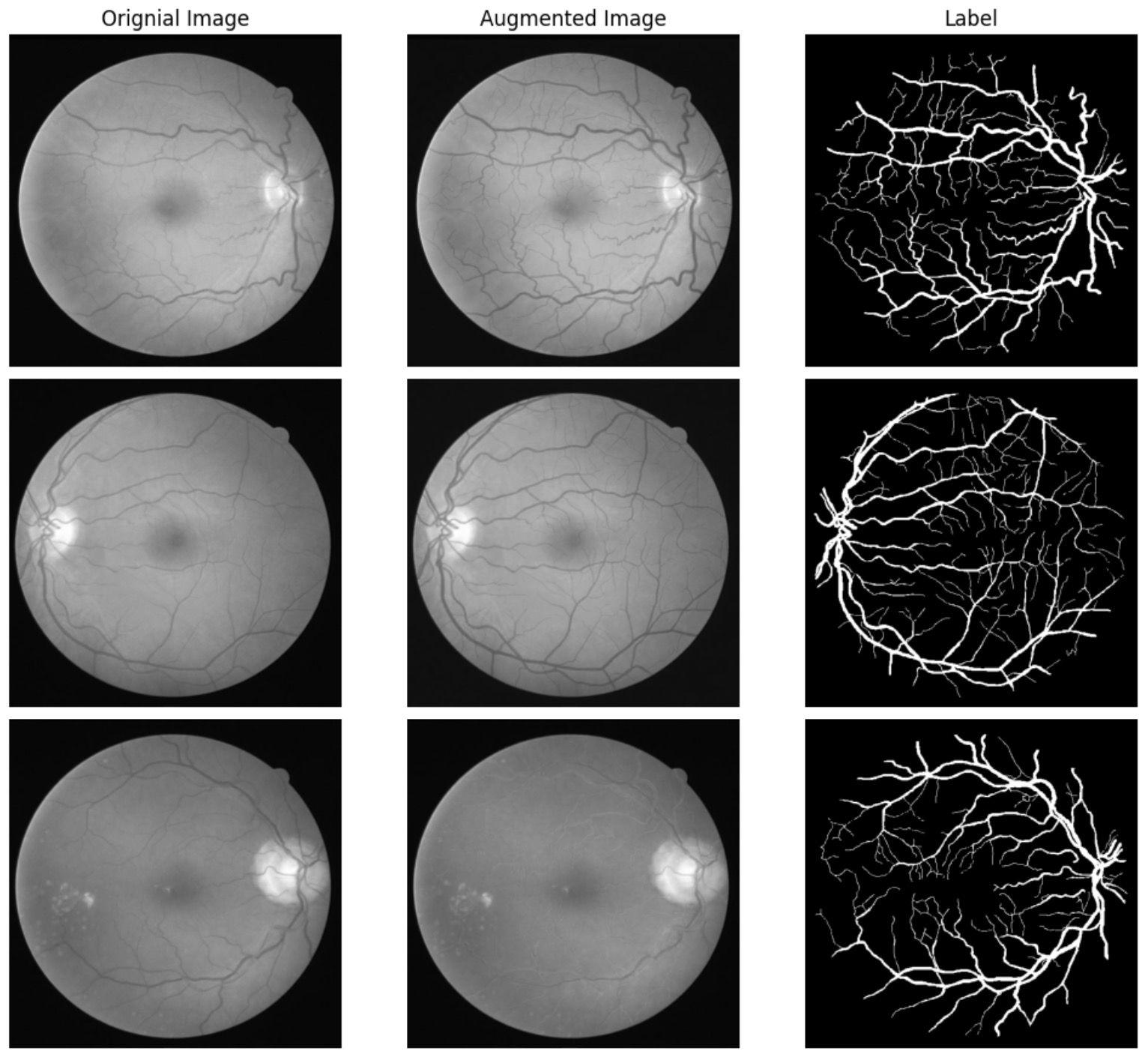}
\caption{Comparison of original images with those augmented using MixUp by summing up the label. A positive $\alpha$ value yields clearer vessel structures in the first and second scenarios. Conversely, the third scenario employs a negative $\alpha$ value, simulating real-world noise and yielding reduced visual discernibility. Combining MixUp by the label with other augmentation techniques introduces greater variance, effectively addressing the limitations imposed by dataset size.}
\label{fig:mixuplabel}
\end{figure}

Another variant of MixUp has been proposed that leverages a hyperparameter, denoted as $\alpha$, to manage the intensity of the overlapping label with the original image. In practice, $\alpha$ follows a standard normal distribution $N(0, 1)$. Positive $\alpha$ values accentuate the vessel structure, mirroring the effect of CLAHE, while negative $\alpha$ values diminish visual discernibility, simulating real-world noise. This visual representation is illustrated in Figure  \ref{fig:mixuplabel}. This variant of MixUp can be used in tandem with other data augmentation techniques, such as colour transformation, to introduce additional variants and bolster the model's robustness. 


\chapter{Experiments and Results}
\label{Chapter4}

\section{Settings}
\subsection{Datasets}
The experiments are tested on two benchmark datasets. DRIVE and CHASE-DB1 with the split ratio of 8:2. Where 80\% images are utilised as training data and 20\% as the validation data.

\begin{figure}[h!]
\centering
\begin{subfigure}{.19\textwidth}
  \centering
  \includegraphics[width=.95\linewidth]{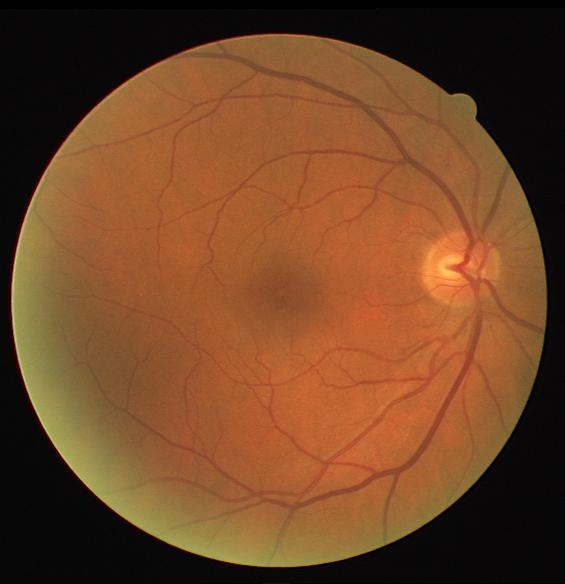}
\end{subfigure}%
\hfill
\begin{subfigure}{.19\textwidth}
  \centering
  \includegraphics[width=.95\linewidth]{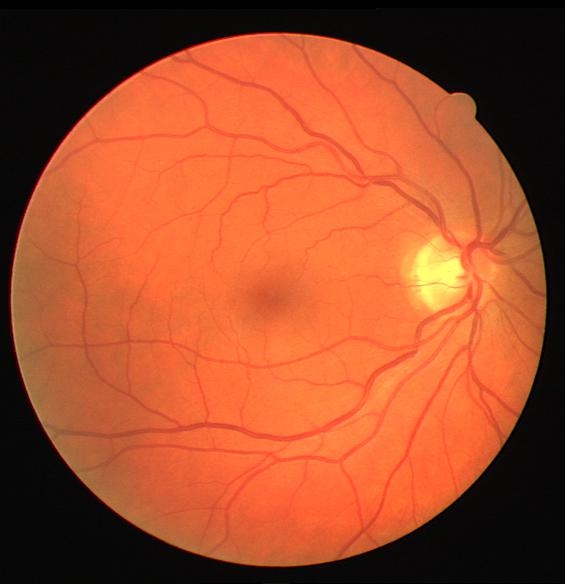}

\end{subfigure}%
\hfill
\begin{subfigure}{.19\textwidth}
  \centering
  \includegraphics[width=.95\linewidth]{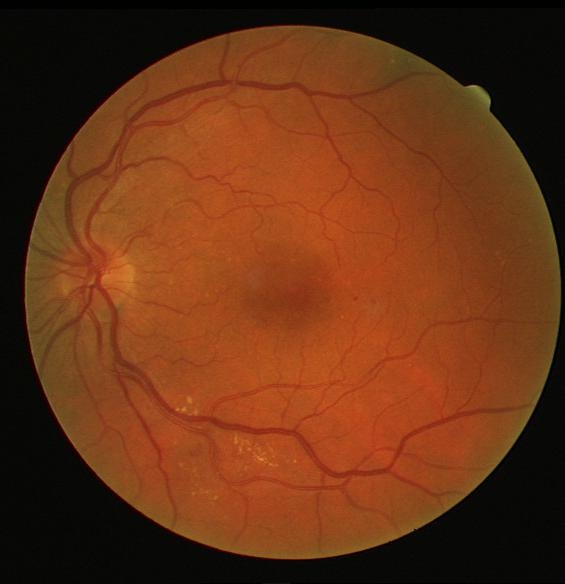}

\end{subfigure}
\hfill
\begin{subfigure}{.19\textwidth}
  \centering
  \includegraphics[width=.95\linewidth]{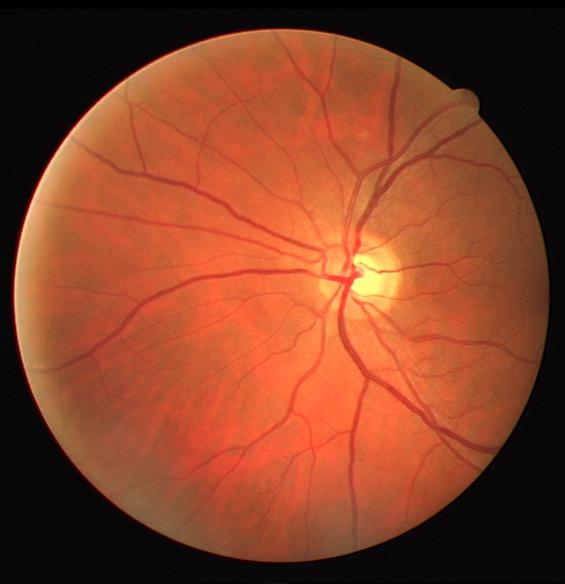}

\end{subfigure}%
\hfill
\begin{subfigure}{.19\textwidth}
  \centering
  \includegraphics[width=.95\linewidth]{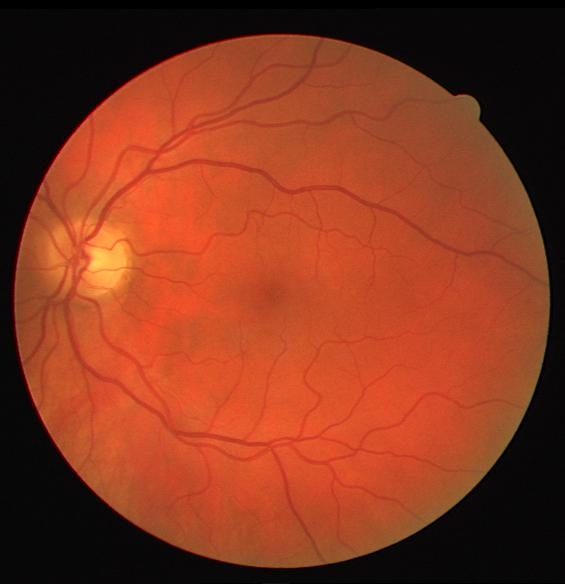}

\end{subfigure}%

\medskip 

\centering
\begin{subfigure}{.19\textwidth}
  \centering
  \includegraphics[width=.95\linewidth]{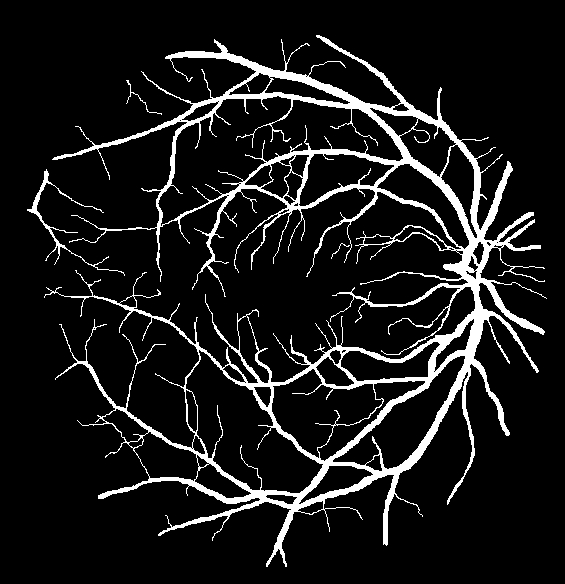}
\end{subfigure}%
\hfill
\begin{subfigure}{.19\textwidth}
  \centering
  \includegraphics[width=.95\linewidth]{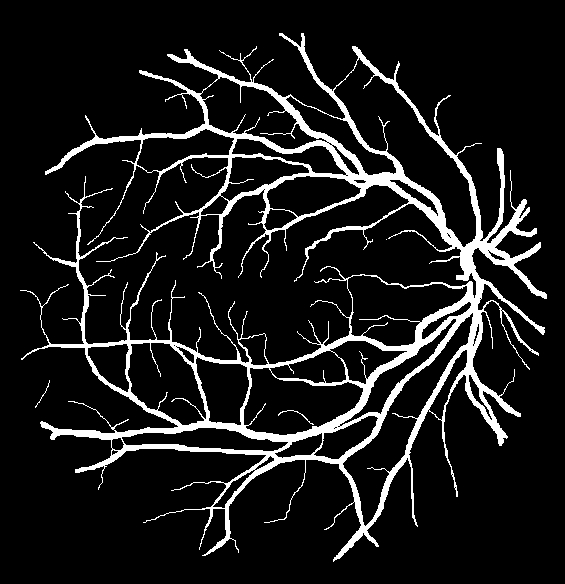}

\end{subfigure}%
\hfill
\begin{subfigure}{.19\textwidth}
  \centering
  \includegraphics[width=.95\linewidth]{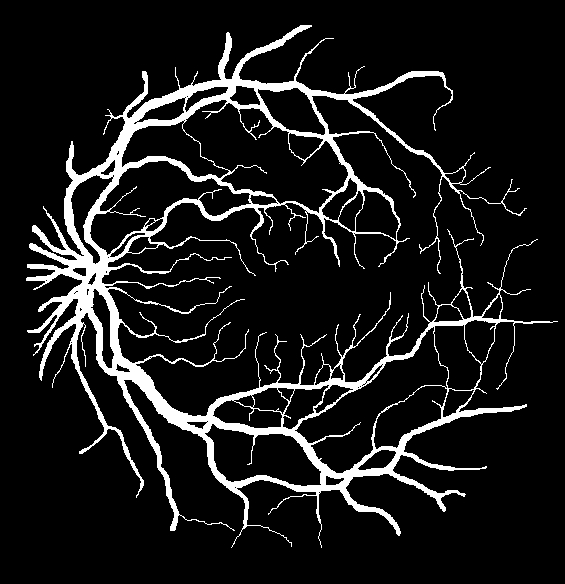}

\end{subfigure}
\hfill
\begin{subfigure}{.19\textwidth}
  \centering
  \includegraphics[width=.95\linewidth]{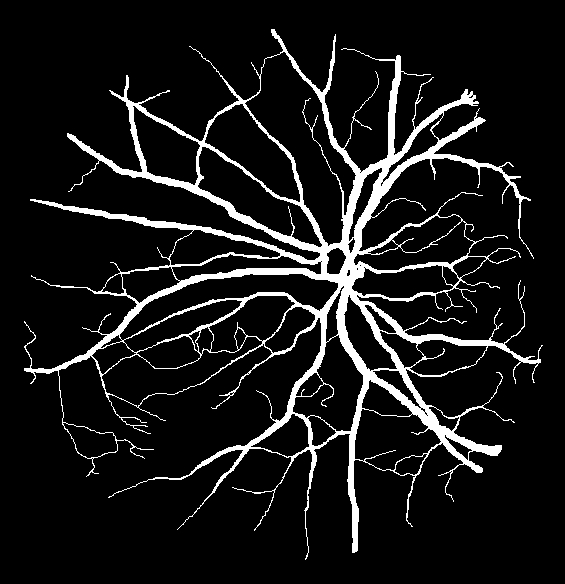}

\end{subfigure}%
\hfill
\begin{subfigure}{.19\textwidth}
  \centering
  \includegraphics[width=.95\linewidth]{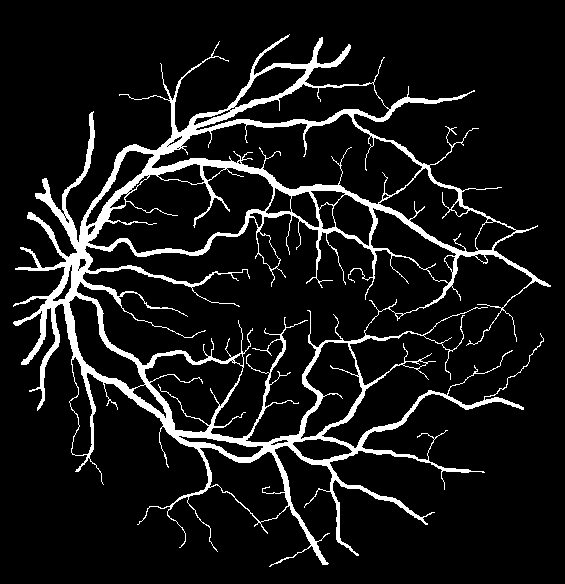}

\end{subfigure}%
\caption{Images and labels in DRIVE dataset.}
\label{fig:drivedata}
\end{figure}

\BE{DRIVE} The DRIVE \cite{staal2004ridge} dataset comprises 40 RGB retinal images, of which 33 images depict healthy eyes, and 7 display mild early diabetic retinopathy. Each image has a resolution of 565 $\times$ 584 $\times$ 3 pixels, shown in Figure \ref{fig:drivedata}.

\begin{figure}[h!]
\centering
\begin{subfigure}{.19\textwidth}
  \centering
  \includegraphics[width=.95\linewidth]{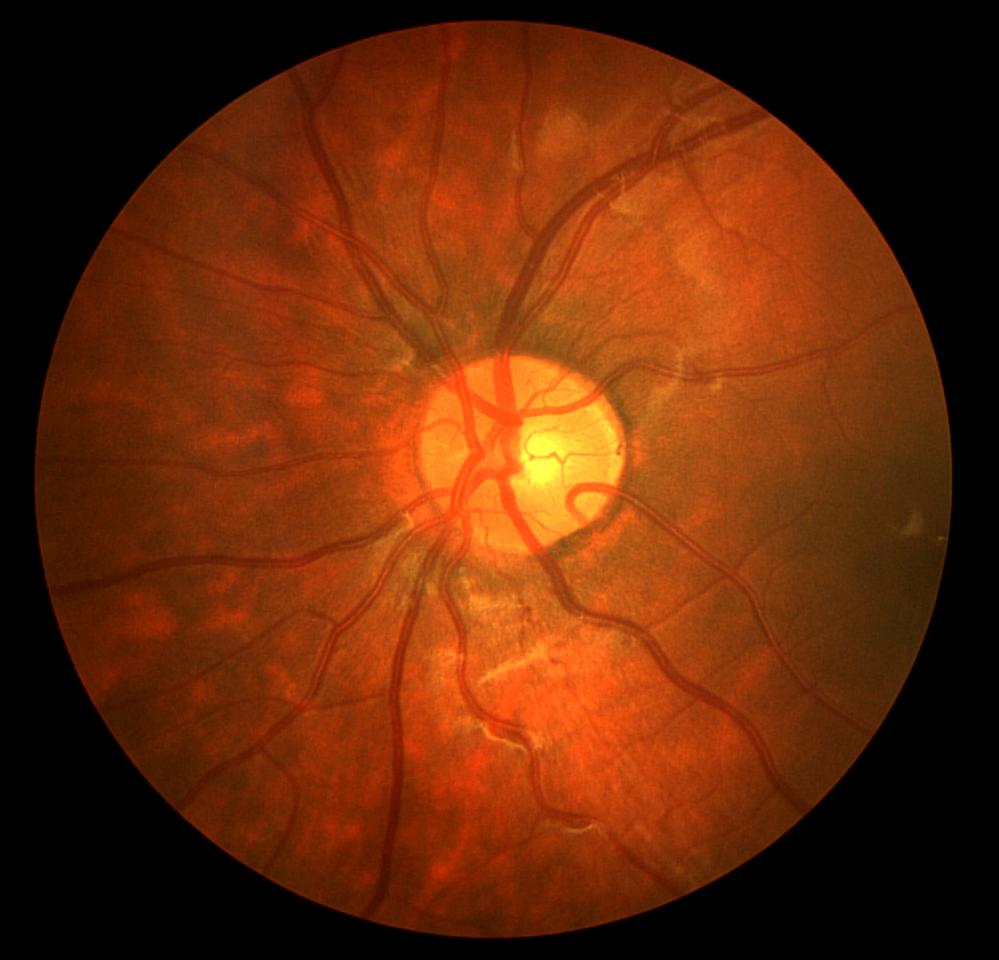}
\end{subfigure}%
\hfill
\begin{subfigure}{.19\textwidth}
  \centering
  \includegraphics[width=.95\linewidth]{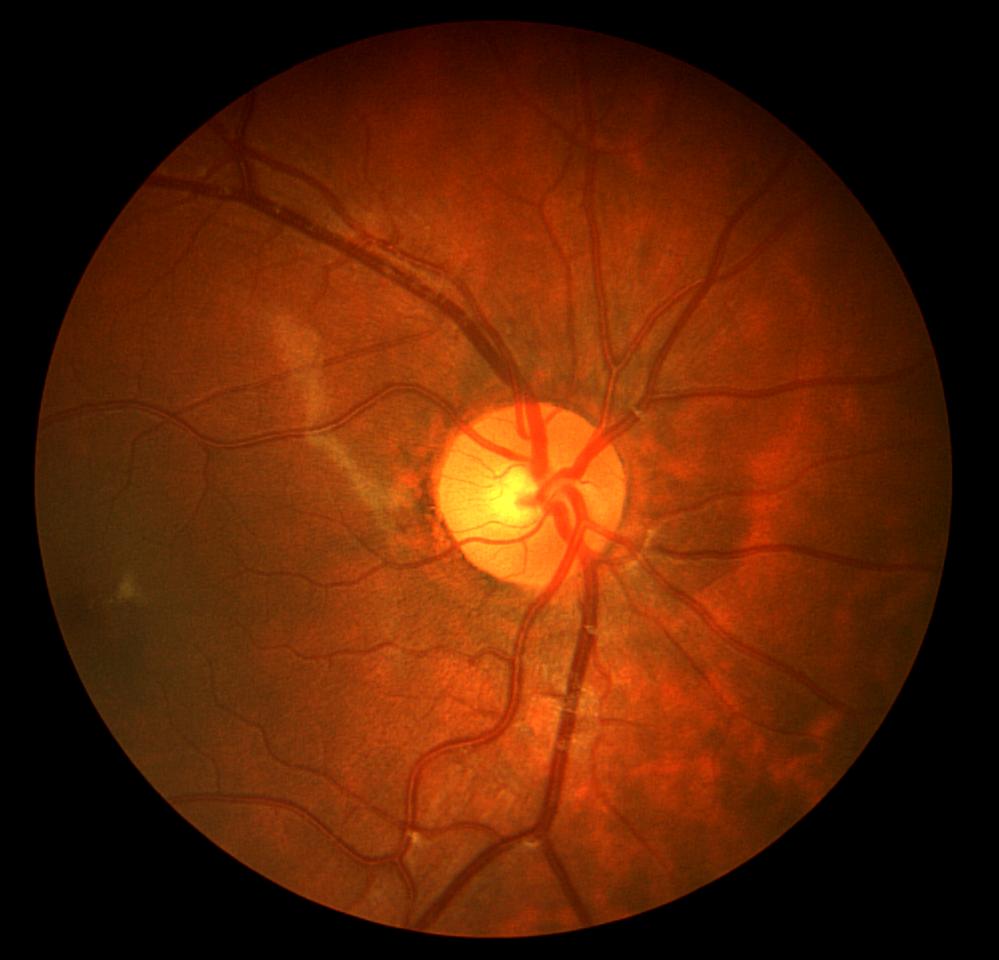}

\end{subfigure}%
\hfill
\begin{subfigure}{.19\textwidth}
  \centering
  \includegraphics[width=.95\linewidth]{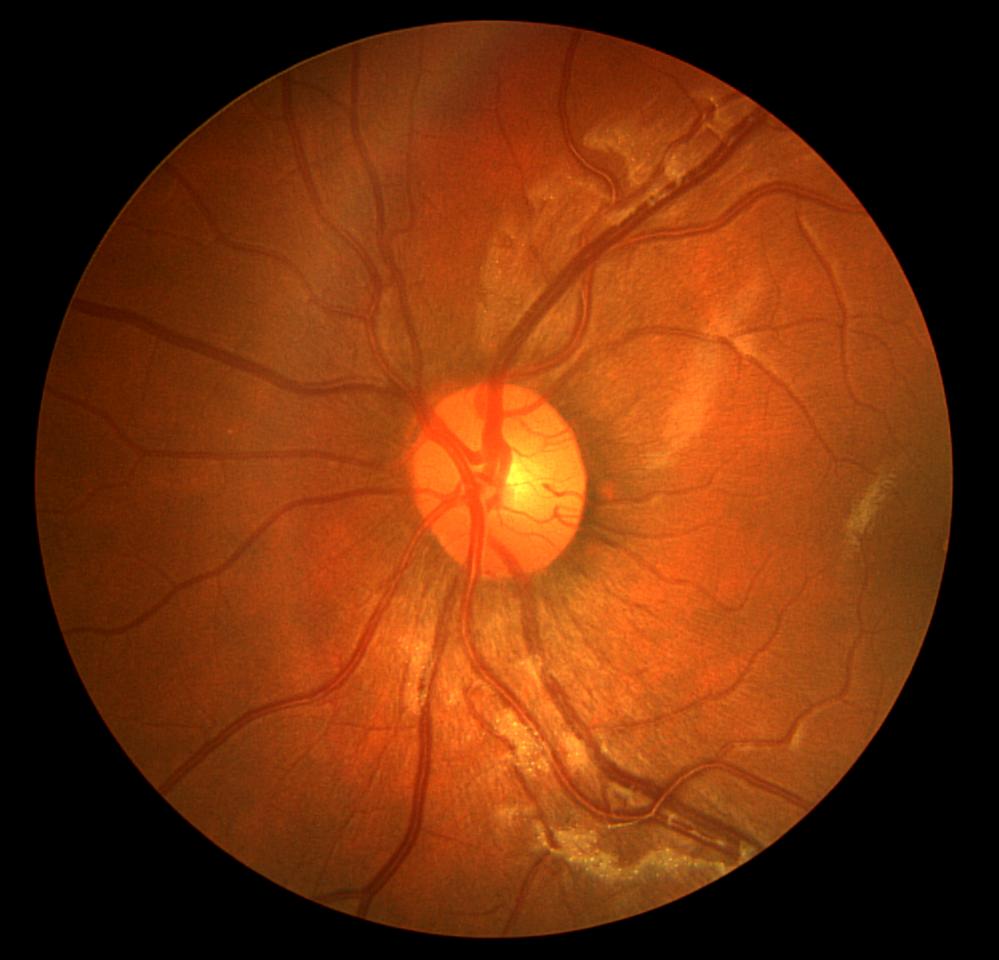}

\end{subfigure}
\hfill
\begin{subfigure}{.19\textwidth}
  \centering
  \includegraphics[width=.95\linewidth]{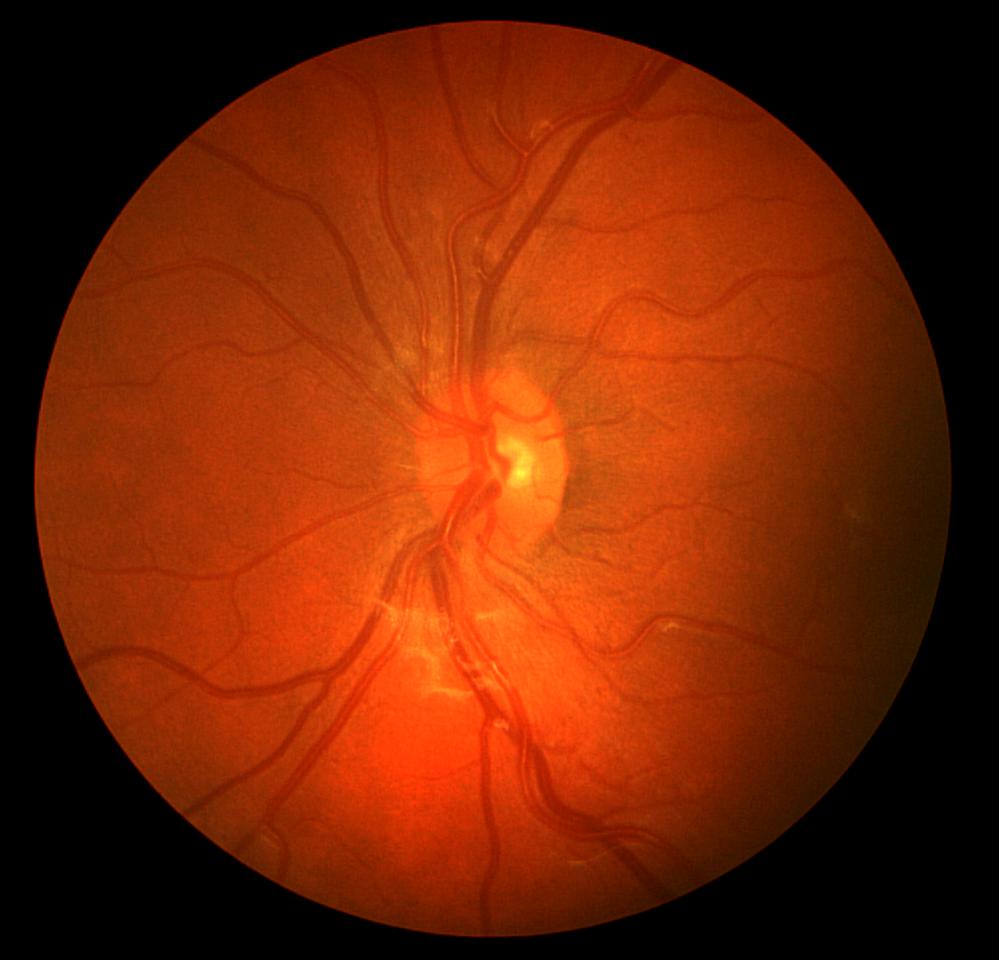}

\end{subfigure}%
\hfill
\begin{subfigure}{.19\textwidth}
  \centering
  \includegraphics[width=.95\linewidth]{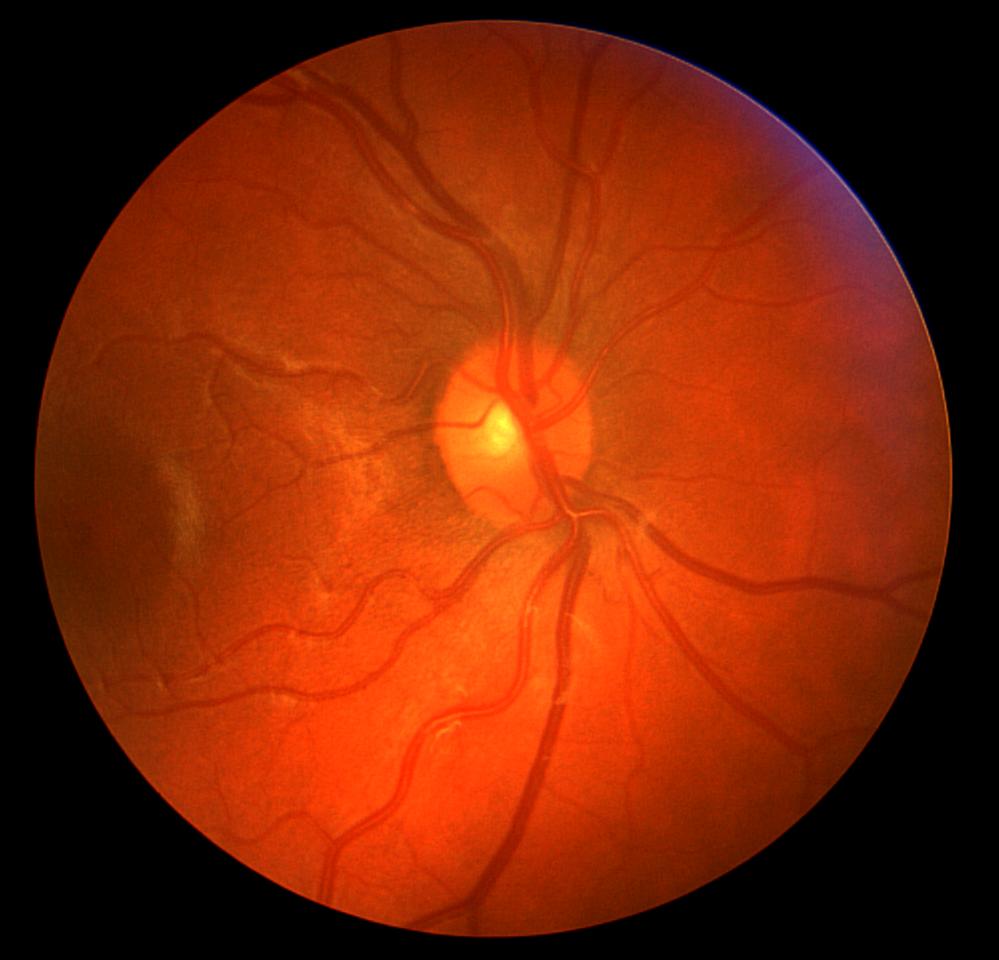}
\end{subfigure}%

\medskip 

\centering
\begin{subfigure}{.19\textwidth}
  \centering
  \includegraphics[width=.95\linewidth]{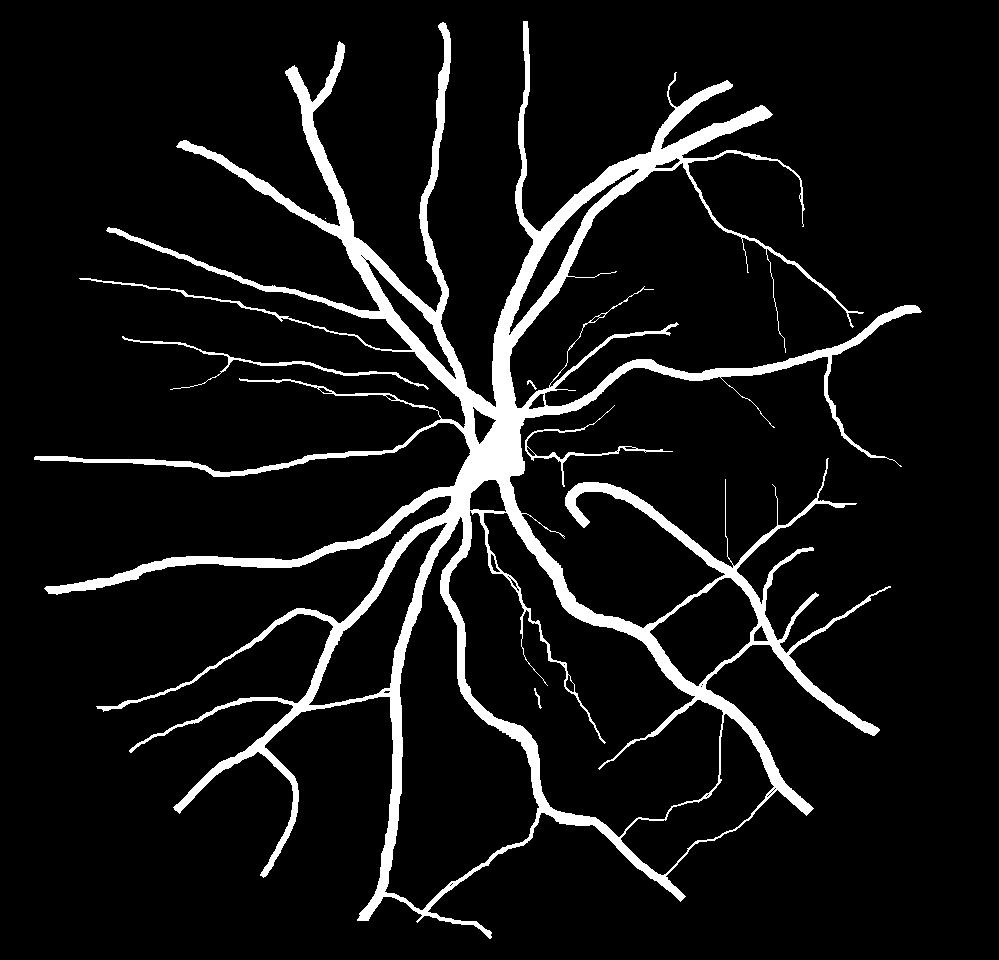}
\end{subfigure}%
\hfill
\begin{subfigure}{.19\textwidth}
  \centering
  \includegraphics[width=.95\linewidth]{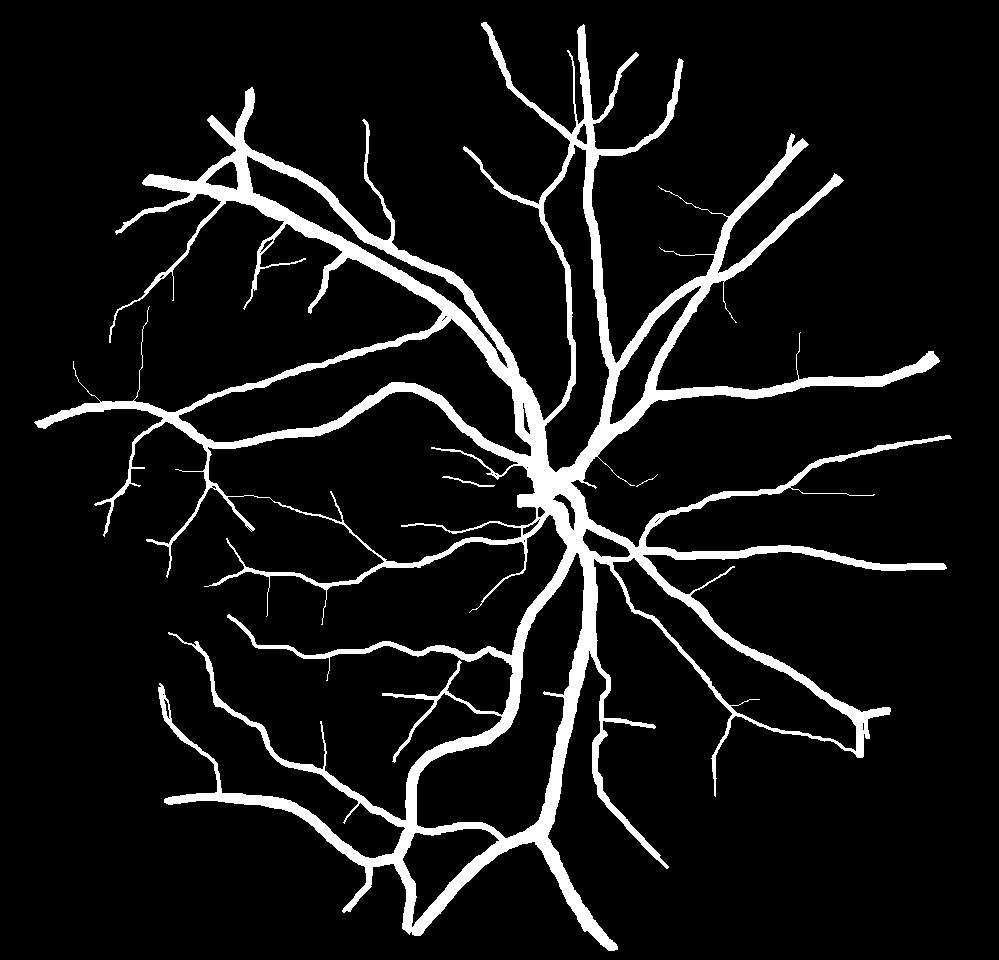}

\end{subfigure}%
\hfill
\begin{subfigure}{.19\textwidth}
  \centering
  \includegraphics[width=.95\linewidth]{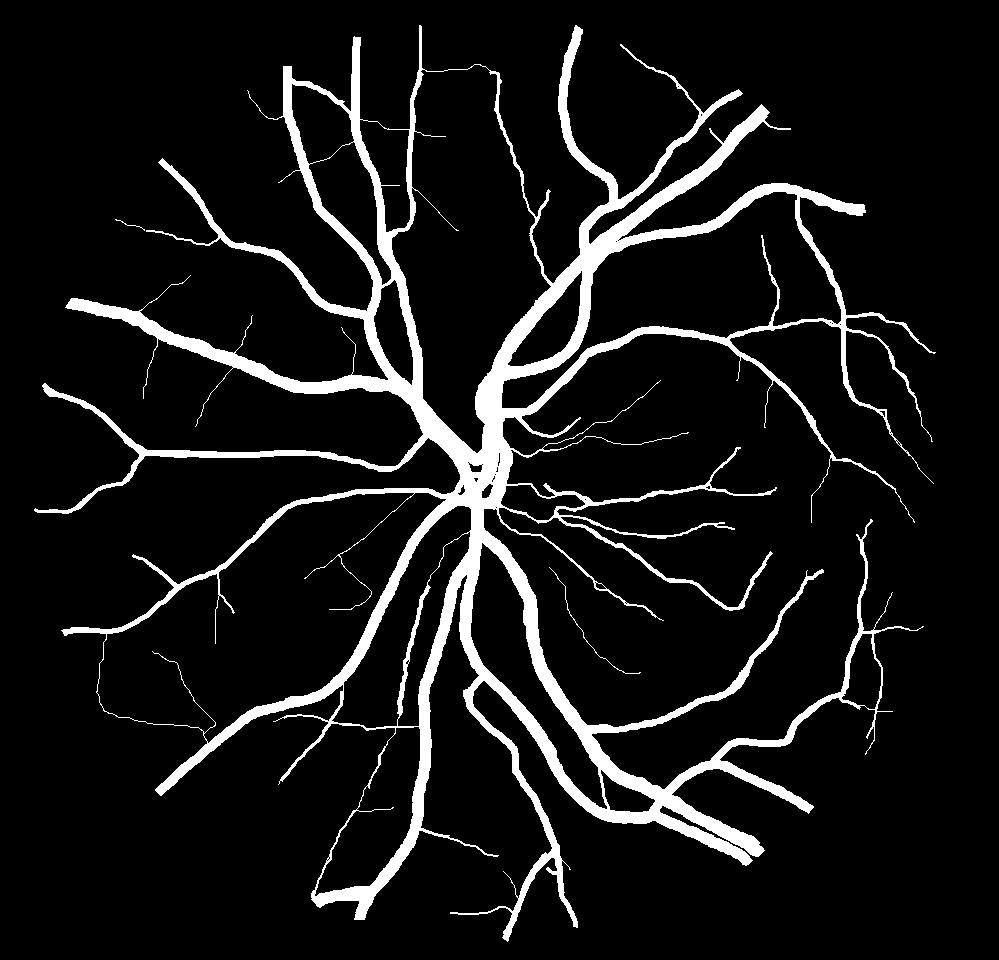}

\end{subfigure}
\hfill
\begin{subfigure}{.19\textwidth}
  \centering
  \includegraphics[width=.95\linewidth]{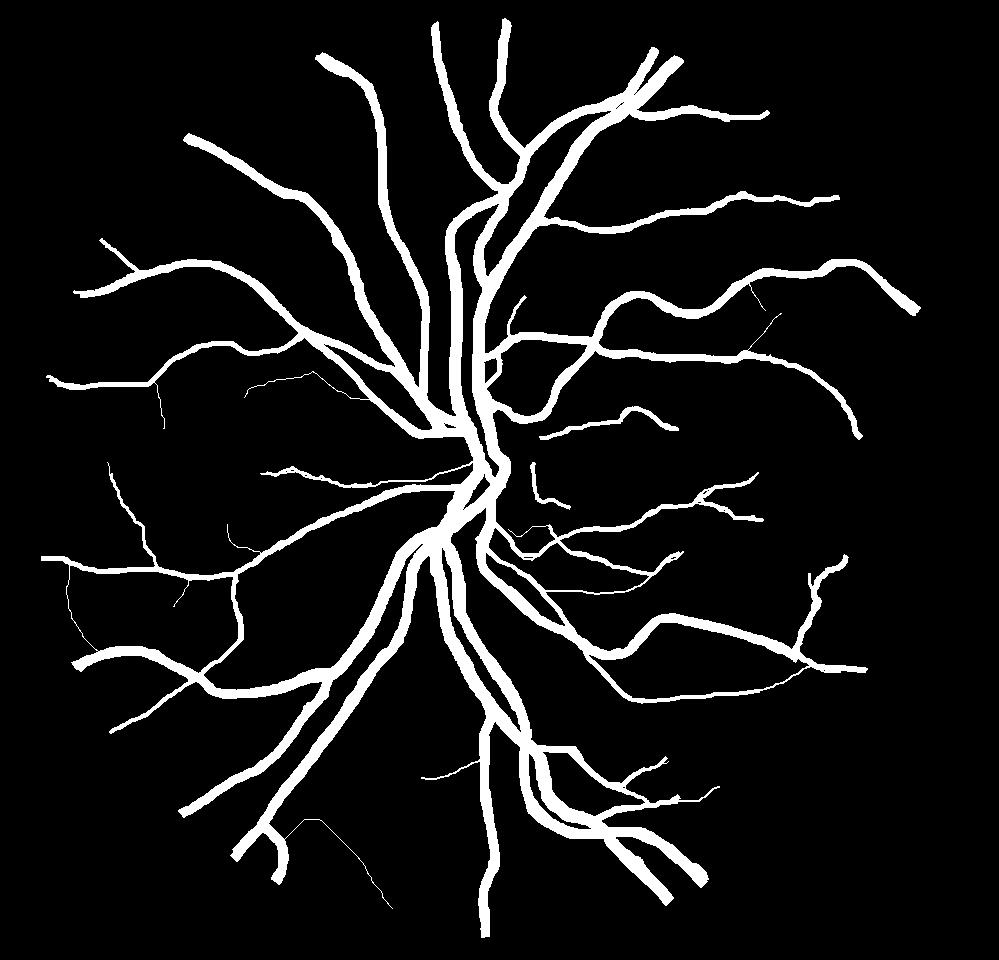}

\end{subfigure}%
\hfill
\begin{subfigure}{.19\textwidth}
  \centering
  \includegraphics[width=.95\linewidth]{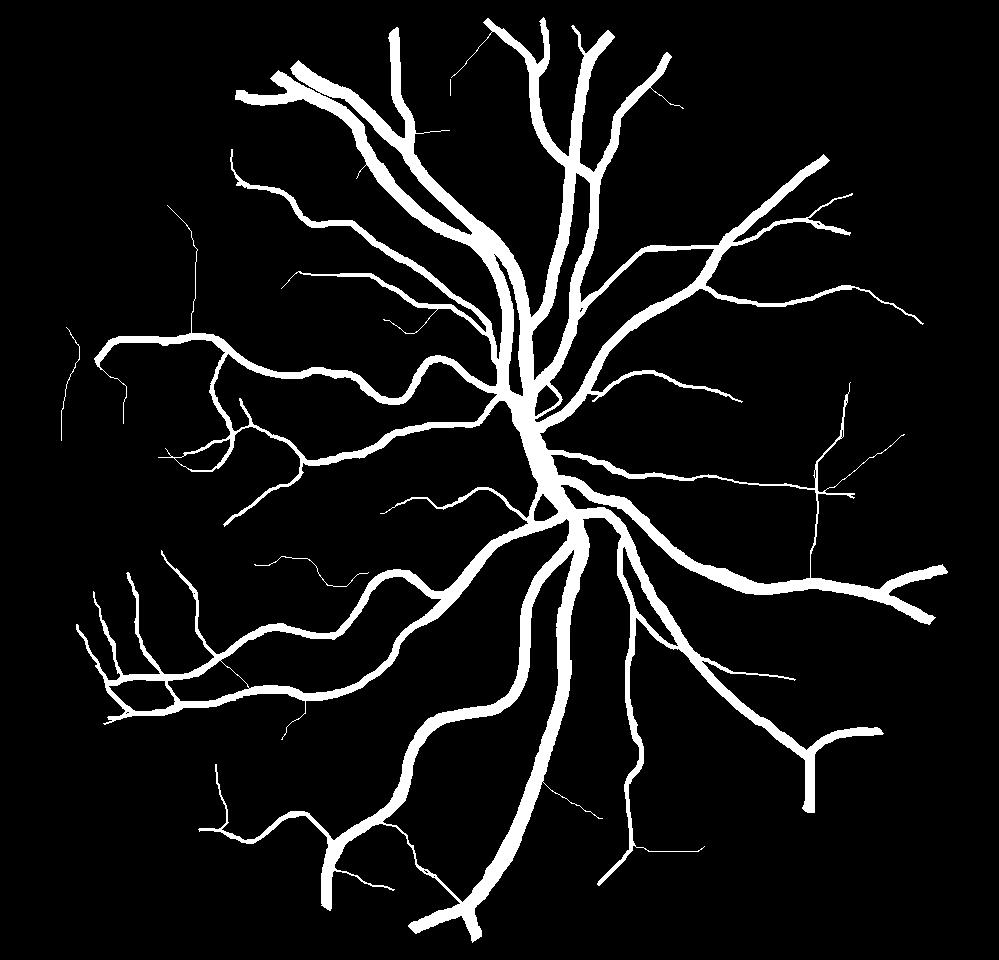}

\end{subfigure}%
\caption{Images and labels in CHASE-DB1 dataset.}
\label{fig:chasedbdata}
\end{figure}

\BE{CHASE-DB1} The CHASE-DB1 \cite{owen2009measuring} dataset includes 28 RGB children's retinal images, each with a resolution of 990 $\times$ 960 $\times$ 3 pixels. The first 20 images serve as training data, while the remaining 8 are designated for testing, shown in Figure \ref{fig:chasedbdata}.

\subsection{Metrics}
The evaluation metrics are defined to analyse and compare the performance of different models quantitatively for many aspects. This section introduces the standard metrics utilised in medical image segmentations in terms of True Positives (TP), True Negatives (TN), False Positives (FP) and False Negatives (FN).

\begin{itemize}
    \item \textbf{TP:} Pixel number of correctly predicted objects.
    \item \textbf{TN:} Pixel number of correctly predicted backgrounds.
    \item \textbf{FP:} Pixel number of incorrectly predicted objects.
    \item \textbf{FN:} Pixel number of incorrectly predicted backgrounds.
\end{itemize}

\BE{Pixel accuracy (ACC)} Pixel accuracy states the ratio of the sum number of the correctly predicted pixels with the number of total pixels of the image. It is well noticed that when there exists an imbalance of distribution between the ROI and background, the pixel accuracy could be unreliable in reflecting the model's performance. It is defined as:

\begin{equation}
Accuracy = \frac{TP + TN}{TP + TN + FP + FN}.
\end{equation}

\BE{Sensitivity (SEN)} Sensitivity (Recall) measures the ability to the disease area. Highly sensitive models can detect target areas (e.g. lesions, organs or tissues) more effectively, thus helping to reduce the likelihood of missed detections. In diagnosing and treating disease, reducing the missed detection rate is essential for the timely detection and resolution of problems. In some scenarios, there is a low tolerance for missed detections, such as cancer screening or detection of other serious diseases. In these cases, sensitivity becomes a key metric for assessing the model's performance. The mathematical equation is:

\begin{equation}
Sensitivity = Recall = \frac{TP}{TP + FN}.
\end{equation}

\BE{Specificity (SEP)} Specificity is an important metric in the medical domain to assess the model's ability to distinguish between target and non-target areas correctly. High specificity represents a low false positive rate. It could be calculated by:

\begin{equation}
Specificity = \frac{TN}{TN + FP}.
\end{equation}

\BE{Precision (PRE)} Precision measures the model's accuracy in identifying and segmenting anatomical structures or lesion areas of interest. Accuracy indicates the proportion of the target region identified by the model that is the target region, which relates to the degree of false positives. In medical image analysis, high accuracy is critical to avoid misdiagnosis and misleading treatment decisions. For example, when detecting a tumour or lesion, and want the model to accurately identify the actual abnormal area rather than mistaking normal tissue for an abnormality:

\begin{equation}
Precision = \frac{TP}{TP + FP}.
\end{equation}

\BE{F1 score (F1)} F1 score is an overall metric balancing Recall and Precision. In medical image analysis, data are often unbalanced; for example, lesion areas may be relatively small. Therefore, reducing false positives (high precision) and ensuring that all regions of interest are covered (high recall) are critical for diagnostic and treatment decisions. In binary classification cases, the F1 score equals the Dice coefficient. The formula of the F1 score is stated by:

\begin{equation}
F1 = \frac{2 \times Precision \times Recall}{Precision + Recall} = Dice.
\end{equation}

\BE{Intersetion over union (IOU)} IOU is also as known as the Jaccard index, which calculates the intersection area between the prediction and the ground truth:

\begin{equation}
IOU = \frac{TP}{TP + FP + FN}.
\end{equation}

\subsection{Implementation Details}
The following parameters summarised in Table \ref{tab:bestexp} obtain the best result. Firstly, the interactions of different data augmentation will lead to higher complexity for hyperparameter tuning. We analysed the effect of the data augmentation, such as MixUp will be introduced in the ablation study part. The best result was obtained with the simple augmentation: rotation in the degree range of 0 and 360, auto contrast, and colour transformation in brightness, hue and saturation with a value of 0.2. Then, to reduce the computational complexity. Note that all the data are normalised, and the images are resized to $560 \times 560$ for the DRIVE dataset. As for the CHASE-DB1 dataset, the images are resized to $1000 \times 1000$. The grey images are utilised since it contains only one channel. Then, we leverage the BCE Loss with 2 views of the contrastive views. The experiments showed that less batch size would lead to better results. Therefore, we set the batch size as 1. The model is combined with the backbone and attention mechanism where we insert SE block into FR-UNet. The model is trained with 50 epochs by Adam optimizer learning rate of 1e-4 and weight decay of 1e-5. And the learning rate is gradually reduced by the Consine Annealing scheduler. The full training process is on the Colab with a free GPU.

\begin{table}[!h]
    \centering
    \resizebox{0.6 \textwidth}{!}{
    \begin{tabular}{|c||c|}
    \hline
    Parameters & Value \\
    \hline
    \centering Augmentation & Rotation, Contrast, Colour, Gery\\
    \hline
    \centering Loss & Binary Cross-Entropy Loss\\
    \hline
    \centering Contrastive Views & 2\\
    \hline
    \centering Batch Size &  1\\
    \hline
    \centering Backbone & FR-UNet\\
    \hline
    \centering Attention Block & SE Block \\
    \hline
    \centering Epochs & 70 \\
    \hline
    \centering Optimizer & Adam  \\
    \hline
    \centering Learning Rate & 1e-4\\
    \hline
    \centering Weight Decay & 1e-5\\
    \hline
    \centering LR scheduler & Cosine Annealing\\
    \hline
    \centering Device & Colab Free GPU (T4)\\
    \hline
    \end{tabular}}
    \caption{Parameters settings.}
    \label{tab:bestexp}
\end{table}

\section{Main Results}
\subsection{Segmentation Visualisation}
To better understand the model's performance, the predicted segmentation results are plotted for comparison with the ground truth. Each graph contains four images. The most left is the original input image, followed by the predicted image and the predicted map, which contains the probability value from 0 to 1 for each pixel. The continuous value could better calculate the gradients. Higher probability stands for a higher confidence level. The predicted image is calculated by a threshold filter containing only two labels: the background and label classes. The threshold is set at 0.5, meaning a value higher than 0.5 is denoted with the vessel object. The final image is the ground truth.

\begin{figure}[h!]
  \centering
  \includegraphics[scale=0.23]{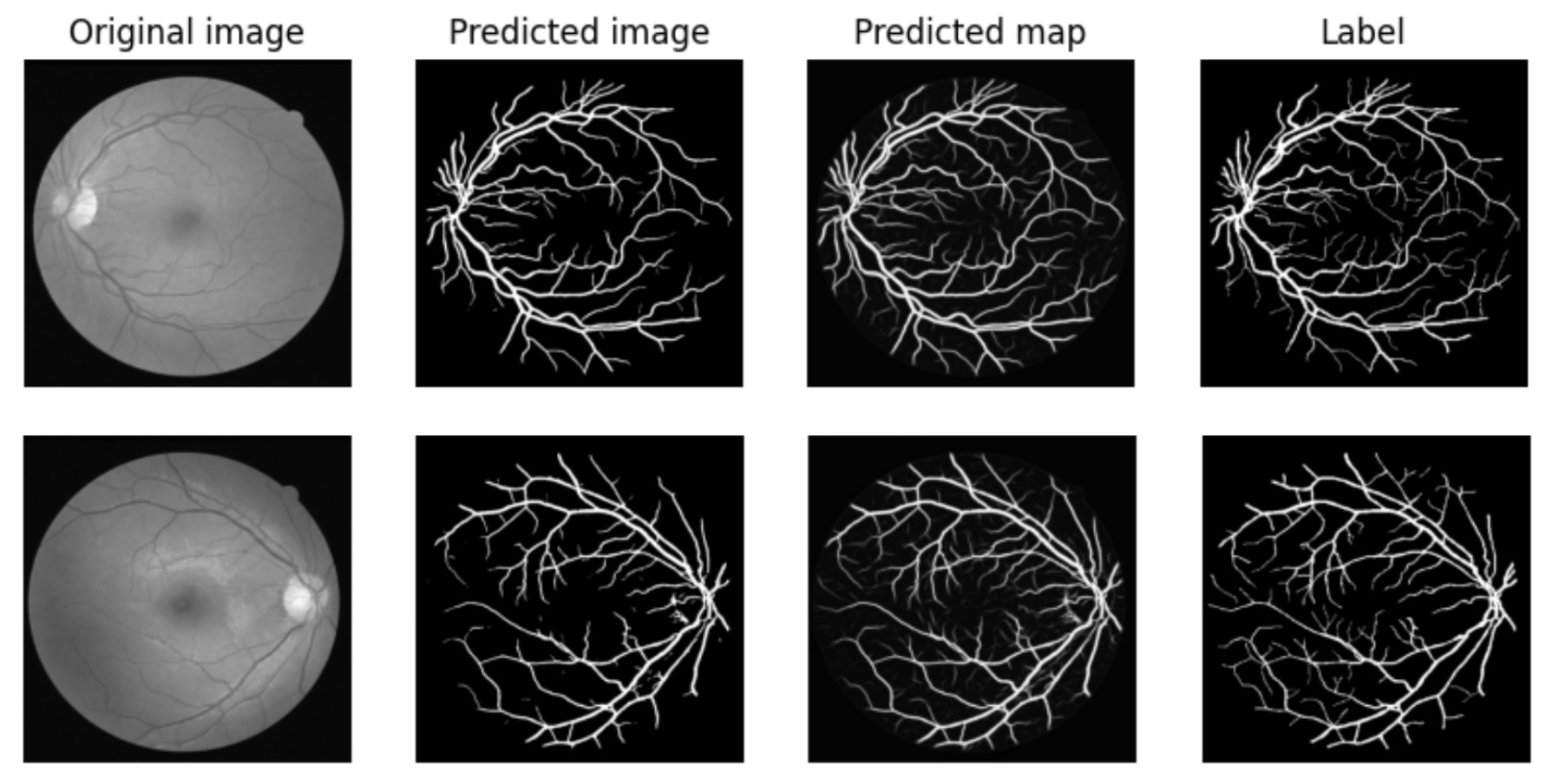}
  \caption{Segmentation result of DRIVE dataset.}
  \label{fig:driveseg}
\end{figure}

\BE{DRIVE} The segmented results in the DRIVE dataset are presented in Figure \ref{fig:driveseg}. The proposed method could well learn the general vessel structure. It could be found that the predicted image is pretty close to the ground truth. However, it could not be well detected with the tiny vessel since the probability value shown by the predicted map with a lower intensity and low probability in detailed vessels. And the predicted vessels are generally thicker than the ground truth label. One hypothesis is that since the trade-off between the performance and the computational cost. The model does not have very large parameters for the detailed information. Moreover, the kernel size may also affect the performance. The larger kernel size will result in a more rough prediction. Overall, the predictions by AUtO have relatively good quality.

\BE{CHASE-DB1} Comparing with DRIVE dataset, the CHASE-DB1 have two major difference. The first one contains a larger resolution, and the other one contains thicker and less tiny vessels. Figure \ref{fig:chasedbdata} shows the predicted result of CHASE-DB1. The proposed method could also predict the larger vessel correctly. However, it still suffers from the tiny details.

\begin{figure}[h!]
  \centering
  \includegraphics[scale=0.23]{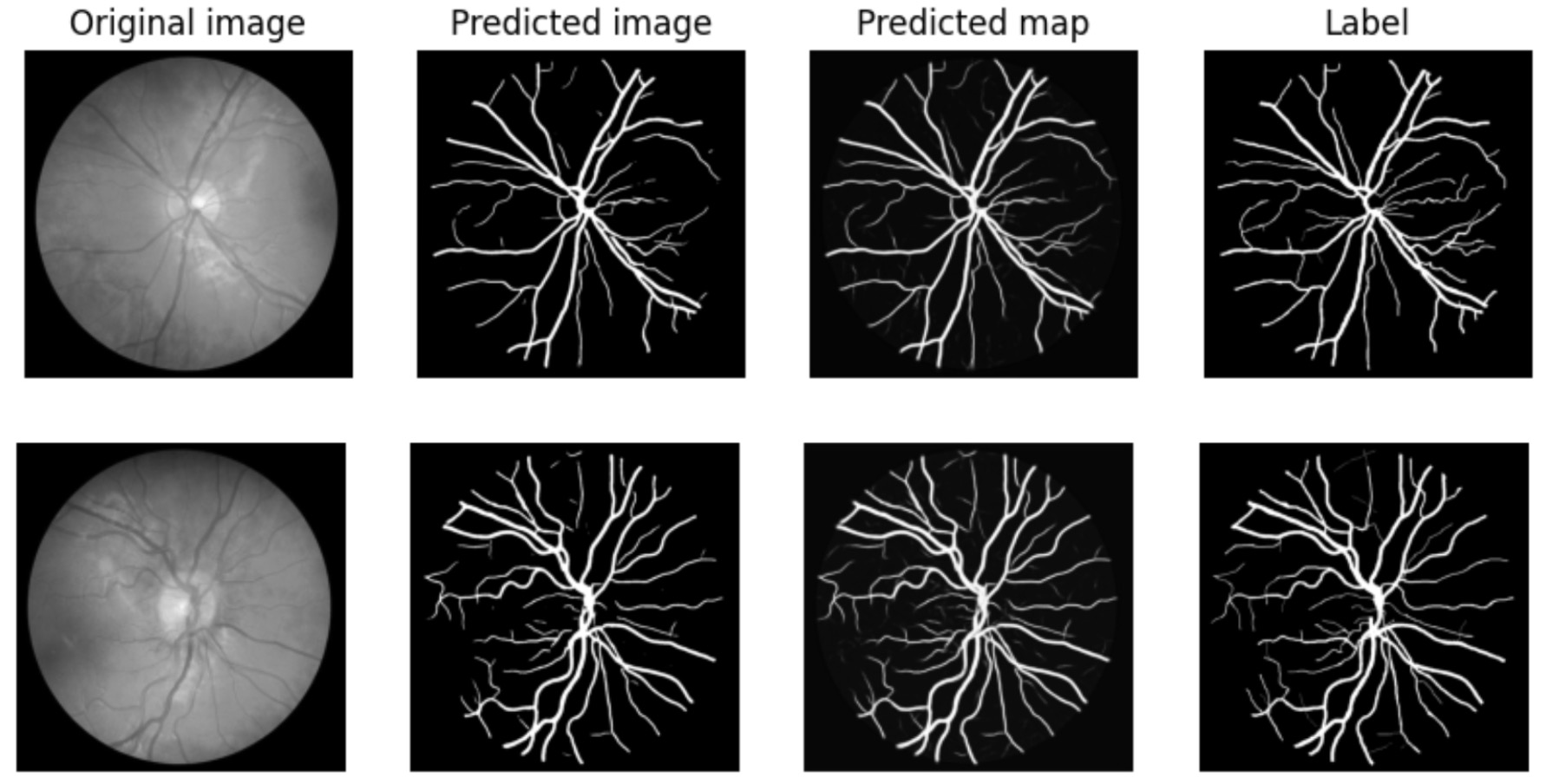}
  \caption{Segmentation result of CHASE-DB1 dataset.}
  \label{fig:chasedbseg}
\end{figure}

\subsection{Quantitative Comparison Analysis}
The performance of various methods is presented in Table \ref{tab:mainresult}. The results demonstrate that UNet family networks generally have fewer training parameters than attention-based and GAN-based networks. With the highest number of training parameters, SGL owes this to its semantic training approach involving multiple sub-predictors. It achieves the highest sensitivity score on the DRIVE dataset (0.8380) and the second-highest on the CHASE-DB1 dataset (0.8690), trailing the benchmark score of 0.8798 by FR-UNet by a margin of 0.0108. Embedding methods like SGL permit compact models to focus on intricate local features rather than global ones, which significantly bolsters sensitivity scores.

The model with the second-largest parameter count is RV-GAN, exhibiting remarkable results with the highest accuracy, specificity, and F1 score on the DRIVE dataset and the highest F1 score on the CHASE-DB1 dataset. Despite its success, training a GAN-based neural network is resource-intensive, requiring 40 training hours \cite{uysal2021exploring}.

FR-UNet, with only 7.37 million parameters, achieves a benchmark AUC score on the DRIVE dataset and the highest sensitivity and IOU scores on the CHASE-DB1 dataset. As a lightweight model with high performance, FR-UNet's efficacy stems from its HR-Net backbone.

Our proposed method, AUtO, is based on the FR-UNet backbone and introduces only 0.03 million additional parameters, ensuring its lightweight nature. AUtO achieved the highest IOU and SPE scores and the second-highest F1 score in the CHASE-DB1 dataset, with only 70 training epochs. Increasing the training epochs to 160 could potentially improve. The F1 and IOU scores to approximately 0.85 and 0.75, respectively. Although training on grey images reduces computational costs, it sacrifices colour information, which may be valuable for decision-making. Therefore, future work will focus on training the model with RGB images and implementing more data augmentation techniques.

However, AUtO's performance in the DRIVE dataset is not as impressive. This is possibly due to using the same hyperparameters as the CHASE-DB1 dataset, indicating the need for additional fine-tuning. Another hypothesis is that the DRIVE dataset's complex vessel structure and noise may lead to overfitting by the attention mechanism. Future experiments will focus on these hypotheses and implement additional regularization.

Moreover, Table \ref{tab:time} presents the efficiency of the models, highlighting that AUtO-Net requires only 0.5 training hours. This is a staggering 20 times faster than FR-UNet and 80 times faster than RV-GAN. Two primary factors contribute to this efficiency. Firstly, compared to RV-GAN, both AUtO and FR-UNet have fewer training parameters to adjust. While FR-UNet employs patch augmentation, dividing the input images into approximately 10 thousand $48 \times 48$ patches, AUtO opts for a more direct approach. Despite the enhancement in model performance, patch operation significantly dents training efficiency. In contrast, AUtO trains directly on raw images using simple data augmentations, thereby reducing the training time to a mere 0.5 hours.

In conclusion, the results clearly demonstrate the performance of the proposed AUtO method in relation to benchmark models. While GAN and knowledge-based methods deliver impressive results, they involve more extensive training parameters. Conversely, FR-UNet and AUtO maintain fewer training parameters without compromising performance. The AUtO framework boasts the most streamlined training procedure, taking only 0.5 hours on the freely available Colab GPU.

\begin{landscape}
\begin{table}[!h]
    \centering
    \resizebox{\linewidth}{!}{
    \begin{tabular}{|c |c||c|c|c|c|c|c||c|c|c|c|c|c||}
    \hline
     &  &\multicolumn{6}{c||}{ DRIVE} & \multicolumn{6}{c||}{ CHASEDB1}\\
    \hline
     Methods & Params (M) & ACC & SEN & SPE & AUC & F1 & IOU &  ACC & SEN & SPE & AUC & F1 & IOU\\
    \hline
    
    UNet \cite{ronneberger2015u} & 7.76 &0.9678 & 0.8057 & 0.9833 & 0.9825 & 0.8141 & 0.6864 & 0.9743 &0.7650 
    & 0.9884& 0.9836 &0.7898 & 0.6526\\
    \hline

    UNet++ \cite{zhou2018unet++} & 9.05 & 0.9679 & 0.7891 & 0.9850 & 0.9825 & 0.8114 & 0.6827 & 0.9739 & 0.8357 & 0.9832 & 0.9881 & 0.8015 & 0.6688 \\
    \hline

    Attention-UNet \cite{oktay2018attention} & 8.73 & 0.9662 & 0.7906 & 0.9831 & 0.9774 & 0.8039 & 0.6721 & 0.9730 & 0.8384 & 0.9820 & 0.9848 & 0.7964 & 0.6617 \\
    \hline

    HR-Net \cite{pohlen2017full} & 9.64 & 0.9704 & 0.8040 & 0.9864 & 0.9869 & 0.8265 & 0.7043 & 0.9758 & 0.8443 & 0.9847 & 0.9902 & 0.8148 & 0.6875\\
    \hline
    
    CS-Net \cite{mou2019cs} & 8.40 & 0.9632 & 0.8170 & 0.9854 & 0.9798 & 0.8039 & 0.7017 & 0.9742 & 0.8400 & 0.9832 & 0.9881 & 0.8042 & 0.6725 \\
    \hline

    AG-Net \cite{zhang2020attention} & - & 0.9692 & 0.8100 & 0.9848 & 0.9856 & - & 0.6965 &  0.9743 & 0.8186 & 0.9848 & 0.9863 & - & 0.6669\\
    \hline

    RVSeg-Net \cite{wang2020rvseg} & \textbf{5.20} & 0.9681 & 0.8107 & 0.9845 & 0.9817 & - & - & 0.9726 & 0.8069 & 0.9836 & 0.9833 & - & -\\
    \hline

    SCS-Net \cite{wu2021scs} & - & 0.9697 & 0.8289 & 0.9838 & 0.9837 & - & - & 0.9744 & 0.8365 & 0.9839 & 0.9867 & - & -\\
    \hline

    VSSC-Net \cite{samuel2021vssc} & 8.05 & 0.9627 & 0.7827 & 0.9821 & 0.9789 & - & - & 0.9633 & 0.7233 & 0.9865 & 0.9706 & - & -\\
    \hline

    MS-Net \cite{wu2018multiscale}& - & 0.9567 & 0.7844 & 0.9819 & 0.9807 & - & - &  0.9637 & 0.7538 & 0.9847 & 0.9825 & - & -\\
    \hline

    Vessel-Net \cite{wu2019vessel} & - & 0.9578 & 0.8038 & 0.9802  & 0.9821 & - & - & 96.61 & 0.8132 & 0.9814 & 0.9661 & - & -\\
    \hline

    
    SGL \cite{zhou2021study} & 15.53 & 0.9705 & \textbf{0.8380} & 0.9834 & 0.9886 & 0.8316 & - & \textbf{0.9771} & 0.8690 & 0.9843 & \textbf{0.9920} & 0.8271 & - \\
    \hline
    
    RV-GAN \cite{kamran2021rv} & 14.81 & \textbf{0.9790} & 0.7927 & \textbf{0.9969} & 0.9887 & \textbf{0.8690} & - & 0.9697 & 0.8199 & 0.9806 & 0.9914 & \textbf{0.8957} & -\\
    \hline
    
    FR-UNet \cite{liu2022full} & 7.37 & 0.9705 & 0.8356 & 0.9837 & \textbf{0.9889} & 0.8316 & \textbf{0.7120} & 0.9748 & \textbf{0.8798} & 0.9814 & 0.9913 & 0.8151 & 0.6882\\
    \hline


    AUtO & 7.40 & 0.9646 & 0.8226 & 0.9786 & 0.9812 & 0.8032 & 0.6714 & 0.9736 & 0.8262 & \textbf{0.9866} & 0.9862 & 0.8346 & \textbf{0.7162}\\
    \hline
    
    \end{tabular}}
    \caption{Comparison of AUtO with SOTA methods for vessel segmentation tasks on DRIVE and CHASE-DB1 dataset.}
    \label{tab:mainresult}
\end{table}
\end{landscape}

\begin{table}[!h]
    \centering
    \resizebox{0.6 \linewidth}{!}{
    \begin{tabular}{|c||c|c|c|c|}
    \hline
     Methods & Training Hours & System RAM & GPU RAM\\
    \hline
    \centering RV-GAN & 40 \cite{uysal2021exploring} & - & - \\
    \hline
    \centering FR-UNet & 10 & 3.3 GB & 8.5 GB \\
    \hline
    \centering AUtO & \textbf{0.5} & \textbf{2.8 GB} & \textbf{4.3 GB}\\
    \hline
    \end{tabular}}
    \caption{Comparison of the model efficiency, where RV-GAN takes around 40 hours \cite{uysal2021exploring}. While FR-UNet only requires 10 hours, AUtO is the most efficient and could be trained within 0.5 hours. FR-UNet and AUtO training hours are trained on free Colab GPU with a fair comparison.}
    \label{tab:time}
\end{table}

\subsection{Analysis of Backbone Networks}
\begin{figure}[h!]
  \centering
  \includegraphics[scale=0.25]{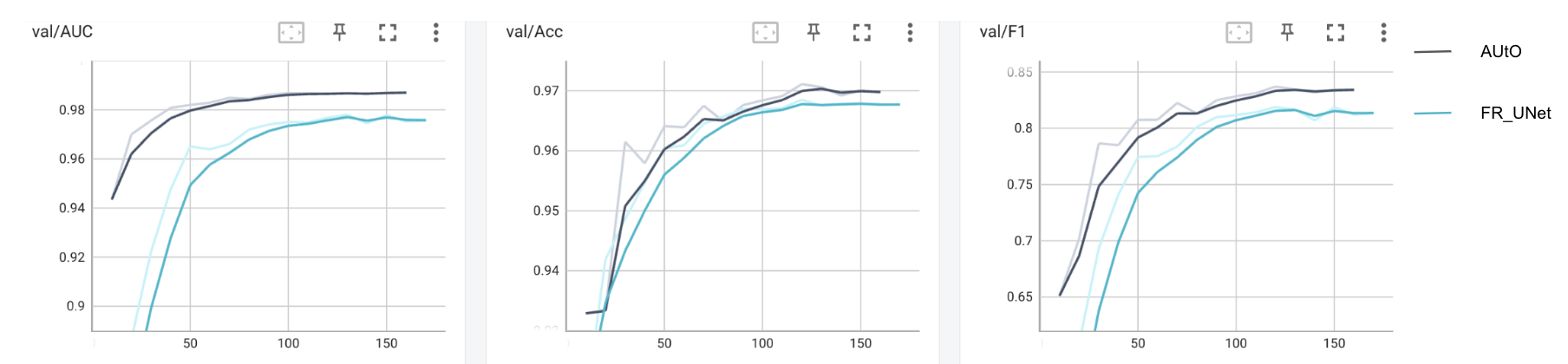}
  \caption{Validation performance between FR-UNet with AUtO.}
  \label{fig:chasedmetric}
\end{figure}

\begin{figure}[h!]
  \centering
  \includegraphics[scale=0.35]{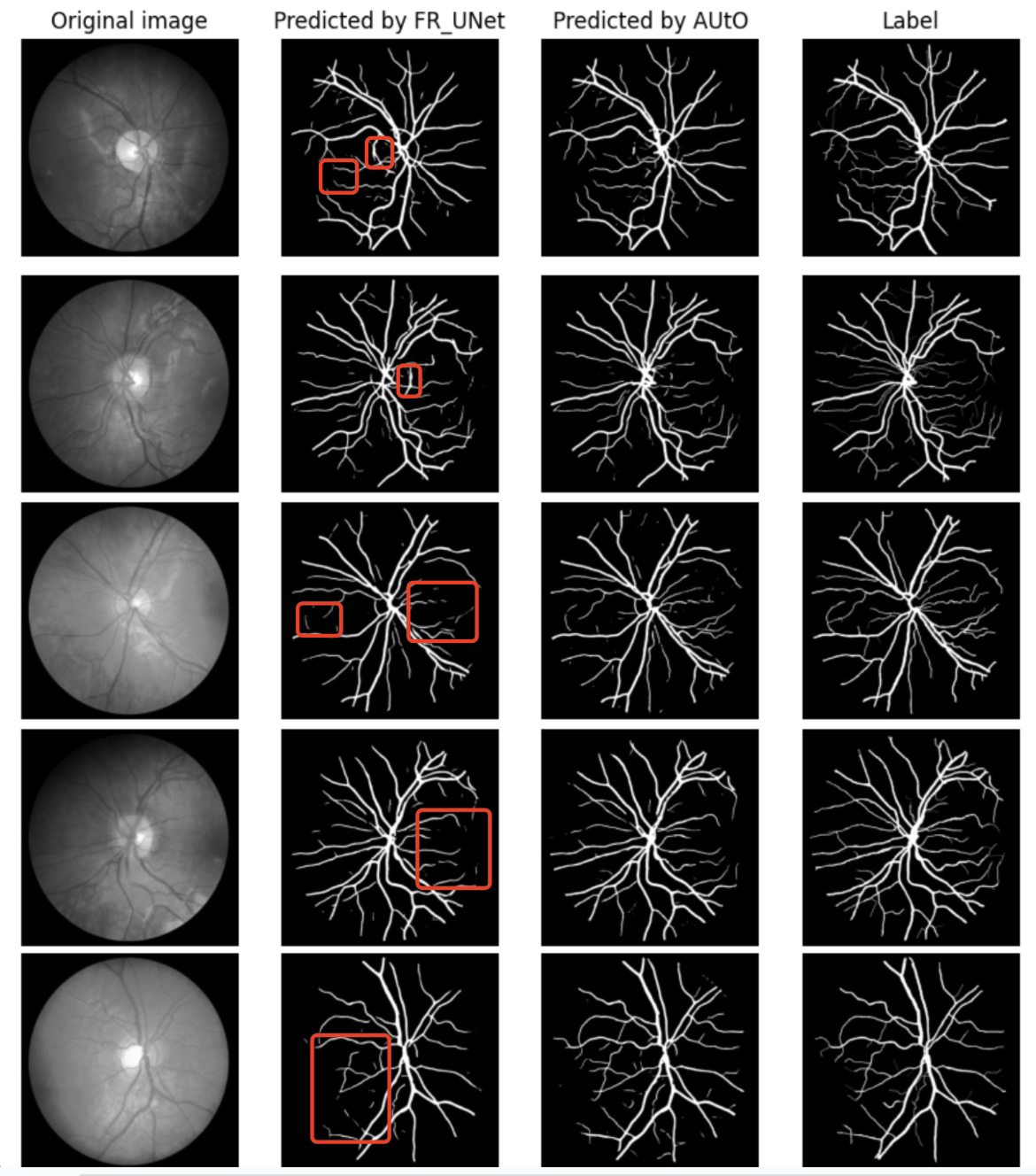}
  \caption{Segmentation predictions between FR-UNet with AUtO in CHASE-DB1.}
  \label{fig:chasedbcompare}
\end{figure}

Figure \ref{fig:chasedmetric} and Figure \ref{fig:chasedbcompare} present a comparative performance assessment of two backbone architectures, FR-UNet and AUtO-Net, on the CHASE-DB1 dataset. Figure \ref{fig:chasedmetric} plots the validation performance, in terms of AUC, ACC, and F1 scores, over approximately 160 epochs. AUtO-Net consistently outperforms FR-UNet across all epochs, demonstrating the substantial improvement imparted by the attention mechanism integrated into this model.

On the other hand, Figure \ref{fig:chasedbcompare} provides a visual comparison of the segmentation results from FR-UNet and AUtO-Net. The superior sensitivity and precision of AUtO-Net are evident in its ability to discern and delineate intricate vessel details with greater accuracy. The red box annotations highlight areas where FR-UNet has misclassified regions, providing a clearer visual contrast.

For instance, in the top images, FR-UNet mistakenly identifies the retina cup as a vessel, a misclassification that AUtO-Net astutely avoids. This example underscores AUtO-Net's superior ability to discern classification boundaries.

Moreover, the bottom four images further illustrate AUtO-Net's superior performance. The reconstructed vessel structures generated by AUtO-Net are markedly more detailed and accurate than FR-UNet.

In conclusion, the attention mechanism within AUtO-Net undoubtedly enhances its performance, making it a more robust and precise choice for vessel segmentation tasks. This comparison underscores the critical role of the right backbone architecture in achieving high-quality segmentation results.

\section{Ablation Study}
\subsection{Modules Analysis}
The ablation study is designed to test the effectiveness of the proposed method: the contrastive multiview learning module and the residual attention module. Therefore, the controlled experiment is designed with four combinations:

\begin{enumerate}
    \item Use baseline only.
    \item Use baseline with the residual attention module.
    \item Use baseline with the contrastive multiview learning module.
    \item Use baseline with both residual attention and contrastive multiview learning module.
\end{enumerate}

Where the baseline method is FR-UNet, and the experiment is done on the DRIVE dataset with a rotation and colour transformation composition for 50 epochs. The result is summarised in Table \ref{tab:module}.

\begin{table}[!h]
    \centering
    \resizebox{0.9\linewidth}{!}{
    \begin{tabular}{|c||c|c|c|c|c|c|c|}
    \hline
     Method & AUC & F1 & ACC & SEN & SPE & PRE & IOU\\
    \hline
    \centering Baseline Only & 0.9628 & 0.7545 & \textbf{0.9623} & 0.6746 & \textbf{0.9901} & \textbf{0.868} & 0.6073\\
    \hline
    \centering Baseline + A & 0.9737 & 0.7706 & 0.9614 & 0.7247 & 0.9854 & 0.8344 & 0.6279\\
    \hline
    \centering Baseline + C & \textbf{0.9743} & \textbf{0.7801} & 0.9618 & 0.7595 & 0.9820 & 0.8083 & \textbf{0.6399}\\
    \hline
    \centering Baseline + A + C & 0.9710 & 0.7739 & 0.9577 & \textbf{0.8279} & 0.9698 & 0.7304 & 0.6333\\
    \hline
    \end{tabular}}
    \caption{Ablation study for the proposed methods, C stands for contrastive multiview learning and A stands for attention block. Compare with the performance of controlled experiments on the baseline method.}
    \label{tab:module}
\end{table}

From the table, the result clearly indicates that both proposed residual blocks with attention mechanisms and contrastive multiview learning framework could lead to a performance increase overall metric (SEN and SPE have a nature trade-off). 

The F1 score with the attention block achieved 0.7706, which is 0.0161 higher than the baseline approach of 0.7545. Moreover, it could be found that the sensitivity (SEN) score also improved from 0.6746 to 0.7247, representing that the attention mechanism indeed have more substantial power to capture tiny vessel structures than pure CNN block. However, the problem is also apparent. Some background noise is also recognised as the vessel object. Therefore, it leads to a drop in specification (SPE) score from 0.9901 to 0.9854.

The baseline with the contrastive multiview learning module obtains the highest F1 score of 0.7801 and the highest IOU score of 0.6399, which is 0.0256 and 0.0326 higher, respectively, than the baseline. Like the attention module, with learning multiple views, the model could also improve the sensitivity from 0.6746 to 0.7595, an approximately 0.0849 increase, and only sacrifices the 0.0081 specification score. Therefore, it could conclude that the contrastive multiview learning framework effectively improves the model's performance.

There is an interesting phenomenon that contrastive multiview learning could still improve the F1 score with the attention-based network. However, the improvement is not obvious compared with the improvement in the convolutional neural network, only 0.0033. However, it could still boost the sensitivity score from 0.6746 of the baseline only and 0.7247 of the baseline with attention block to 0.8279. It could conclude that the proposed contrastive multiview learning module is a general approach that works for CNNs and attention-based networks.

\subsection{Attention Block}
Table \ref{tab:ab} illustrates the comparison between models with and without the self-attention mechanism. The results show that incorporating an attention block does not significantly increase the model parameters, with an increase of only 0.1 million. Moreover, the model's performance with attention blocks is generally superior, with the F1 score approximately 0.01 higher and the IOU Score around 0.013 higher. This evidence supports the hypothesis that the attention mechanism indeed enhances the model's learning capacity.

However, it is worth noting that the difference in AUC between the models is insignificant, suggesting that the attention mechanism might not contribute significantly to the model's overall ability to distinguish between different classes. This observation prompts us to consider the specific areas where the attention mechanism improves performance. The attention mechanism is likely particularly beneficial in focusing on intricate details or subtle features that a model could overlook without attention.

Another point worth considering is the balance between the slight improvement in performance and the additional computational complexity brought about by the attention block. Given that the performance boost is not overwhelming and the model parameters increase, albeit not substantially, it is crucial to consider the computational resources and training time in a practical implementation scenario.

\begin{table}[!h]
    \centering
    \resizebox{0.7 \textwidth}{!}{
    \begin{tabular}{|c||c|c|c|c|c|}
    \hline
     Attention  & Params (M) & AUC & F1 & ACC & IOU\\
    \hline
    \centering No & \textbf{7.37} & 0.9800 & 0.7889 & 0.9662 & 0.6524 \\
    \hline

    \centering Yes & 7.40\ & \textbf{0.9800} & \textbf{0.7990} & \textbf{0.9667} & \textbf{0.6653} \\
    \hline
    
    \end{tabular}}
    \caption{Comparison of the effectiveness of the attention block. The models are trained on the DRIVE dataset without data augmentation for 50 epochs.}
    \label{tab:ab}
\end{table}

Table \ref{tab:abreduce} examines the influence of the computation-performance trade-off reduction ratio $r$. It is observed that a smaller reduction ratio $r$ leads to larger model parameters. However, the increase is not substantial, moving from a reduction ratio of 16 with 7.40 million parameters to a ratio of 1 with 7.66 million parameters. The table illustrates that optimal performance is achieved when the reduction ratio is 2, as indicated by the highest AUC, F1, ACC, and IOU scores of 0.9816, 0.8041, 0.9673, and 0.6726, respectively. Other metrics, such as SEN, SPE, and PRE, are approximately in the second rank.

While the superior performance at a reduction ratio of 2 is noteworthy, it also prompts a deeper analysis of the nature of the trade-off being made. The slight parameter increase does not seem to result in significant performance gains beyond a reduction ratio of 2. The minor variation in performance metrics across different reduction ratios suggests that further reduction in the ratio may lead to diminishing returns regarding model performance.

Another observation point is the fluctuation in the SEN, SPE, and PRE metrics across different reduction ratios. While the ratio of 2 offers the best overall performance, the highest SEN score is achieved at a ratio of 16, and the best SPE and PRE scores are observed at 1 and 4, respectively. This indicates that different aspects of model performance may be sensitive to the reduction ratio in different ways, reinforcing the need to carefully consider the trade-offs involved when selecting an optimal reduction ratio.

\begin{table}[!h]
    \centering
    \resizebox{\linewidth}{!}{
    \begin{tabular}{|c||c|c|c|c|c|c|c|c|}
    \hline
     Reduction & Params (M) & AUC & F1 & ACC & SEN & SPE & PRE & IOU \\
    \hline
    \centering 16 & \textbf{7.40} & 0.98 & 0.799 &0.9667 & \textbf{0.7928} & 0.983 & 0.811 & 0.6653 \\
    \hline
    \centering 8  & 7.41 & 0.9793 & 0.7972 &0.9653 & 0.7808 & 0.9834 & 0.8213 & 0.6634\\
    \hline
    \centering 4 & 7.45 & 0.9779 &0.7928 &0.9666 & 0.7435 & 0.9883 & \textbf{0.8544} & 0.6572 \\
    \hline
    \centering 2 & 7.51 & \textbf{0.9816} & \textbf{0.8041} & \textbf{0.9673} & 0.7871 & 0.9842 & 0.8247 & \textbf{0.6726} \\
    \hline
    \centering 1 & 7.66 & 0.98 & 0.7978 & 0.9673 & 0.7717 & \textbf{0.9855} & 0.8296 & 0.6641 \\
    \hline
    \end{tabular}}
    \caption{Comparison of the model performance with different reduction ratios in the SE attention block.}
    \label{tab:abreduce}
\end{table}

\subsection{Augmentations}
Table \ref{tab:abaug} provides a comprehensive overview of the impact of various data augmentation strategies on model performance. A noteworthy observation from the table is the impressive improvement in model performance when the Contrast Limited Adaptive Histogram Equalization (CLAHE) augmentation technique is employed. This technique achieves the highest AUC, F1, and IOU scores of 0.9824, 0.8045, and 0.6732, respectively. A plausible explanation for this improvement could be the ability of CLAHE to enhance the contrast between the vessel structures and background noise, thereby facilitating a more accurate classification boundary.

Another augmentation strategy that positively impacts performance is the use of flips, as it offers the highest ACC score of 0.9676. Introducing geometric variations through flips could enhance the model's robustness without creating a substantial distribution gap between the training and testing data.

Interestingly, the sharpened augmentation leads to the highest SPE and PRE scores of 0.9893 and 0.8927, respectively, despite a decline in the sensitivity score from 0.7928 to 0.7079. This could be attributed to the fact that the sharpening operation, while enhancing clarity, might cause the model to overlook smaller vessels, making it more challenging to capture intricate details.

The table also sheds light on the efficacy of the proposed MixUp by images method, which attains the highest SEN score of 0.9002. This suggests that the model is proficient at detecting smaller vessel structures. This could potentially be ascribed to the mitigation of the imbalanced sampling issue between the background and vessel classes by MixUp. However, despite this advantage, the overall performance leaves room for improvement. The sum operation might cause a shift in intensity between the training and testing distributions. Hence, future work could focus on refining this aspect of the MixUp method.

Another variant of MixUp, the MixUp by labels, achieves the second-highest scores for SPE and PRE metrics at 0.9882 and 0.8473, respectively. However, it experiences a decline in the SEN score to 0.7100. One possible conjecture for this trade-off could be that as the decision boundary in the training data becomes clearer, the threshold for classifying the vessel class increases. This, in turn, could lead to misclassification of smaller vessels.

\begin{table}[!h]
    \centering
    \resizebox{\linewidth}{!}{
    \begin{tabular}{|c||c|c|c|c|c|c|c|}
    \hline
     Augmentation & AUC & F1 & ACC & SEN & SPE & PRE & IOU\\
    \hline
    \centering None & 0.9800 & 0.7990 &0.9667 & 0.7928 & 0.9830 & 0.8110 & 0.6653\\
    \hline
    \centering CLAHE & \textbf{0.9824} & \textbf{0.8045} & 0.9647 & 0.8304 & 0.9796 & 0.7811 & \textbf{0.6732} \\
    \hline
    \centering Flip & 0.9801 & 0.8028 & \textbf{0.9676} & 0.8164 & 0.9810 & 0.7908 & 0.6707 \\
    \hline
    \centering Rotation & 0.9788 & 0.7932 & 0.9659 & 0.8108 & 0.9796 & 0.7773 & 0.6575 \\
    \hline
    \centering Normalisation & 0.9799 & 0.7971 & 0.9674 & 0.7678 & 0.9860 & 0.8347 & 0.6627 \\
    \hline
    \centering Sharpen & 0.9633 & 0.7714 & 0.9654 & 0.7079 & \textbf{0.9893} & \textbf{0.8927} & 0.6301 \\
    \hline
    \centering Random Contrast & 0.9723 & 0.7706 & 0.9600 & 0.7282 & 0.9835 & 0.8243 & 0.6282 \\
    \hline
    \centering Random Colour & 0.9765 & 0.8015 & 0.9662 & 0.7716 & 0.9857 & 0.8382 & 0.6689 \\
    \hline
    \centering Gaussian Blur & 0.9660 & 0.7477 & 0.9560 & 0.7019 & 0.9820 & 0.8106 & 0.5987 \\
    \hline
    \centering Elastic & 0.9791 & 0.7830 & 0.9656 & 0.7482 & 0.9857 & 0.828 & 0.6435 \\
    \hline
    \centering MixUp-Image & 0.9804 & 0.7398 & 0.9469 & \textbf{0.9002} & 0.9514 & 0.6326 & 0.5887 \\
    \hline
    \centering MixUp-Label & 0.9619 & 0.7688 & 0.9645 & 0.7100 & 0.9882 & 0.8473 & 0.6250 \\
    \hline
    \end{tabular}}
    \caption{The model performance with different augmentations, testing on the DRIVE dataset with 50 epochs.}
    \label{tab:abaug}
\end{table}

\subsection{Contrastive Multiview Learning}
The impact of contrastive multiview learning on the efficacy of different data augmentation strategies is presented in Table \ref{tab:contrastab}. Two salient observations can be drawn from this analysis.

First, the incorporation of contrastive loss consistently enhances the model's performance across all data augmentations. This enhancement is manifested by improving key performance metrics such as ACC, AUC, F1, and IOU scores. It suggests that the contrastive loss framework reinforces the learning process, empowering the model to derive robust features that improve its predictive capabilities.

Second, contrastive loss is not merely a regularisation term but rather a dynamic facilitator that calibrates the balance between sensitivity (SEN) and specificity (SPE). The model exhibits a discrepancy between these two metrics without contrastive loss, hinting at an unstable training process. This instability is likely triggered by the data augmentation itself, which could introduce unexpected variances. However, when contrastive loss is added, the model shows a more balanced performance between sensitivity and specificity, indicating a more stable learning environment. This balance is particularly critical as a trade-off often exists between SEN and SPE - an increase in one may lead to a decrease in the other, given a fixed model performance.

To illustrate, augmentations such as random colour and CLAHE improved the overall accuracy and intersection over union scores when the contrastive loss was integrated. On the other hand, while achieving the highest specificity and precision scores, sharpen and Gaussian blur augmentations demonstrated substantial improvement in sensitivity and F1 scores with contrastive loss, indicating an overall more balanced performance. Furthermore, MixUp-Image and MixUp-Label augmentations, which initially had a high sensitivity score but lower specificity, saw a boost in overall performance and stability with contrastive learning.

In conclusion, contrastive multiview learning consistently enhances model performance across different data augmentations. More importantly, it stabilises during training, reinforcing the balance between sensitivity and specificity. These findings underscore the importance of incorporating contrastive multiview learning in deep learning models, especially when working with data augmentations.

\begin{table}[!h]
    \centering
    \resizebox{\linewidth}{!}{
    \begin{tabular}{|c|c||c|c|c|c|c|c|c|}
    \hline
      Augmentation & Contrast & AUC & F1 & ACC & SEN & SPE & PRE & IOU\\
    \hline
    \centering Random & No & 0.9765 & 0.8015 & 0.9662 & 0.7716 & 0.9857 & 0.8382 & 0.6689 \\

    \centering Colour & Yes & 0.9799 & 0.8036 & \textbf{0.9673} & 0.7945 & 0.9836 & 0.8160 & \textbf{0.6719}\\
    \hline

    \centering CLAHE & No & \textbf{0.9824} & \textbf{0.8045} & 0.9647 & 0.8304 & 0.9796 & 0.7811 & 0.6732 \\
    
    \centering  & Yes & 0.9773 & 0.7991 & 0.9651 & 0.7636 & 0.9858 & 0.8449 & 0.6662\\
    \hline
    
    \centering Sharpen & No & 0.9633 & 0.7714 & 0.9654 & 0.7079 & \textbf{0.9893} & \textbf{0.8927} & 0.6301 \\

    \centering  & Yes & 0.9790 & 0.7900 & 0.9670 & 0.7409 & 0.9882 & 0.8502 & 0.6531\\
    \hline
    
    \centering Gaussian & No & 0.9660 & 0.7477 & 0.9560 & 0.7019 & 0.9820 & 0.8106 & 0.5987 \\
    
    \centering Blur & Yes & 0.9767 & 0.7825 & 0.9642 & 0.7414 & 0.9860 & 0.8336 & 0.6429\\
    \hline
    
    \centering Random & No & 0.9723 & 0.7706 & 0.9600 & 0.7282 & 0.9835 & 0.8243 & 0.6282 \\
    \centering Contrast & Yes & 0.9731 & 0.7736 & 0.9623 & 0.7512 & 0.9825 & 0.8069 & 0.6333\\
    \hline

    \centering MixUp & No & 0.9582 & 0.7248 & 0.9422 & \textbf{0.9022} & 0.9462 & 0.6122 & 0.5711 \\
    
     \centering Image & Yes & 0.9791 & 0.7457 & 0.9488 & 0.8928 & 0.9558 & 0.6491 & 0.5961\\
    \hline
    
    \centering MixUp & No & 0.9619  & 0.7688 & 0.9645 & 0.7100 & 0.9882 & 0.8473 & 0.6250 \\
    
    \centering Label & Yes & 0.9699 & 0.7992 & 0.9642 & 0.7732 & 0.9841 & 0.8354 & 0.6660\\
    \hline
    
    \end{tabular}}
    \caption{Comparison of the model performance With and Without contrastive loss with different data augmentations.}
    \label{tab:contrastab}
\end{table}

\subsection{Batch Size}
Batch size selection is a strategic model training decision that influences computational efficiency and stability. Larger batch sizes allow for enhanced GPU parallelisation, potentially accelerating training speed. However, they also necessitate increased GPU memory. Conventional wisdom suggests that larger batch sizes promote a more stable training process. However, our experimental results, as summarised in Table \ref{tab:abbs}, seem to challenge this assertion.

Contrary to the general belief, we found smaller batch sizes yielding superior performance. Indeed, the highest performance across all metrics—AUC, F1, ACC, SEN, SPE, PRE, and IOU—was achieved with a batch size of 1. As the batch size increased, there was a visible decline in performance.

Several plausible explanations arise for these counter-intuitive findings. First, each input data in our study is a high-resolution image, abundant with information. This means that even a single image provides sufficient data for model training.
Second, each image is associated with some background noise, such as extreme values caused by lighting conditions. Increasing batch sizes might amplify this noise variance, leading to lower model performance.

Lastly, the total dataset comprises approximately 30 images only, each exhibiting similar vessel patterns. Therefore, a single image in a batch already provides a representative snapshot of the entire dataset, ensuring robust training.

In summary, our analysis implies that smaller batch sizes offer advantages in this context due to the high information content in each input image and the similar pattern across the dataset. These findings may encourage re-evaluating batch size selection strategies, particularly in scenarios dealing with high-resolution images and small datasets, practically for medical image segmentation. However, the small batch size can not fully utilise the GPU hardware to compute parallelly, resulting in a slower training time.

\begin{table}[!h]
    \centering
    \resizebox{0.8\linewidth}{!}{
    \begin{tabular}{|c||c|c|c|c|c|c|c|}
    \hline
     Batch Size & AUC & F1 & ACC & SEN & SPE & PRE & IOU\\
    \hline
    \centering 1 & \textbf{0.9825} & \textbf{0.8047} & \textbf{0.9674} & \textbf{0.8322} & \textbf{0.9794} & \textbf{0.7800} & \textbf{0.6734}\\
    \hline
    \centering 2 & 0.9786 & 0.7880 & 0.9642 & 0.8217 & 0.9768 & 0.7576 & 0.6502 \\
    \hline
    \centering 3 & 0.9673 & 0.7749 & 0.9627 & 0.7977 & 0.9772 & 0.7541 & 0.6325 \\
    \hline
    \end{tabular}}
    \caption{Comparison of the model performance with different batch sizes from 1 to 3 on the DRIVE dataset.}
    \label{tab:abbs}
\end{table}

\subsection{Loss Functions}
Loss functions are pivotal in defining the optimisation objectives in segmentation tasks. In our study, we conducted a comparative evaluation of three prominent loss functions: Binary Cross-Entropy Loss (BCE Loss), Focal Loss, and Dice Loss (refer to Chapter \ref{Chapter2} for a detailed discussion).

The BCE Loss, grounded in KL divergence, guides the model to learn the target distribution. The Dice Loss is derived from the F1 score, while the Focal Loss is specifically designed to handle imbalanced class distributions. Our evaluation also included combinations of these loss functions. The results of this analysis are summarised in Table \ref{tab:lossab}.

Our analysis revealed that BCE Loss generally yielded higher AUC, ACC, SPE, and PRE scores. In contrast, the Dice Loss led to a superior SEN score, albeit at the cost of relatively lower SPE and PRE scores. Interestingly, the FocalBCE Loss did not appear to be suitable for this vessel segmentation task among the loss functions considered.

However, the combination of Dice and FocalBCE Loss achieved the highest SEN score of 0.8513. This demonstrates that a well-considered combination of loss functions can leverage their strengths, enhancing performance on specific metrics.

In conclusion, our study highlights the importance of an appropriate choice of the loss function in segmentation tasks. As the choice significantly influences model performance, it should be made judiciously, considering the specific requirements and characteristics of the task at hand.

\begin{table}[!h]
    \centering
    \resizebox{0.9\linewidth}{!}{
    \begin{tabular}{|c||c|c|c|c|c|c|c|}
    \hline
    Loss & AUC & F1 & ACC & SEN & SPE & PRE & IOU\\
    \hline
    \centering BCE & \textbf{0.9800} & 0.7990 & \textbf{0.9667} & 0.7928 & \textbf{0.9830} & \textbf{0.8110} & 0.6653\\
    \hline
    \centering Dice & 0.9687 & 0.7974 & 0.9650 & 0.8264 & 0.9780 & 0.7755 & 0.6632 \\
    \hline
    \centering FocalBCE & 0.9780 & 0.7935 & 0.9648 & 0.8109 & 0.9793 & 0.7836 & 0.6579 \\
    \hline
    \centering Dice + FocalBCE & 0.9724 & 0.7947 & 0.9633 & \textbf{0.8513} & 0.9738 & 0.7493 & 0.6595 \\
    \hline
    \centering BCE + Dice & 0.9764 & \textbf{0.8001} & 0.9658 & 0.8204 & 0.9794 & 0.7856 & \textbf{0.6669} \\
    \hline
    \end{tabular}}
    \caption{Comparison of the model performance by different loss functions on the DRIVE dataset.}
    \label{tab:lossab}
\end{table}
\chapter{Conclusion}
\label{Chapter5}

\section{Summary}

As a culmination of a rigorous honours year, AUtO-Net was proposed to address two critical issues in medical image segmentation: data limitation and the high computational cost associated with deep learning methods. The AUtO-Net showcases four major characteristics, exploiting an augmentation-driven contrastive multiview learning framework in synergy with a hybrid attention-CNN structure.

\BE{Superior performance} AUtO-Net sets a new standard in performance on the retina dataset CHASE-DB1, achieving the highest SPE and IOU scores of 0.9866 and 0.7162, respectively and attaining the highest F1 score of 0.8346 in the UNet family around 0.0195 higher than existing methods.

\BE{Efficiency} When juxtaposed with other state-of-the-art methods, AUtO-Net's training process is significantly more efficient, completing in a mere 30 minutes compared to the 40 hours required by RV-GAN. This translates to an 80-fold acceleration, demonstrating AUtO-Net's remarkable computational efficiency. This efficiency stems from the modification in contrastive multiview learning, focusing solely on the loss function without necessitating additional parameters. The integration of CNN and attention mechanisms further optimizes performance whilst maintaining computational complexity at a manageable level.

\BE{End-to-End Integration} While traditional contrastive learning methods for segmentation are not end-to-end and require a multi-stage training process, AUtO-Net deviates from this norm. It offers a complete end-to-end model, eliminating the need for pre-training encoders with large datasets and the subsequent fine-tuning of the decoder for segmentation. This innovative approach simplifies implementation and makes AUtO-Net a viable model for real-world deployments and applications.

\BE{Generalizability} AUtO-Net extends beyond the specific domain of medical image segmentation. It is not a narrow, application-specific model but rather a broad conceptual and philosophical framework. The contrastive multiview learning framework and the plug-and-play attention block within AUtO-Net can be applied to any machine learning task, including but not limited to Natural Language Processing, Computer Vision, and Audio Recognition. The core premise of the AUtO-Net contrastive multiview learning framework is to amalgamate multiple views into a single, coherent representation to understand invariant characteristics better. The attention mechanism reinforces this by learning longer dependencies, further enhancing performance.

\section{Limitations and Improvements}
However, there are still many works to be done in the future. Here are some limitations which could be improved.

\BE{Data augmentations analysis} A thorough and systematic understanding of data augmentation's effectiveness is necessary but challenging due to the varying nature of data and use cases, which can yield disparate performances. The complexity of conducting experiments to test augmentations is not linear. Each data augmentation comes with multiple hyperparameters that require meticulous selection. Furthermore, the composite effects of multiple data augmentations can affect one another, complicating the assessment of their performance. Therefore, future research should propose new experiments and methodologies to examine the interdependencies of augmentation methods. Two potential solutions exist.

The first solution is to test every possible combination of augmentation methods. However, given the high complexity associated with even a limited number of augmentations, a more quantitative and theoretical analysis may prove more feasible. In future studies, random sampling-based methods such as Monte Carlo Analysis (MCA) might offer a suitable approach to modelling the performance of each augmentation method.

Additionally, the analyses and results presented in Chapter \ref{Chapter3} and Chapter \ref{Chapter4} suggest a distribution gap between the augmented data and the testing data. Although the proposed contrastive multiview learning method could somewhat mitigate this issue, it nonetheless persists. To fully exploit the potential of data augmentation, more advanced methods might be adopted in the future. For instance, domain adaptation \cite{farahani2021brief} could be introduced, treating the augmented data as the source and the training data as the target data. This approach would enable the model to learn domain-invariant features, thereby enhancing its performance. Moreover, the RV-GAN model \cite{kamran2021rv}, an unsupervised approach that generates pseudo labels, might offer inspiration. Given the existence of GAN-based domain adaptation methods \cite{ganin2015unsupervised}, integrating these three techniques into a unified network might fully harness the performance potential of data augmentation.

\BE{Improvement for MixUp} The primary objective of the proposed MixUp method is to enhance the model's generalisation capability. However, experimental results indicate that the MixUp augmentation contributes marginally to peak performance, suggesting potential areas for improvement in its implementation. Additional ablation studies could be conducted to refine the logic underlying MixUp. Its default implementation involves the summing of two randomly selected images, a process that may undermine the fundamental purpose of data augmentation: to simulate additional real-world data and address the issue of limited data size. This summing operation typically shifts the data intensity for each image, resulting in generally brighter images. Given these observations, future research could explore various ways to refine the MixUp logic. One possibility is to employ max, min, or mean operations instead of the sum operation. Another potential approach could involve using a kernel to randomly select sections from multiple images and assemble them into a new image.

\BE{Contrastive multiview learning extensions} Originating as a framework for self-supervised learning, contrastive multiview learning could potentially be adapted to an unsupervised method within the AUtO-Net model, given an adequate volume of training data. Indeed, a considerable amount of unlabelled retina data is readily available on the internet, presenting an excellent opportunity to gather these training images via a web crawler. Therefore, future research geared towards expanding the model's generalisability and reducing data requirements could significantly contribute to the field of medical image segmentation. Additionally, these web images could serve as a new benchmark training dataset, addressing a current gap and enabling the testing of unsupervised vessel segmentation algorithms. Moreover, the hypothesis that increasing the number of views can enhance model performance remains unverified empirically. Therefore, further comprehensive experiments and ablation studies will be required to test this proposition.

\section{Discussions and Insights}
\begin{figure}[h!]
  \centering
  \includegraphics[width = \textwidth]{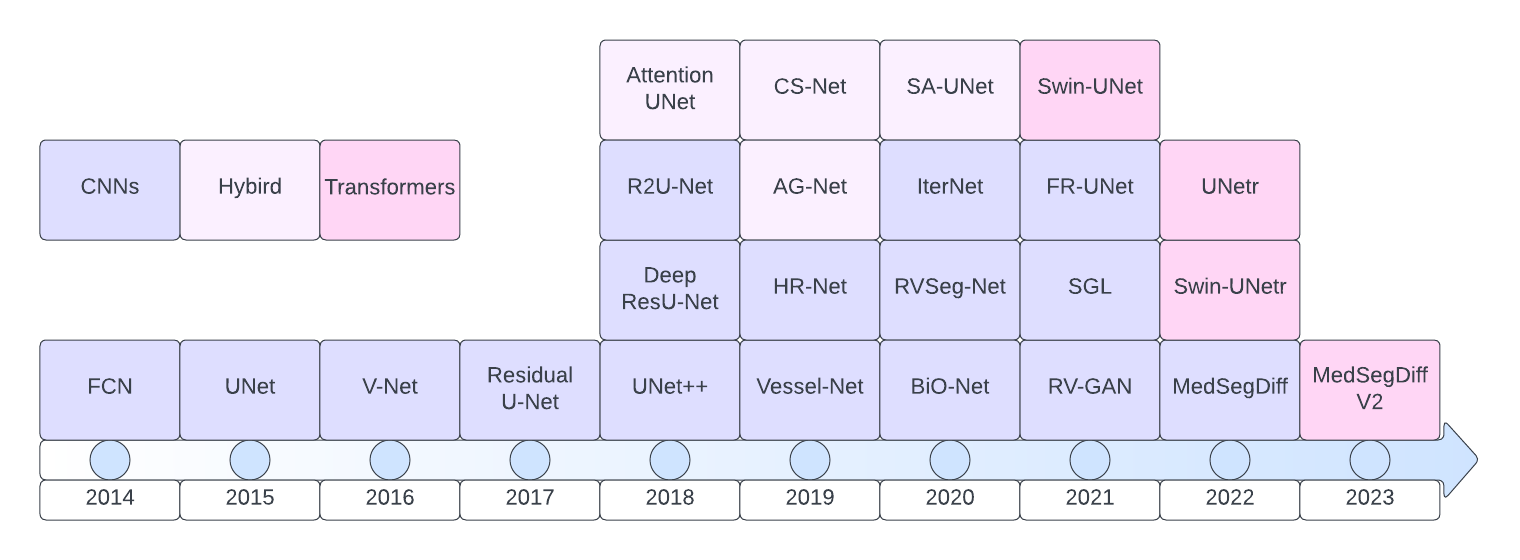}
  \caption{The summarised timeline for proposed deep segmentation networks.}
  \label{fig:capstone}
\end{figure}

There is a trend for the deep learning approach to become the conventional algorithm rather than threshold-based segmentation algorithms for medical image segmentation.

Threshold-based segmentation techniques excel in simple image tasks with well-defined bimodal histograms. However, real-world applications often involve complex images laden with noise originating from lighting conditions or the medical image formation process. These complications result in unsuitable thresholds, suboptimal performance, and difficulties in parameter selection. Rule-based approaches neglect spatial information, struggle to handle noise and blurred boundaries and underperform in general segmentation tasks \cite{295913,haralick1985image}. Hence, algorithms must adapt to intricate image inputs while exhibiting resilience against noise, such as colour and position variations. In contrast, deep learning approaches have revolutionized complex medical image segmentation, emerging as a formidable tool. The impact of deep neural networks is indisputable. However, as the number of hidden layers and parameters increases, neural networks become synonymous with computationally expensive methods.

Figure \ref{fig:capstone} provides a comprehensive summary of the evolution of deep learning models from 2015 to 2023. The trend clearly shows a transition from traditional Convolutional Neural Networks (CNNs) towards hybrid attention networks and the recent emergence of Vision Transformers. This progression suggests that Vision Transformers may become the dominant architecture, potentially superseding conventional CNNs. Moreover, it is now recognised that the segmentation task can also be conceptualised as an image generation task. Notably, the Denoising Diffusion Probabilistic Model (DDPM \cite{ho2020denoising}) sets a new benchmark in performance \cite{wu2022medsegdiff, wu2023medsegdiff}, even when compared with Swin Transformer in medical image segmentation.

However, the computational cost remains a significant hurdle in developing and implementing increasingly complex deep learning models. Several prospective research directions could help mitigate this challenge. First, while the hybrid model, which integrates an attention block into CNNs as AUtO-Net does, offers a robust solution, other possibilities exist. One promising approach is leveraging the advantages of pre-trained models to enhance computational efficiency, particularly within the context of large pre-trained models. The process of training an extensive model with billions of parameters from scratch can be prohibitively time-consuming and financially burdensome. Therefore, exploring methods for efficiently fine-tuning large models \cite{liu2022few} presents a promising research direction. For instance, the AIM model \cite{yang2023aim} effectively employs trainable adapters while freezing the pre-trained model, thereby making the number of trainable parameters manageable. It is rare to find medical vision models that effectively utilise pre-trained models. As such, developing methods for efficient fine-tuning of pre-trained models holds significant potential for advancements in the field.

\bibliography{references}
\bibliographystyle{ieeetr}

\end{document}